\pdfoutput=1
\documentclass[11pt]{article}

\usepackage[letterpaper,margin=1in]{geometry}
\usepackage[round,sort]{natbib}
\usepackage{times}
\usepackage{amsmath,amssymb,amsthm,mathtools}
\usepackage{comment}
\usepackage{bm}
\usepackage{graphicx}
\usepackage{booktabs}
\usepackage{array}
\usepackage{commath}
\usepackage{empheq}
\usepackage{threeparttable}
\usepackage{float}
\usepackage{placeins}
\usepackage{xcolor}
\usepackage{hyperref}
\usepackage{url}

\hypersetup{colorlinks=true,linkcolor=blue!60!black,citecolor=blue!60!black,urlcolor=blue!60!black}

\mathtoolsset{showonlyrefs=true}
\newtheorem{theorem}{Theorem}
\newtheorem{lemma}{Lemma}

\DeclareMathOperator{\diag}{diag}

\title{A Scaling Theory of Multi-Epoch Pretraining:\\
Late-Time Overfitting and Ensembling}

\newcommand{\authorentry}[3]{%
\textbf{#1}\hfill{\scshape #3}\\
\textit{#2}\\[0.6em]}
\author{%
\begin{minipage}{0.92\textwidth}
\raggedright\normalsize
\authorentry{Zhen Yang$^{*}$}{University of Texas at Austin}{zhen.yang@utexas.edu}
\authorentry{Yidi Miao$^{*}$}{Carnegie Mellon University}{yidim@andrew.cmu.edu}
\authorentry{Blake Bordelon}{University of Texas at Austin}{blake.bordelon@austin.utexas.edu}
\end{minipage}}
\date{}

\begin{document}

\maketitle
{\renewcommand{\thefootnote}{$*$}\footnotetext{Equal contribution.}}

\begin{abstract}
In the near future language model pretraining will become bound by total data rather than total compute. Such a regime requires revisiting scaling laws and scaling law theory. Motivated by prior works that have found utility in ensembling, we study how the number of training epochs and ensemble size jointly shape generalization when training data are fixed. In language-model experiments on FineWeb, logit averaging reduces late-training deterioration, delays optimal stopping, and can outperform wider or deeper single models at matched parameter-token budgets over a common training horizon. To analyze this interaction between data reuse and ensembling, we derive a dynamical mean-field theory (DMFT) for multi-epoch stochastic gradient descent in power-law random-feature ensembles, tracking error correlations across epochs and models. Ensembling reduces the prediction variance responsible for late-time growth in expected test error, changing the optimal stopping time. We use these stopping times to derive compute-allocation laws at fixed data and individual model sizes. Under this allocation, ensemble size grows with compute at least as fast as training time per member, while excess test loss above the limiting loss decreases as a power of compute.

\end{abstract}

\section{Introduction}

Scaling laws for one-pass training prescribe how model size and training data
should grow with compute \citep{kaplan2020scaling,hoffmann2022training}. The
supply of high-quality unique text, however, need not grow at the same rate as
compute \citep{villalobos2024run,muennighoff2023scaling, kim2025pre}. Once a corpus is
reused, additional optimization has two opposing effects: it continues to fit
the available data, but it can also amplify noise tied to that particular
dataset, model, or training order. The resulting validation curve is often
non-monotone even though training loss keeps falling.

\textbf{Non-Monotone $\to$ Monotone Validation with Epoching.} In this data-reuse regime, simple small scale language-model experiments can reproduce how validation loss evolves over repeated training epochs and how
ensembling changes this trajectory.
At fixed model size, across
\(10\)--\(100\) million unique FineWeb tokens, validation loss reaches its
minimum after roughly 9--14 epochs and then rises with further training
(Figure~\ref{fig:datasize_sweep}). Averaging independently trained models'
logits reduces this rise and can outperform a wider or deeper model at
matched compute (Figure~\ref{fig:ens_val_vs_t} and
Appendix~\ref{app:compute_matching}). These observations motivate two
questions: which component of late-training error can be reduced by
averaging independently trained models, and how does this reduction
change the loss-minimizing training time?

\textbf{Theoretical Model.} We address these questions using an ensemble of \(E\) random-feature
models, each with \(N\) trainable parameters and an independently sampled
feature matrix. The models train on the same \(P\) examples using
fixed-order stochastic gradient descent (SGD), which reuses the same
ordered minibatches in every epoch. We derive a dynamical mean-field
theory (DMFT) for the training and test losses of individual models and
their averaged predictions. The theory describes these losses through
error correlations within and across epochs and between ensemble members,
together with response functions that measure sensitivity to small
perturbations and quantify overfitting dynamics that build up across epochs.

\paragraph{Relevant Prior Works} Our closest theoretical starting point is the DMFT framework of
\citet{bordelon2024dynamical}, which describes finite-data and
finite-width learning dynamics, including overfitting and ensembling.
We extend this framework to fixed-order, multi-epoch SGD and derive
ensemble-dependent stopping and compute-allocation laws for noiseless
gradient flow at \(N=P\).

The analysis also draws on spectral learning curves
\citep{bordelon2020spectrum,maloney2022solvable,simon2021eigenlearning},
high-dimensional SGD dynamics
\citep{mignacco2020dynamical,gerbelot2022rigorous,paquette2021sgd},
and scaling-law analyses based on Volterra equations
\citep{paquette20244+} and linear-regression risk bounds
\citep{lin2024scaling,yan2025larger}. Our experiments are motivated
by data-constrained scaling \citep{muennighoff2023scaling} and
fixed-data, high-compute pretraining \citep{kim2025pre}.

\vspace{3pt}
\textbf{Contributions.}
(i) We derive a multi-epoch SGD dynamical mean field theory (DMFT) and prove a finite-horizon loss-approximation bound for Gaussian features. Our theory quantifies how both train loss and validation loss evolve over both steps within epochs and across epochs. The theory requires computing correlation and response functions which characterize both optimization progress across steps and also overfitting effects across epoch timescales. 
(ii) We derive ensemble-dependent stopping laws for noiseless Gaussian
gradient flow at \(N=P\), and compute-allocation laws under power-law source/capacity conditions at fixed data and member sizes, with each
ensemble trained to its own optimal stopping time. 
(iii) In language-model experiments on fixed FineWeb corpora, we show
that logit averaging reduces late-training deterioration and delays
optimal stopping. Ensembles can outperform wider or deeper single
models at matched parameter-token budgets over a common training
horizon.

\FloatBarrier

\vspace{-8pt}
\section{Multi-Epoch Pretraining Phenomena}
\label{sec:phenomena}

We measure validation cross-entropy for autoregressive Transformers
trained repeatedly on fixed FineWeb subsets, varying model and corpus
size and averaging logits from independently trained models.
Appendix~\ref{app:experimental_protocol} gives training and evaluation details.

\begin{figure}[t]
    \centering
    \includegraphics[width=0.95\linewidth]{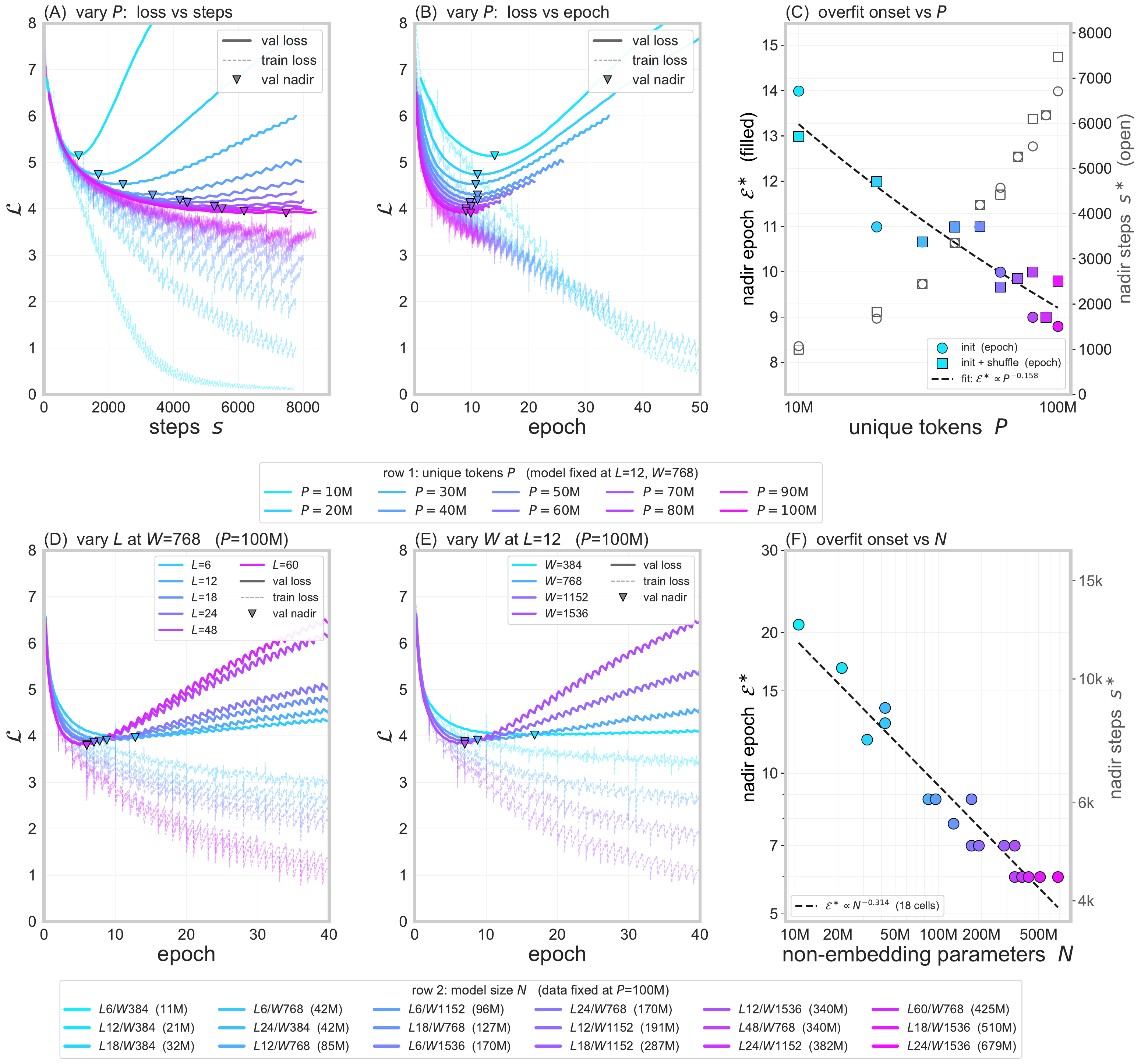}
    \vspace{-4pt}
    \caption{\textbf{Repeated training lowers training loss but eventually raises validation loss.}
    Top: corpus-size sweep at fixed model size; bottom: model-size sweep at fixed
    corpus size. Dashed/solid curves show training/validation loss; triangles mark minima.
    (C, F) Optimal stopping in epochs (filled) and steps (open).
    Larger models reach their optimum in fewer epochs; optimal epochs vary less
    across corpus sizes. Settings: Table~\ref{tab:experiment_design}.}
    \label{fig:datasize_sweep}
    \vspace{-6pt}
\end{figure}

\textbf{Repeated training eventually hurts validation performance.}
Validation loss follows a U-shaped trajectory while training loss
continues to fall (Figure~\ref{fig:datasize_sweep}), consistent with
earlier observations under repeated pretraining data
\citep{muennighoff2023scaling,kim2025pre}.
At fixed model size, validation loss reaches its minimum after roughly
$9$--$14$ epochs across the tested corpus sizes; at fixed data size,
larger models reach their minima earlier
(Section~\ref{sec:model_size}).

\begin{figure}[t]
    \centering
    \includegraphics[width=0.90\linewidth]{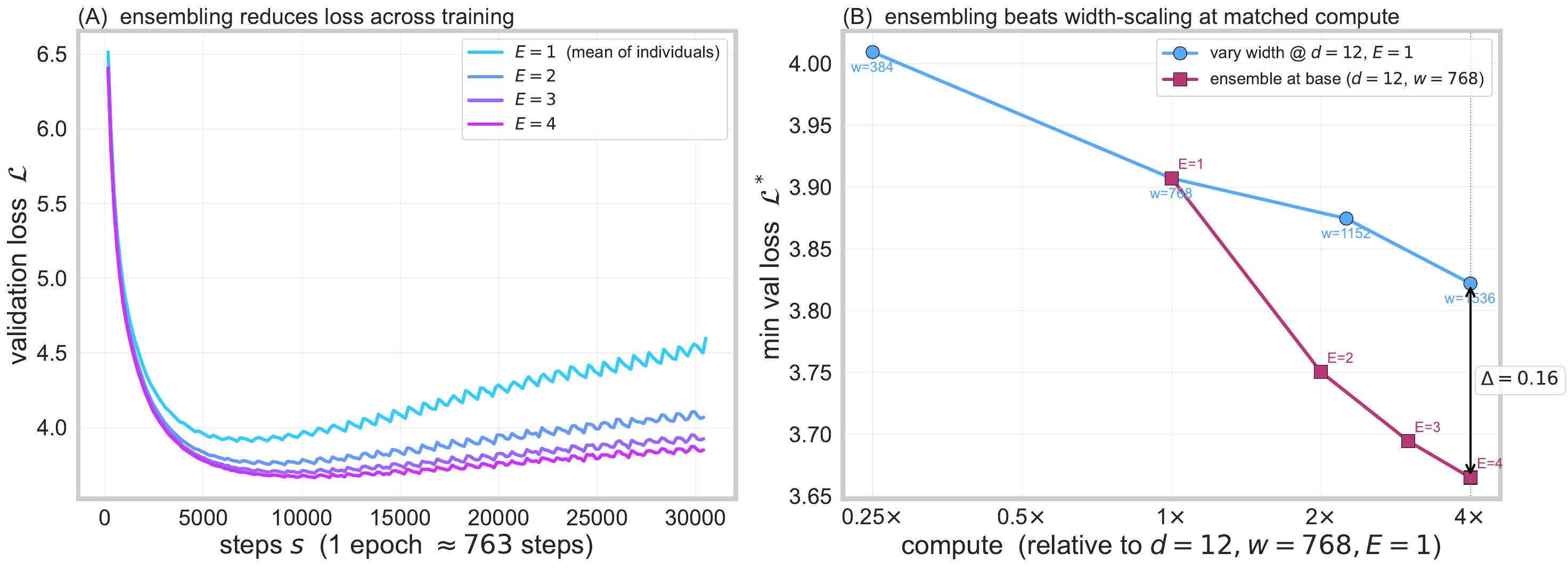}
    \vspace{-4pt}
    \caption{\textbf{Ensembling reduces late-time deterioration and improves loss at matched training budgets.}
    (A) Validation trajectories by ensemble size. (B) Four base models outperform
    one twice-wider model by $0.16$ nats. Both panels use $P=100$M; settings and evaluation coverage: Table~\ref{tab:experiment_design}.}
    \label{fig:ens_val_vs_t}
    \vspace{-6pt}
\end{figure}

\textbf{Ensembling improves the use of a fixed training budget.}
Logit averaging lowers validation loss and reduces its late-time rise
(Figure~\ref{fig:ens_val_vs_t}A).
At matched parameter-token compute, an ensemble of four base models
attains a validation loss $0.16$ nats per token lower than a single
model with twice the width
(Figure~\ref{fig:ens_val_vs_t}B; Appendix~\ref{app:compute_matching}).
Related work also finds gains from ensembling on fixed data
\citep{kim2025pre} and allocating multi-epoch budgets to diverse
predictors \citep{mandal2026q0}.

These observations motivate a model of repeated data use that tracks both
within-model errors and the errors shared across ensemble members.

\vspace{-8pt}
\section{A Dynamical Mean-Field Theory of Fixed-Data Reuse}
\label{sec:dmft}

\subsection{Random-Feature Model and Multi-Epoch Dynamics}

\textbf{Model and observables.}
We study a random-feature predictor of width \(N\), with \(N\) trainable
parameters and \(P\) training examples. SGD uses minibatches of size \(B\):
\begin{equation}
    f(\bm x)
    =
    \frac1N\bm w^\top\bm A\bm\psi(\bm x),
    \qquad
    y(\bm x)
    =
    \bm w^\star\!\cdot\bm\psi(\bm x)+\sigma\epsilon,
    \label{eq:main_model}
\end{equation}
where the training feature vectors are independent draws from
\(\mathcal N(\bm0,\bm\Lambda)\), with
\(\bm\Lambda=\diag(\lambda_1,\lambda_2,\ldots)\).
The matrix \(\bm A\in\mathbb R^{N\times M}\) has independent standard Gaussian
entries, and \(\epsilon\sim\mathcal N(0,1)\). The population features,
random-feature matrix, and label noises are mutually independent.
Here \(\sigma^2\) is the variance of additive label noise, drawn once per
training example, and \(M\) is the population feature dimension. We initialize
the trainable weights at \(\bm w_0(0)=0\); ensemble members use independent
draws of \(\bm A\) on the same dataset. The DMFT
allows \(\sigma\geq0\). We first analyze late-time scaling for noiseless
targets and then give the label-noise extension; the compute-allocation result
uses the noiseless scaling form. Setting \(\sigma=0\) removes label noise,
while the dataset and random features remain random.
Table~\ref{tab:theory_notation} collects the size and training-horizon
notation. For the scaling analysis, we allow infinitely
many population features with summable eigenvalues,
\(\sum_{k=1}^\infty\lambda_k<\infty\), as in dimension-free
analyses of random-feature regression \citep{defilippis2024dimension}.
Appendix~\ref{app:asymptotic_settings} separates this setting from an isotropic
finite-dimensional proportional specialization.

\begingroup
\begin{table}[H]
    \centering
    \caption{Size and training-horizon notation for the random-feature model.}
    \label{tab:theory_notation}
    \begin{tabular}{@{}ll@{\qquad}ll@{}}
        \toprule
        Symbol & Meaning & Symbol & Meaning \\
        \midrule
        \(N\) & Model width & \(M\) & Population feature dimension \\
        \(P\) & Training samples & \(B\) & Samples per update \\
        \(E\) & Ensemble members & \(\mathcal E\) & Training epochs \\
        \(T=P/B\) & Updates per epoch & \(S=T\mathcal E\) & Total updates \\
        \bottomrule
    \end{tabular}
\end{table}
\endgroup

The discrepancy between the target and learned feature coefficients is
\begin{equation}
    \bm v_n^0(t)
    :=
    \bm w^\star-\frac1N\bm A^\top\bm w_n(t).
    \label{eq:main_residual}
\end{equation} For a
fixed realization of the training data and random features, the test
mean-squared error is
\begin{equation}
    \mathcal L_{N,B}(n,t)
    =
    \bm v_n^0(t)^\top\bm\Lambda\bm v_n^0(t)+\sigma^2.
    \label{eq:main_test_loss}
\end{equation}

\textbf{Epoch and minibatch indices.}
We use \(n\in\{0,\ldots,\mathcal E-1\}\) for the epoch index and
\(t\in\{0,\ldots,T-1\}\) for the step within that epoch. The global update
index and effective training time are
\[
    s=nT+t,\qquad \tau=\eta s.
\]
Thus \(s/T\) counts the elapsed epochs. In the fixed-order setting, the same feature block
\(\bm\Psi(t)\in\mathbb R^{B\times M}\) and its fixed label-noise vector
\(\bm\epsilon(t)\) reappear at position \(t\) in every epoch. The exact residual
recursion is
\begin{align}
    \bm v_n^0(t+1)
    &=
    \bm v_n^0(t)
    -\eta
    \left(\frac1N\bm A^\top\bm A\right)
    \frac1B\bm\Psi(t)^\top
    \left(\bm\Psi(t)\bm v_n^0(t)+\sigma\bm\epsilon(t)\right),
    \nonumber\\
    \bm v_{n+1}^0(0)&=\bm v_n^0(T).
    \label{eq:main_residual_update}
\end{align}
Here \(\eta\) is the \(N\)-rescaled learning rate; the corresponding learning
rate on \(\bm w\) is \(\eta N\).

\textbf{Causality and fixed-data reuse.}
To simplify the equations, we use the combined epoch-step index
\(\alpha=(n,t)\), where \(n\) denotes the epoch and \(t\) the step within that
epoch. The causal matrix \(\bm\Theta\) and the projection \(\mathcal D\)
distinguish ordinary temporal dependence from repeated-data dependence:
$$
    \Theta_{\alpha , \beta}
    =
    \delta_{nn'}\mathbf 1\{t>t'\}+\mathbf 1\{n>n'\},
    \;\; \alpha=(n,t),\;\; \beta=(n',t'),
    \;\;
    [\mathcal D(\bm Q)]_{\alpha, \beta} 
    =
    Q_{(n,t),(n',t')}\delta_{tt'}.
    \label{eq:main_D}
$$
The matrix \(\bm\Theta\) sums updates that occur before a given state. The map
\(\mathcal D\) sets entries with \(t\neq t'\) to zero, thereby retaining
cross-epoch correlations only between repeated uses of the same minibatch.
\vspace{-10pt}
\subsection{High Level Findings from Theory}
\vspace{-5pt}
\label{sec:theory_findings}

\textbf{Overfitting Effects from Multi-Pass SGD and Cross-Epoch Response Functions.}
The theory describes how cross-epoch correlations build up as the same
data are reused. These effects are captured by the DMFT response and
correlation functions, which predict how validation loss and training
loss separate as epoch count grows.

\textbf{Regimes of Training in Terms of Compute.}
Our theory predicts that Chinchilla-style compute-optimal scaling breaks
down around the compute scale \(C\sim P^\chi\), where \(\chi=b+1\) for
the power-law spectrum studied below.
\vspace{-2pt}
\begin{itemize}
    \vspace{-5pt}
    \item \textbf{Effectively Online Regime}:
    At low compute, when data-reuse effects are small, the leading
    scaling follows the effectively online approximation used for
    one-pass training.
    \item \textbf{Possible Overfit Regime}:
    Once the model size selected by the online allocation becomes
    comparable to the dataset size, overfitting can affect the optimal
    use of compute. Optimal stopping and approaches such as ensembling
    then become important.
    With more compute, training time, model size, and ensemble count
    can be adjusted to further reduce test loss using the same data.
\end{itemize}

\textbf{Optimal Stopping and Ensembling as a Function of Compute.}
For noiseless gradient flow at the interpolation threshold, the theory
predicts growing expected validation loss at late times. This growth
can be controlled through a combination of early stopping and
ensembling. We characterize how the optimal stopping time and ensemble
count depend on compute at fixed data and member size
(Section~\ref{sec:compute_allocation}).

\subsection{DMFT Characterization}

Averaging over the random matrices reused in
\eqref{eq:main_residual_update}, DMFT gives a Gaussian single-site
process
\citep{mignacco2020dynamical,gerbelot2022rigorous,bordelon2024dynamical,bordelon2026disordered}.
The response \(R_{\alpha,\beta}\) measures the effect at \(\alpha\) of a
perturbation at \(\beta\); \(C_{\alpha,\beta}\) is a two-time correlation.
Bold symbols denote matrices on the \(S=T\mathcal E\) epoch-step indices.
The response equations are
\begin{align}
    \bm H_k
    &=
    \left[\bm I+\eta\lambda_k\bm\Theta\bm R_3\bm R_1\right]^{-1},
    &
    \bar{\bm H}&=\sum_k\lambda_k\bm H_k,
    \nonumber\\
    \bm R_1
    &=
    \left[\bm I+\frac\eta B
    \mathcal D(\bar{\bm H}\bm\Theta\bm R_3)\right]^{-1},
    &
    \bm R_3
    &=
    \left[\bm I+\frac\eta N
    \bm R_1\bar{\bm H}\bm\Theta\right]^{-1}.
    \label{eq:main_response_closure}
\end{align}

The transfer matrix \(\bm H_k\) propagates spectral mode \(k\). The responses
\(\bm R_1\) and \(\bm R_3\) contain the finite-data and finite-width feedback,
respectively.

Writing \(v^1_\alpha\) for the residual on a current minibatch example,
including label noise, the correlations and losses are
\begin{align}
    [\bm C_0]_{\alpha,\beta}
    &=\sum_k\lambda_k
      \left\langle v^0_{k,\alpha}v^0_{k,\beta}\right\rangle,
    &
    [\bm C_1]_{\alpha,\beta}
    &=\left\langle v^1_\alpha v^1_\beta\right\rangle,
    \nonumber\\
    \mathcal L_{\mathrm{test}}(\alpha)
    &=[\bm C_0]_{\alpha\alpha}+\sigma^2,
    &
    \widehat{\mathcal L}_{\mathrm{train}}(\alpha)
    &=[\bm C_1]_{\alpha\alpha}.
    \label{eq:main_dmft_losses}
\end{align}
The brackets average over the effective process. The correlation equations
close with \eqref{eq:main_response_closure}; the minibatch-feature field has
covariance proportional to \(\mathcal D(\bm C_1)\), retaining correlations
between repeated uses of the same minibatch. Appendices~\ref{app:cavity_deriv}
and~\ref{app:msr_path_integral} derive the process, and
Appendices~\ref{app:single_site_processes}--\ref{app:vectorized_dmft} give the
full equations.

\begin{figure}[t]
    \centering
    \makebox[0.455\linewidth][c]{\includegraphics[height=1.8in]{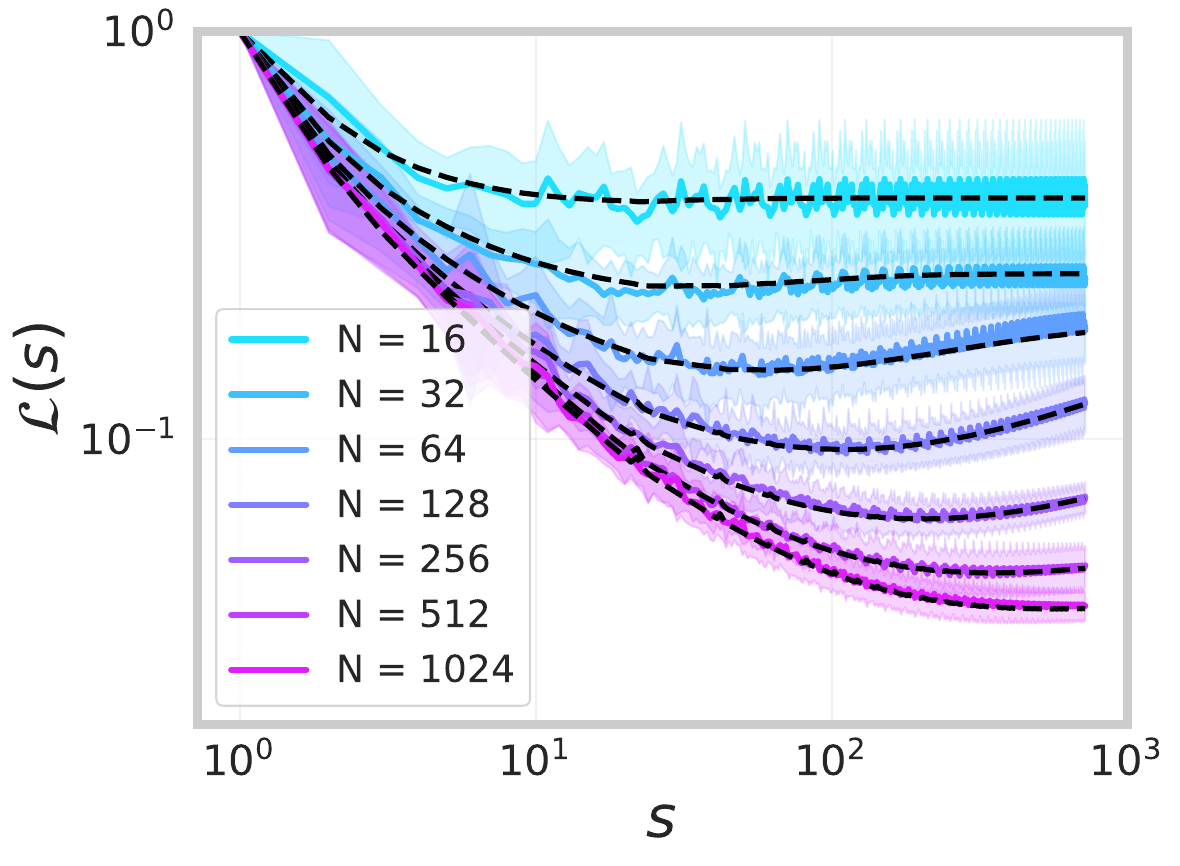}}\hspace{0.02\linewidth}%
    \makebox[0.455\linewidth][c]{\includegraphics[height=1.8in]{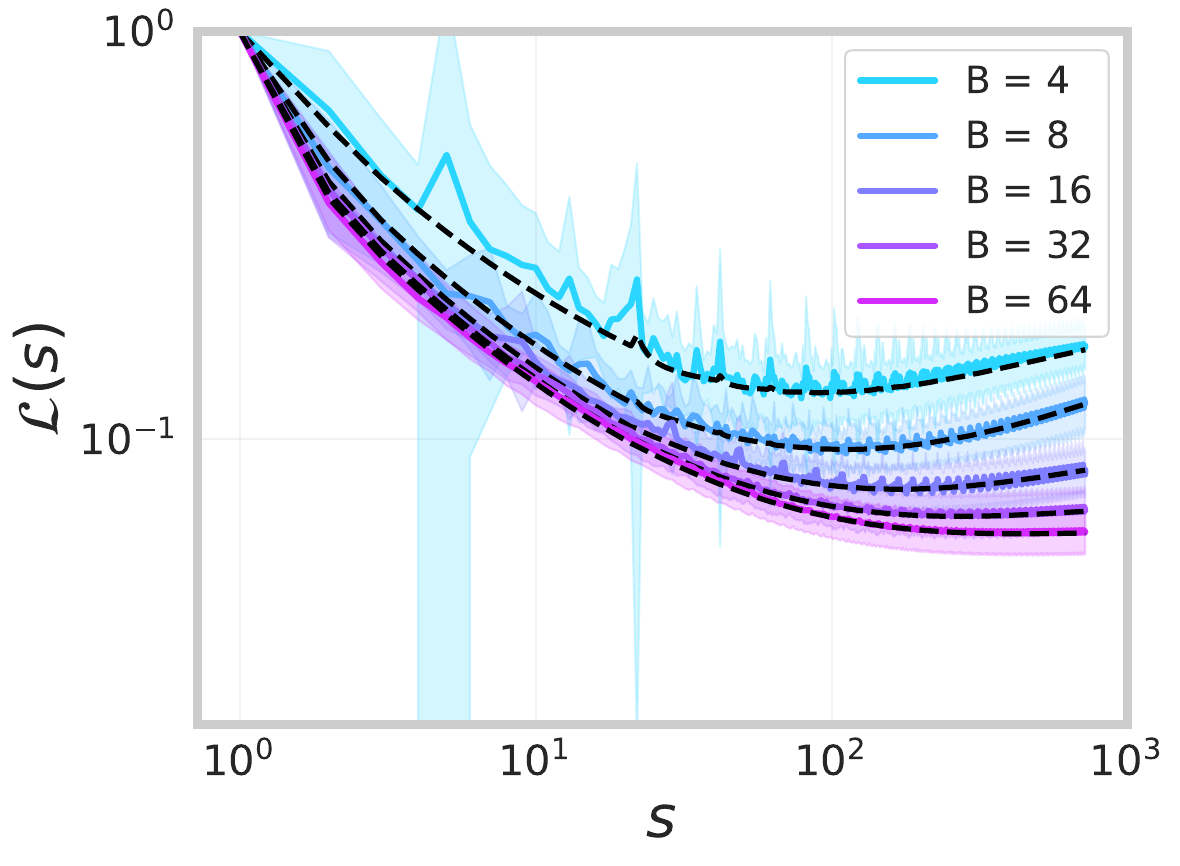}}\\[2pt]
    \makebox[0.455\linewidth][c]{\includegraphics[height=1.8in]{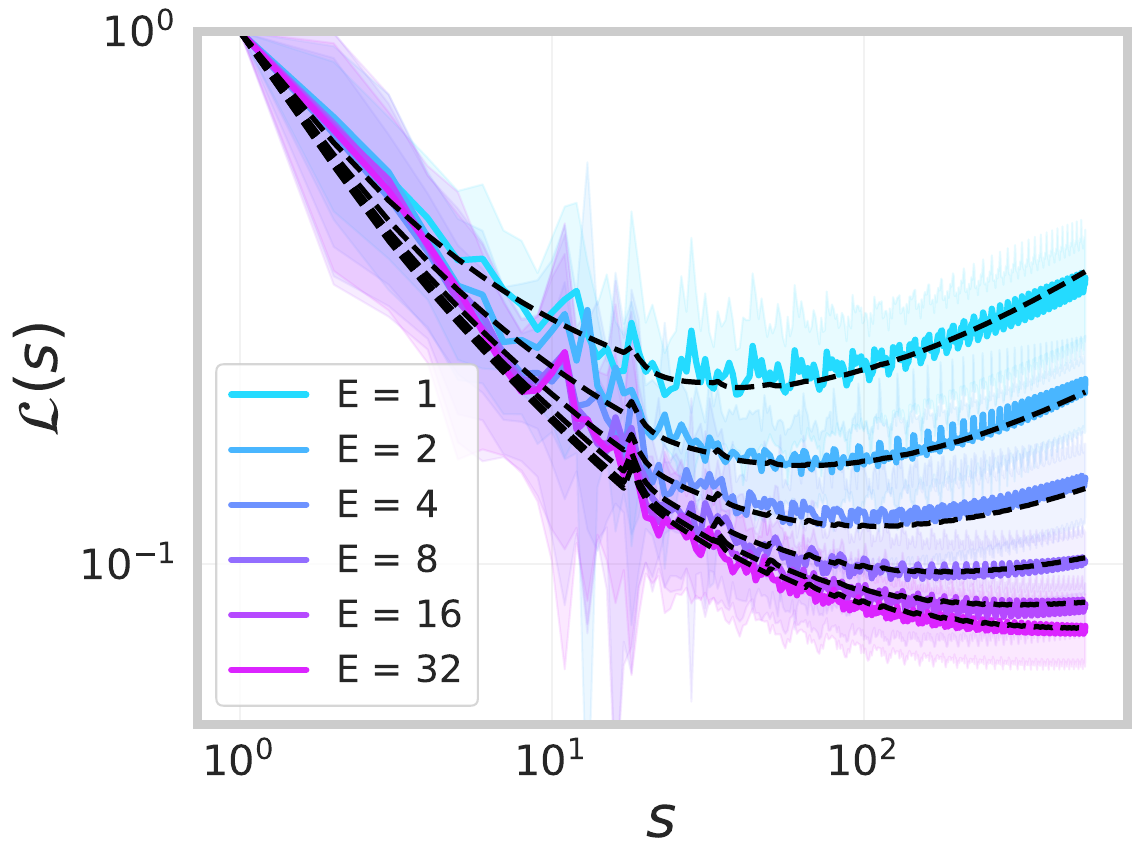}}\hspace{0.02\linewidth}%
    \makebox[0.455\linewidth][c]{\includegraphics[height=1.8in]{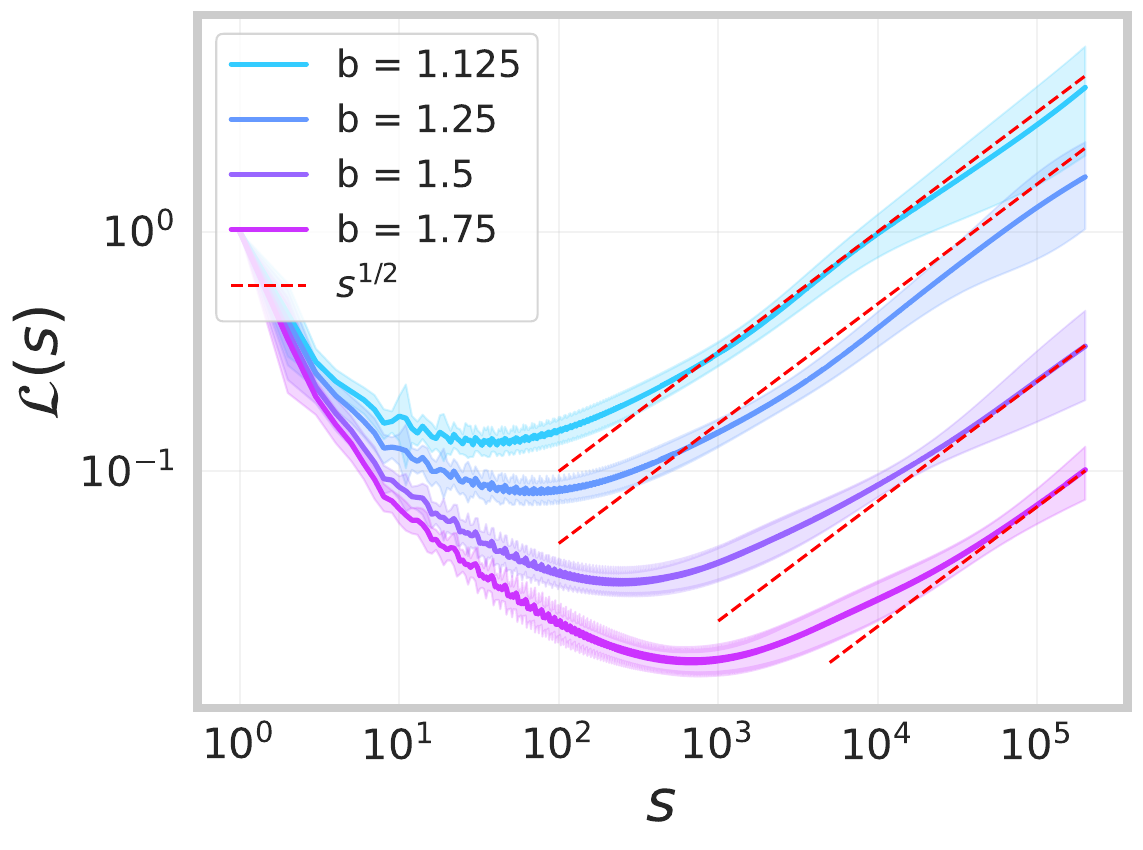}}
    \vspace{-4pt}
\caption{Microscopic random-feature simulations and DMFT predictions
    (dashed black) versus global step. Panels vary model width \(N\), batch size
    \(B\), ensemble size \(E\), and capacity exponent \(b\) at interpolation.}
    \label{fig:dmft_examples}
\vspace{-10pt}
\end{figure}

Figure~\ref{fig:dmft_examples} compares the closed DMFT equations with direct
simulation of the microscopic random-feature model. The theory tracks changes
in width \(N\), batch size \(B\), and ensemble size \(E\), as well as the
late-time growth at interpolation.

\vspace{-8pt}
\subsection{Late-Time Growth Near Interpolation}\label{sec:late_growth}

We study the gradient-flow limit under power-law source and capacity conditions
\citep{bordelon2024dynamical}:
\begin{equation}
    \lambda_k\sim k^{-b},
    \qquad
    \lambda_k(w_k^\star)^2\sim k^{-a},
    \qquad a>1,\quad b>1.
    \label{eq:main_power_law}
\end{equation}
Here \(b\) describes feature-spectrum decay and \(a\) describes target power;
\(b>1\) makes the eigenvalue tail summable. Write
\(\mathcal L(\tau,N,P,E)\) for ensemble test loss at effective time \(\tau\).
We first take \(\sigma=0\) and \(1<a<b+1\), for which the full signal source
scales as \(P^{-(a-1)}\)
(Appendix~\ref{app:scaling_form_interpolation_growth}).

At \(N=P\), the response becomes marginal at low frequencies on the scale
\(\tau_c\asymp P^b\). Under the conditions of
Theorem~\ref{thm:app_late_time_critical_growth}, the growing contribution
\(\mathcal L_{\mathrm{grow}}(\tau)\) to a single model's test loss
(\(E=1\)) is
\begin{equation}
    \mathcal L_{\mathrm{grow}}(\tau)
    \asymp
    P^{-(a-1+b/2)}\tau^{1/2},
    \qquad
    \tau\gtrsim\tau_c\asymp P^b.
    \label{eq:main_late_growth}
\end{equation}
On the regular GF branch at fixed \(N/P\ne1\), a positive low-frequency
denominator cuts off this critical growth.
Before finite data limit learning, the unlearned modes give a contribution
of order \(\tau^{-(a-1)/b}\) for \(\tau\ll P^b\). To determine when to stop
an ensemble, we also need the time dependence of the error that remains
after averaging over models, which we analyze next.

Label noise adds a source to the correlation equations without changing
the responses. Its conditional late-time growth and weak-noise stopping
conditions are derived in Appendices~\ref{app:label_noise_growth}
and~\ref{app:weak_label_noise_stopping}; the stopping and allocation laws
below concern noiseless targets.

\subsection{Ensembling and Optimal Stopping}\label{sec:mechanism}

Consider \(E\) models with independent random projections trained on a
shared dataset in a common fixed order. Write \(C_{\mathrm{self}}(\tau)\)
for one member's expected squared error and \(C_{\mathrm{cross}}(\tau)\)
for the expected product of errors from two distinct members on the same
input. At \(N=P\), the ensemble loss is
\begin{equation}
    \mathcal L(\tau,P,P,E)
    =C_{\mathrm{cross}}(\tau)
    +\frac{C_{\mathrm{self}}(\tau)-C_{\mathrm{cross}}(\tau)}{E}.
    \label{eq:main_local_scaling}
\end{equation}
The difference \(C_{\mathrm{self}}-C_{\mathrm{cross}}\) is the prediction
variance across random projections, averaged over datasets. Ensembling
reduces it by \(1/E\) at fixed time, while cross includes both bias and
error due to the shared data, as in previous analyses of ensembling
\citep{adlam2020understanding,geiger2020scaling,bordelon2024dynamical,ruben2023learning,ruben2025no}.
Appendix~\ref{app:ensembling_theory} derives the ensemble process.

For Gaussian data and random features, zero initialization, and the
pure-power continuum spectrum \(\lambda_k=k^{-b}\),
\((w_k^\star)^2=k^{b-a}\), with \(b>1\) and \(1<a<b+1\), noiseless
GF equations at \(N=P\) give
\begin{align}
    C_{\mathrm{cross}}(\tau)
    &=C_\infty+J\tau^{-1/2}+D\tau^{-1}+O(\tau^{-3/2}),
    \label{eq:main_cross_late}\\
    C_{\mathrm{self}}(\tau)&=H\sqrt\tau+O(1),
    \qquad D,H>0.
    \label{eq:main_self_late}
\end{align}
Here \(C_\infty\asymp P^{-(a-1)}\), and \(J\) has the sign of \(a-b\).
Since cross stays bounded, only the variance grows at late times.
At fixed \(P\) and large \(E\), the loss-minimizing time \(\tau_\star(P,E)\)
balances this growth with the cross correction: \(J\tau^{-1/2}\) for \(a>b\), or \(D/\tau\) when
\(a=b\) and \(J=0\). For \(a<b\), cross approaches \(C_\infty\) from below
and attains a finite-time minimum; stopping times approach its minimizer set.
These cases give the stopping laws in Table~\ref{tab:main_stopping_compute}
(Appendix~\ref{app:continuum_cross_stopping}).

The \(a=b\) cancellation follows from equal squared target coefficients,
not merely equal tail exponents. For the finite noiseless Gaussian model
with zero initialization, \(N=P\), a positive spectrum, \(M>P+1\), and
\((w_k^\star)^2=q>0\), Theorem~\ref{thm:app_new_finite_stopping} proves
\(\tau_\star\asymp E^{2/3}\) for the expected loss at fixed \(P,M\) as
\(E\to\infty\). The additional \(P^b\) scaling comes from the continuum
calculation.

To connect this variance reduction with our language-model experiments,
we consider logit averaging under cross-entropy.
For a fixed dataset and evaluation input, a Taylor expansion of
cross-entropy gives a leading \(1/E\) correction under logit averaging
at a fixed checkpoint (Appendix~\ref{app:logit_ensemble_ce}).

\vspace{-8pt}
\subsection{Compute Allocation at Optimal Stopping}
\label{sec:compute_allocation}

We study how to allocate additional compute between ensemble size
and training time when dataset and individual model sizes are fixed.
Under the noiseless continuum assumptions of
Section~\ref{sec:mechanism}, we fix \(P\) and \(N=P\).
For each ensemble size \(E\), let \(\tau_\star(P,E)\) be the
loss-minimizing training time and \(\mathcal L^*(P,E)\) the
corresponding minimum test loss.
At fixed batch size and learning rate, this training requires compute
\(EP\tau_\star(P,E)\), giving the allocation problem
\begin{equation}
    \min_{E\in\mathbb N_{\ge1}}\mathcal L^*(P,E)
    \quad\text{subject to}\quad
    EP\tau_\star(P,E)\le C,\qquad P\text{ fixed}.
    \label{eq:main_stopped_allocation}
\end{equation}
Ensembling reduces prediction variance, so the minimum loss is
nonincreasing in \(E\). An optimal choice \(E_\star(C;P)\) is therefore
the largest ensemble that can reach its loss minimum within budget.

Write \(\mathcal L^*(C;P)\) for the resulting minimum loss.
The excess loss in Table~\ref{tab:main_stopping_compute} is measured
relative to \(\mathcal L_\infty(P)\), the best test loss approached
as ensemble size grows and training time is chosen optimally.
This limit is of order \(P^{-(a-1)}\) and reflects the error shared
across models (Appendix~\ref{app:compute_optimal_scaling}).
For \(a<b\), the excess-loss rate assumes positive minimum
prediction variance over the times minimizing \(C_{\mathrm{cross}}\).

\begingroup
\begin{table}[H]
    \centering
    \caption{Optimal stopping and compute allocation at fixed \(N=P\).
    Entries give scalings up to positive constants as \(E\to\infty\)
    for stopping and \(C/P^{b+1}\to\infty\) for allocation and loss.}
    \label{tab:main_stopping_compute}
    \begin{tabular}{@{}cccc@{}}
        \toprule
        Condition & Optimal stopping & Ensemble size & Excess loss \\
        & \(\tau_\star(P,E)/P^b\) & \(E_\star(C;P)\)
        & \(P^{a-1}[\mathcal L^*(C;P)-\mathcal L_\infty(P)]\) \\
        \midrule
        \(a>b\) & \(E\)
        & \((C/P^{b+1})^{1/2}\)
        & \((C/P^{b+1})^{-1/4}\) \\
        \(a=b\) & \(E^{2/3}\)
        & \((C/P^{b+1})^{3/5}\)
        & \((C/P^{b+1})^{-2/5}\) \\
        \(a<b\) & \(1\)
        & \(C/P^{b+1}\)
        & \((C/P^{b+1})^{-1}\) \\
        \bottomrule
    \end{tabular}
\end{table}
\endgroup

At large budgets, ensemble size grows at least as fast as training
time per model. For \(a>b\), the two grow with the same power of
compute; for equal squared target coefficients (\(a=b\)), ensemble
size grows faster. In both cases, averaging suppresses the growing
variance enough to make longer training worthwhile.
For \(a<b\), the optimal training horizon remains bounded, so further
compute increases support more ensemble members.

Allowing any affordable checkpoint gives a broader optimization over
ensemble size and training time. At \(N=P\) with equal squared target
coefficients, this compute frontier retains the \(-2/5\) excess-loss
exponent, with a different leading coefficient
(Appendix~\ref{app:width_ensemble_allocation}).

For finite-step SGD, batch size and learning rate also affect the
effective training time available at fixed compute
(Appendix~\ref{app:finite_step_batches}).
Figure~\ref{fig:rf_compute_summary} illustrates these  choices
in the finite random-feature model.

\vspace{-8pt}
\subsection{Testing the Theoretical Predictions}\label{sec:tests}

{We first give a finite-horizon guarantee for the loss approximation, then examine the stopping and compute-allocation predictions numerically.}

\noindent\textbf{Finite-horizon accuracy.}
For Gaussian population features, fixed feature dimension \(M\), and a
fixed horizon \(S\), Theorem~\ref{thm:app_finite_horizon_error} bounds the
absolute loss error by
\(\widetilde{\mathcal O}_{M,S}(N^{-1/2}+B^{-1/2})\), simultaneously for
all \(\alpha\leq S\), at a sufficiently small fixed learning rate.
The same rate holds for relative loss error when the reference loss is
bounded below on this horizon. Appendix~\ref{app:error_estimate_proof}
proves the bound by comparison with population gradient descent.

\begin{figure}[t]
    \centering
    \includegraphics[width=0.80\linewidth]{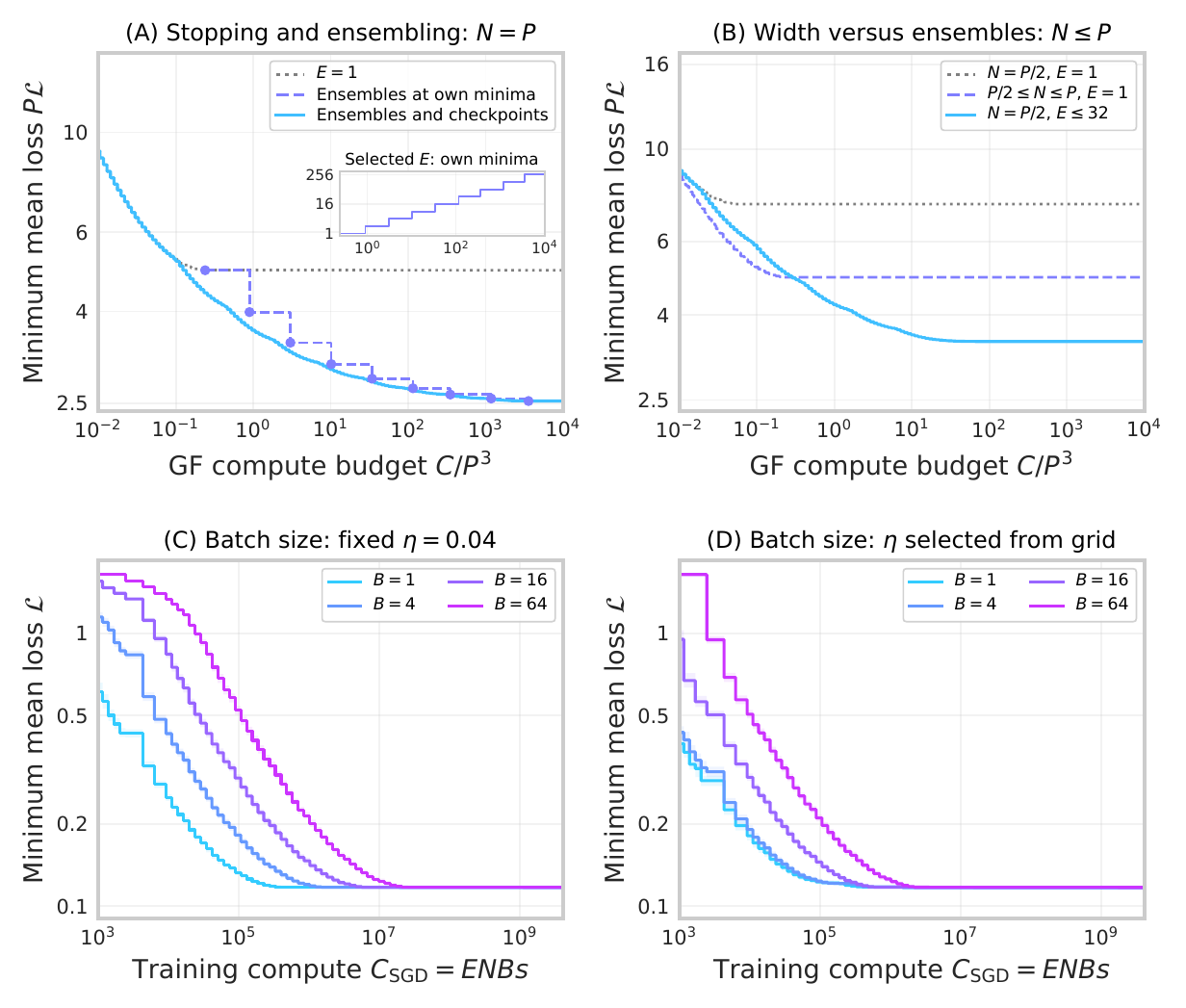}
    \vspace{-6pt}
\caption{\textbf{Compute allocation at fixed data size.}
    Noiseless random-feature simulations with \(a=b=2\).
    (A) In GF at \(N=P\), ensembling lowers loss beyond the single-model
    minimum. Inset: selected \(E\) when ensembles are trained to their own
    minima. (B) Starting from \(N=P/2\), ensembling outperforms increasing
    width up to \(P\) at large budgets on this grid.
    (C, D) For cyclic SGD at \(N=P/2\) and \(E=1\), selecting learning
    rates separately narrows the small-batch advantage.
    Shading indicates 95\% dataset-bootstrap intervals.
    See Appendix~\ref{app:rf_numerics} for settings and comparisons.}
    \label{fig:rf_compute_summary}
\vspace{-15pt}
\end{figure}

\noindent\textbf{Compute allocation in the random-feature model.}
Figure~\ref{fig:rf_compute_summary} compares ways to use additional compute
at fixed data size. In (A), the dashed curve and inset implement the stopped
allocation in \eqref{eq:main_stopped_allocation}; the solid curve jointly
selects ensemble size and an affordable checkpoint. Panel (B) starts below
interpolation and compares increasing width with ensembling, while (C, D) show how
learning-rate selection changes the batch-size comparison.
Appendix~\ref{app:rf_numerics} gives the full learning curves, self/cross
correlations, stopping-law comparisons, and numerical checks, together
with allocation results for widths above interpolation.

\vspace{-8pt}
\subsection{Language-Model Experiments}\label{sec:model_size}

{We now return to the FineWeb experiments to examine whether ensembling
suppresses late-time deterioration and delays optimal stopping in language
models, and to compare its benefits with those of increasing model size.}
The experiments use the FineWeb setup introduced in
Section~\ref{sec:phenomena}; Appendix~\ref{app:experimental_protocol} collects
the configurations and fitting procedures.

\noindent\textbf{Ensembling suppresses the late-time rise.}
The loss trajectories in Figure~\ref{fig:ens_val_vs_t} show that averaging
independently trained models reduces late-time deterioration. At fixed
training steps, validation loss also decreases with ensemble size for both
shared and independently shuffled data orders. Appendix~\ref{app:ensemble_protocol} shows that reducible loss decays approximately as $E^{-1}$ for both, as the theory predicts.

\begin{figure}[t]
    \centering
\includegraphics[height=0.34\linewidth]{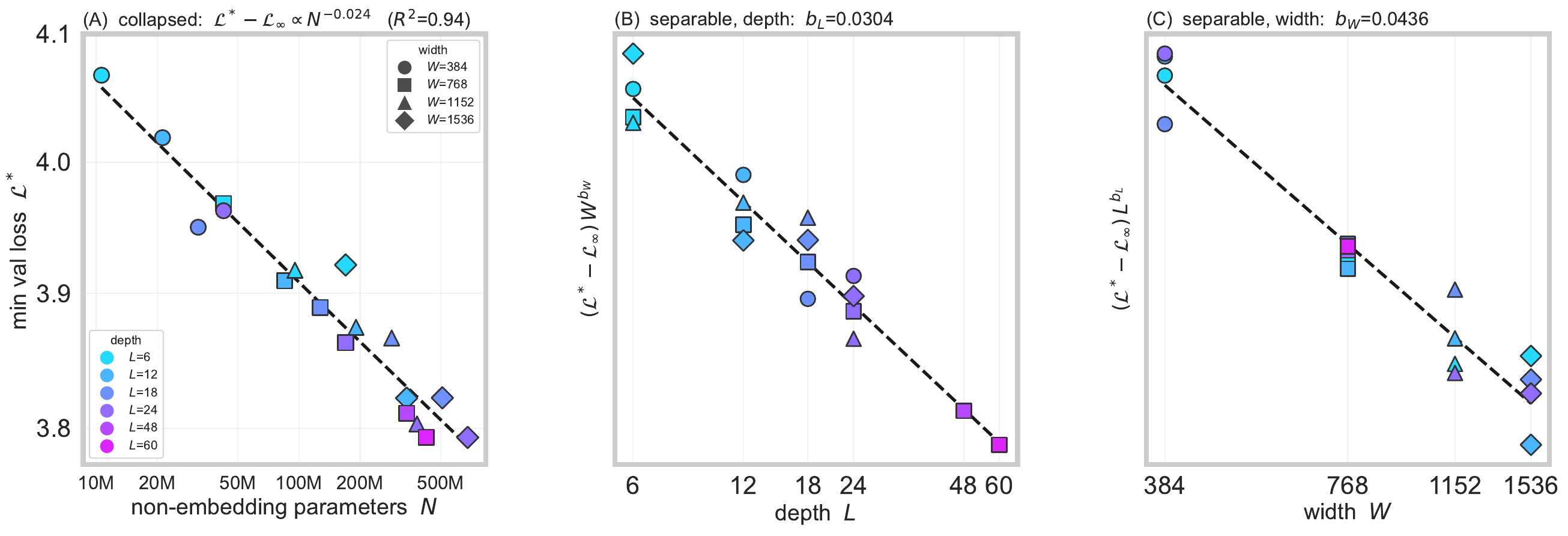}
\vspace{-11pt}
\caption{\textbf{Larger models attain lower loss, with additional depth and width effects.}
At $P=100$M: (A) minimum validation loss versus non-embedding parameter count;
(B, C) depth and width dependence after removing the other variable's fitted
contribution (joint $R^2=0.97$). Shape denotes width and colour denotes depth.
Shared floor: Appendix~\ref{app:asymptotes}; fitted coefficients:
Table~\ref{tab:stopping_exponents}.}
\label{fig:model_size_law}
\vspace{-6pt}
\end{figure}

\noindent\textbf{Model size affects attainable loss and optimal stopping.}
We next vary model depth and width while holding the corpus at $100$M tokens.
Figure~\ref{fig:model_size_law}A shows that larger models generally attain
lower validation loss. The separable relation
\(
\mathcal L^*(L,W)-\mathcal L_\infty=c\,L^{-b_L}W^{-b_W}
\)
captures additional variation across architectures
(Figure~\ref{fig:model_size_law}B,C). At
approximately $170$M parameters, a deeper, narrower model attains a loss
$0.06$ nats per token below a shallower, wider model. Model size therefore
summarizes the main trend, with architecture accounting for some of the
remaining variation.

Larger models also reach their minimum validation loss earlier: the optimal epoch
count falls from $21$ to $6$ over the tested model-size range
(Figure~\ref{fig:datasize_sweep}F). By comparison, it varies from $14$ to $9$
across the data-size sweep at fixed model size. Over the tested ranges, optimal epoch counts decrease more rapidly with
model size than with corpus size. Appendix~\ref{app:scaling_details} collects the fitted exponents and compares
attainable loss along the model and data axes.

\begin{figure}[t]
    \centering
\includegraphics[height=0.327\linewidth]{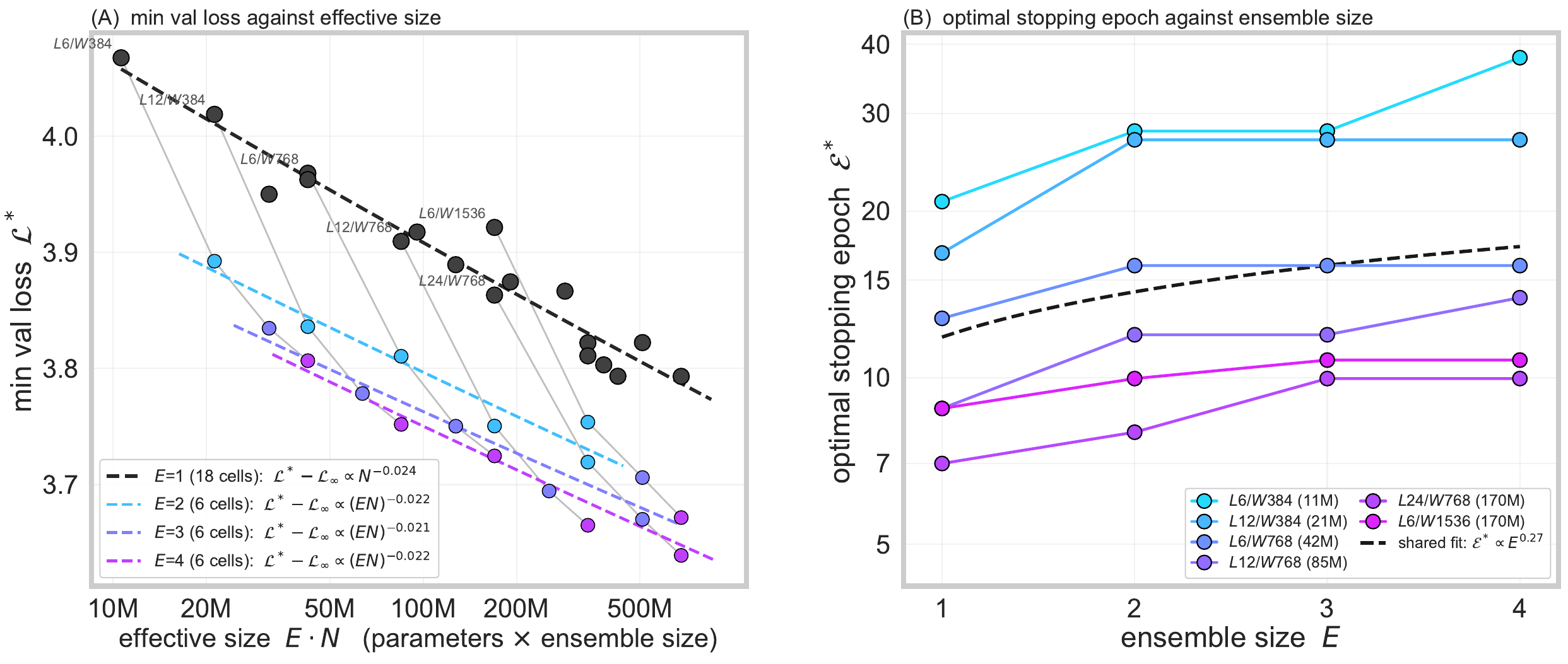}
\vspace{-4pt}
\caption{\textbf{Ensembles outperform larger single models and reach their minimum later.}
At $P=100$M: (A) minimum validation loss versus effective size $EN$;
black circles show single models, coloured points ensembles of six configurations.
(B) Optimal stopping epoch versus ensemble size. Dashed lines show fitted
power laws, with a shared slope in (B); details: Appendix~\ref{app:scaling_details}.}
\label{fig:ensemble_vs_size}
\vspace{-12pt}
\end{figure}

\noindent\textbf{The ensemble advantage persists across model sizes.}
We compare ensembles of six model configurations with larger single models,
holding the corpus and training horizon fixed. Matching effective size
$EN$ (total non-embedding parameters) therefore matches parameter-token
training budgets. Four-member ensembles achieve validation losses
$0.12$--$0.20$ nats lower than the corresponding wider or deeper single
models (Figure~\ref{fig:ensemble_vs_size}A). Ensembling also shifts the
minimum to later training epochs (Figure~\ref{fig:ensemble_vs_size}B).
This later stopping is qualitatively consistent with the RF mechanism in
Section~\ref{sec:mechanism}: averaging suppresses a growing error, allowing further training to remain beneficial.

\vspace{-10pt}
\section{Discussion and Conclusion}
\vspace{-5pt}

Our results show how repeated training epochs and ensembling jointly
shape generalization on a fixed dataset. The multi-epoch DMFT tracks
error correlations within and across epochs and between ensemble
members. For noiseless Gaussian RF gradient flow at \(N=P\), a
component of the expected test error grows at late training times.
Averaging the predictions of \(E\) models reduces this component by
a factor of \(E\), changing the balance that determines optimal
stopping. The pure-power continuum analysis translates the resulting
stopping laws into rules for allocating compute between ensemble size
and training duration, with data and member sizes fixed and each
ensemble trained to its own loss minimum.

Our language-model experiments show lower validation loss and later
optimal stopping with ensembling, qualitatively consistent with this
mechanism. They also show gains over wider or deeper single models
at matched parameter-token training budgets. A quantitative connection
to the RF theory requires accounting for feature learning and optimizer
dynamics, and relating RF feature and sample counts to Transformer
parameter and token counts.

Our label-noise analysis provides a starting point for extending
these stopping and allocation laws to noisy targets. Reshuffling
requires further analysis of how changing the data order affects
correlations across epochs. Together, the theory and experiments
motivate choosing the number of training epochs and ensemble members
jointly when allocating compute to a fixed dataset.

\bibliographystyle{plainnat}
\bibliography{ref}

\begin{thebibliography}{25}
\providecommand{\natexlab}[1]{#1}
\providecommand{\url}[1]{\texttt{#1}}
\expandafter\ifx\csname urlstyle\endcsname\relax
  \providecommand{\doi}[1]{doi: #1}\else
  \providecommand{\doi}{doi: \begingroup \urlstyle{rm}\Url}\fi

\bibitem[Adlam and Pennington(2020)]{adlam2020understanding}
Ben Adlam and Jeffrey Pennington.
\newblock Understanding double descent requires a fine-grained bias-variance
  decomposition.
\newblock \emph{Advances in neural information processing systems},
  33:\penalty0 11022--11032, 2020.

\bibitem[Bordelon and Pehlevan(2026)]{bordelon2026disordered}
Blake Bordelon and Cengiz Pehlevan.
\newblock Disordered dynamics in high dimensions: Connections to random
  matrices and machine learning.
\newblock \emph{arXiv preprint arXiv:2601.01010}, 2026.

\bibitem[Bordelon et~al.(2020)Bordelon, Canatar, and
  Pehlevan]{bordelon2020spectrum}
Blake Bordelon, Abdulkadir Canatar, and Cengiz Pehlevan.
\newblock Spectrum dependent learning curves in kernel regression and wide
  neural networks.
\newblock In \emph{International Conference on Machine Learning}, pages
  1024--1034. PMLR, 2020.

\bibitem[Bordelon et~al.(2024)Bordelon, Atanasov, and
  Pehlevan]{bordelon2024dynamical}
Blake Bordelon, Alexander Atanasov, and Cengiz Pehlevan.
\newblock A dynamical model of neural scaling laws.
\newblock In \emph{International Conference on Machine Learning}, pages
  4345--4382. PMLR, 2024.

\bibitem[Canatar et~al.(2021)Canatar, Bordelon, and
  Pehlevan]{canatar2021spectral}
Abdulkadir Canatar, Blake Bordelon, and Cengiz Pehlevan.
\newblock Spectral bias and task-model alignment explain generalization in
  kernel regression and infinitely wide neural networks.
\newblock \emph{Nature communications}, 12\penalty0 (1):\penalty0 2914, 2021.

\bibitem[Crisanti and Sompolinsky(2018)]{crisanti2018path}
A~Crisanti and H~Sompolinsky.
\newblock Path integral approach to random neural networks.
\newblock \emph{Physical Review E}, 98\penalty0 (6):\penalty0 062120, 2018.

\bibitem[Defilippis et~al.(2024)Defilippis, Loureiro, and
  Misiakiewicz]{defilippis2024dimension}
Leonardo Defilippis, Bruno Loureiro, and Theodor Misiakiewicz.
\newblock Dimension-free deterministic equivalents and scaling laws for random
  feature regression.
\newblock \emph{Advances in Neural Information Processing Systems},
  37:\penalty0 104630--104693, 2024.

\bibitem[Geiger et~al.(2020)Geiger, Jacot, Spigler, Gabriel, Sagun, d’Ascoli,
  Biroli, Hongler, and Wyart]{geiger2020scaling}
Mario Geiger, Arthur Jacot, Stefano Spigler, Franck Gabriel, Levent Sagun,
  St{\'e}phane d’Ascoli, Giulio Biroli, Cl{\'e}ment Hongler, and Matthieu
  Wyart.
\newblock Scaling description of generalization with number of parameters in
  deep learning.
\newblock \emph{Journal of Statistical Mechanics: Theory and Experiment},
  2020\penalty0 (2):\penalty0 023401, 2020.

\bibitem[Gerbelot et~al.(2022)Gerbelot, Troiani, Mignacco, Krzakala, and
  Zdeborova]{gerbelot2022rigorous}
Cedric Gerbelot, Emanuele Troiani, Francesca Mignacco, Florent Krzakala, and
  Lenka Zdeborova.
\newblock Rigorous dynamical mean field theory for stochastic gradient descent
  methods.
\newblock \emph{arXiv preprint arXiv:2210.06591}, 2022.

\bibitem[Helias and Dahmen(2020)]{helias2020statistical}
Moritz Helias and David Dahmen.
\newblock \emph{Statistical field theory for neural networks}, volume 970.
\newblock Springer, 2020.

\bibitem[Hoffmann et~al.(2022)Hoffmann, Borgeaud, Mensch, Buchatskaya, Cai,
  Rutherford, Casas, Hendricks, Welbl, Clark, et~al.]{hoffmann2022training}
Jordan Hoffmann, Sebastian Borgeaud, Arthur Mensch, Elena Buchatskaya, Trevor
  Cai, Eliza Rutherford, Diego de~Las Casas, Lisa~Anne Hendricks, Johannes
  Welbl, Aidan Clark, et~al.
\newblock Training compute-optimal large language models.
\newblock \emph{arXiv preprint arXiv:2203.15556}, 2022.

\bibitem[Kaplan et~al.(2020)Kaplan, McCandlish, Henighan, Brown, Chess, Child,
  Gray, Radford, Wu, and Amodei]{kaplan2020scaling}
Jared Kaplan, Sam McCandlish, Tom Henighan, Tom~B Brown, Benjamin Chess, Rewon
  Child, Scott Gray, Alec Radford, Jeffrey Wu, and Dario Amodei.
\newblock Scaling laws for neural language models.
\newblock \emph{arXiv preprint arXiv:2001.08361}, 2020.

\bibitem[Kim et~al.(2025)Kim, Kotha, Liang, and Hashimoto]{kim2025pre}
Konwoo Kim, Suhas Kotha, Percy Liang, and Tatsunori Hashimoto.
\newblock Pre-training under infinite compute.
\newblock \emph{arXiv preprint arXiv:2509.14786}, 2025.

\bibitem[Lin et~al.(2024)Lin, Wu, Kakade, Bartlett, and Lee]{lin2024scaling}
Licong Lin, Jingfeng Wu, Sham~M Kakade, Peter~L Bartlett, and Jason~D Lee.
\newblock Scaling laws in linear regression: Compute, parameters, and data.
\newblock \emph{arXiv preprint arXiv:2406.08466}, 2024.

\bibitem[Maloney et~al.(2022)Maloney, Roberts, and Sully]{maloney2022solvable}
Alexander Maloney, Daniel~A Roberts, and James Sully.
\newblock A solvable model of neural scaling laws.
\newblock \emph{arXiv preprint arXiv:2210.16859}, 2022.

\bibitem[Mandal et~al.(2026)Mandal, Berman, Vegesna, and Dahal]{mandal2026q0}
Bishwas Mandal, Shmuel Berman, Akshay Vegesna, and Samip Dahal.
\newblock {q0}: Primitives for hyper-epoch pretraining.
\newblock \emph{arXiv preprint arXiv:2606.03938}, 2026.
\newblock \doi{10.48550/arXiv.2606.03938}.
\newblock URL \url{https://arxiv.org/abs/2606.03938}.

\bibitem[Mignacco et~al.(2020)Mignacco, Krzakala, Urbani, and
  Zdeborov{\'a}]{mignacco2020dynamical}
Francesca Mignacco, Florent Krzakala, Pierfrancesco Urbani, and Lenka
  Zdeborov{\'a}.
\newblock Dynamical mean-field theory for stochastic gradient descent in
  gaussian mixture classification.
\newblock \emph{Advances in Neural Information Processing Systems},
  33:\penalty0 9540--9550, 2020.

\bibitem[Muennighoff et~al.(2023)Muennighoff, Rush, Barak, Le~Scao, Tazi,
  Piktus, Pyysalo, Wolf, and Raffel]{muennighoff2023scaling}
Niklas Muennighoff, Alexander Rush, Boaz Barak, Teven Le~Scao, Nouamane Tazi,
  Aleksandra Piktus, Sampo Pyysalo, Thomas Wolf, and Colin~A Raffel.
\newblock Scaling data-constrained language models.
\newblock \emph{Advances in Neural Information Processing Systems},
  36:\penalty0 50358--50376, 2023.

\bibitem[Paquette et~al.(2021)Paquette, Lee, Pedregosa, and
  Paquette]{paquette2021sgd}
Courtney Paquette, Kiwon Lee, Fabian Pedregosa, and Elliot Paquette.
\newblock Sgd in the large: Average-case analysis, asymptotics, and stepsize
  criticality.
\newblock In \emph{Conference on Learning Theory}, pages 3548--3626. PMLR,
  2021.

\bibitem[Paquette et~al.(2024)Paquette, Paquette, Xiao, and
  Pennington]{paquette20244+}
Elliot Paquette, Courtney Paquette, Lechao Xiao, and Jeffrey Pennington.
\newblock 4+ 3 phases of compute-optimal neural scaling laws.
\newblock \emph{arXiv preprint arXiv:2405.15074}, 2024.

\bibitem[Ruben and Pehlevan(2023)]{ruben2023learning}
Ben Ruben and Cengiz Pehlevan.
\newblock Learning curves for noisy heterogeneous feature-subsampled ridge
  ensembles.
\newblock \emph{Advances in Neural Information Processing Systems},
  36:\penalty0 50041--50093, 2023.

\bibitem[Ruben et~al.(2025)Ruben, Tong, Chaudhry, and Pehlevan]{ruben2025no}
Benjamin~Samuel Ruben, William~Lingxiao Tong, Hamza~Tahir Chaudhry, and Cengiz
  Pehlevan.
\newblock No free lunch from random feature ensembles: Scaling laws and
  near-optimality conditions.
\newblock In \emph{International Conference on Machine Learning}, pages
  52198--52224. PMLR, 2025.

\bibitem[Simon et~al.(2021)Simon, Dickens, Karkada, and
  DeWeese]{simon2021eigenlearning}
James~B Simon, Madeline Dickens, Dhruva Karkada, and Michael~R DeWeese.
\newblock The eigenlearning framework: A conservation law perspective on kernel
  regression and wide neural networks.
\newblock \emph{arXiv preprint arXiv:2110.03922}, 2021.

\bibitem[Villalobos et~al.(2024)Villalobos, Ho, Sevilla, Besiroglu, Heim, and
  Hobbhahn]{villalobos2024run}
Pablo Villalobos, Anson Ho, Jaime Sevilla, Tamay Besiroglu, Lennart Heim, and
  Marius Hobbhahn.
\newblock Will we run out of data? limits of llm scaling based on
  human-generated data, 2024.

\bibitem[Yan et~al.(2025)Yan, Wen, Li, Luo, Chen, and Lyu]{yan2025larger}
Tingkai Yan, Haodong Wen, Binghui Li, Kairong Luo, Wenguang Chen, and Kaifeng
  Lyu.
\newblock Larger datasets can be repeated more: A theoretical analysis of
  multi-epoch scaling in linear regression.
\newblock \emph{arXiv preprint arXiv:2511.13421}, 2025.

\end{thebibliography}

\clearpage
\appendix

\section{Experimental Protocol and Additional Fits}
\label{app:experimental_protocol}

This appendix collects the experiment designs, training settings, measurement
conventions, and fit results supporting Section~\ref{sec:phenomena} and the
language-model experiments. Figures and numerical results retain their
reported training recipes.

\subsection{Experiment designs and training settings}

\begin{table}[ht]
\centering
\caption{Language-model experiments. All listed sweeps use constant learning
rates; $\lambda$ is the AdamW weight-decay coefficient.}
\label{tab:experiment_design}
\begin{tabular}{@{}>{\raggedright\arraybackslash}p{0.22\linewidth}>{\raggedright\arraybackslash}p{0.38\linewidth}>{\raggedright\arraybackslash}p{0.33\linewidth}@{}}
\toprule
Experiment & Models and data & Horizon and replication \\
\midrule
Data-size sweep\newline Fig.~\ref{fig:datasize_sweep}, top &
$L=12$, $W=768$ (85M); $P\in\{10,20,\ldots,100\}$M; $\lambda=0$ &
About 8,000 updates ($1.0$--$1.1$B tokens), or 11--101 epochs; one run per strategy at each $P$ \\
\addlinespace
Model-size grid\newline Figs.~\ref{fig:datasize_sweep}, bottom, \ref{fig:model_size_law}, \ref{fig:optimal_stopping} &
$L\in\{6,12,18,24\}\times W\in\{384,768,1152,1536\}$, plus $L\in\{48,60\}$ at $W=768$; $P=100$M; $\lambda=0$ &
40 epochs; one run per cell; 18 cells spanning 11M--680M parameters \\
\addlinespace
Matched-budget comparison\newline Figs.~\ref{fig:ens_val_vs_t}, \ref{fig:app_compute_matched} &
Base $L=12$, $W=768$; ensembles versus wider/deeper single models; $P=100$M; $\lambda=0$ &
40-epoch training budget; ensembles in Fig.~\ref{fig:ens_val_vs_t} evaluated from saved model states through epoch 31 \\
\addlinespace
Ensemble-size scaling\newline Fig.~\ref{fig:ensemble_scaling} &
$L=12$, $W=768$; $P=20$M; $\lambda=0$; shared or independent data orders &
40 epochs; 20 trained members per strategy; 10 random member orderings per strategy \\
\addlinespace
Ensembles across model sizes\newline Fig.~\ref{fig:ensemble_vs_size} &
Six replicated cells (11M--170M); $P=100$M; $\lambda=0$; $E=2,3,4$ &
40 epochs; at least four independent members per cell; up to 680M total counted parameters \\
\addlinespace
Additional joint-fit data &
Seven-cell model-size ladder at $P=20$M; $\lambda=0$ &
50 epochs; one run per configuration \\
\bottomrule
\end{tabular}
\end{table}

\begin{table}[h]
\centering
\caption{Language-model training configuration.}
\label{tab:protocol}
\begin{tabular}{@{}lp{0.74\linewidth}@{}}
\toprule
Data & FineWeb; GPT-2 BPE tokenizer (vocabulary 50{,}257); 10M held-out validation tokens \\
Architecture & Decoder-only Transformer, depth $L$, width $W$, head dimension 64 ($W/64$ heads); rotary embeddings, QK-norm, RMSNorm without weights, no biases, no dropout; SwiGLU MLP with hidden width $8W/3$; untied input and output embeddings; learned residual, input-injection, and U-Net skip gains; sliding-window attention on three of every four layers \\
Parameterization & CompleteP with base $L{=}12$, $W{=}768$; truncated-normal initialization, std 0.02 at base width \\
Sequence length & 2048 tokens \\
Batch size & 131{,}072 tokens (64 sequences) per update \\
Optimizer & AdamW, $\beta=(0.8,0.95)$, $\epsilon=10^{-10}$ \\
Learning rates & matrices 0.01, embeddings 0.0375, output head 0.0005, scalar gains 0.025 \\
Schedule & constant, no warmup ($\lambda{=}0$ runs); last 20\% linear decay to zero (runs with $\lambda>0$ and cooldown) \\
Precision, hardware & bfloat16 autocast; one NVIDIA H100-80GB per run \\
Seeds & one run per configuration; ensemble members differ in initialization and, where stated, data order \\
\bottomrule
\end{tabular}
\end{table}

\FloatBarrier
\subsection{Evaluation, ensembling, and compute accounting}

\begin{table}[ht]
\centering
\caption{Measurement and selection conventions.}
\label{tab:measurement_conventions}
\begin{tabular}{@{}>{\raggedright\arraybackslash}p{0.24\linewidth}>{\raggedright\arraybackslash}p{0.71\linewidth}@{}}
\toprule
Quantity & Definition or procedure \\
\midrule
Validation loss & Cross-entropy in nats per token on a held-out FineWeb split. \\
\addlinespace
Epoch and step & Each update processes 131,072 tokens; one epoch is $P/131{,}072$ updates (763 at $P=100$M; 152 at $P=20$M). \\
\addlinespace
Evaluation cadence & Every 76 updates in the data sweep and 152 updates in the model grids, corresponding to about 10M and 19.9M training tokens. \\
\addlinespace
Saved model state & A checkpoint stores the model parameters at a training step. The optimal checkpoint is the evaluated state with minimum validation loss. \\
\addlinespace
Optimal stopping & $s^*$ is the evaluated step with minimum validation loss. $\mathcal E^*=s^*/(P/131{,}072)$ is the corresponding epoch count. \\
\addlinespace
Data-sweep aggregation & One minimum per strategy at each corpus size; the reported minimum loss and optimal epoch average the two strategies' minima. \\
\addlinespace
Ensemble prediction & Average member logits before computing validation cross-entropy. Initialization-only members share the data order; initialization-plus-shuffle members vary both. \\
\bottomrule
\end{tabular}
\end{table}

\paragraph{Ensemble fits and resampling.}\label{app:ensemble_protocol}
At optimizer step $s$, each data-order strategy is fitted separately with
\begin{equation}
\mathcal L_{s,E}=\alpha_s E^{-\beta}+\mathcal L_{s,\infty},
\label{eq:main_ensemble_fit}
\end{equation}
with one exponent $\beta$ per strategy and an amplitude $\alpha_s$ and asymptote $\mathcal L_{s,\infty}$ at each step.
The asymptote is strategy-specific because initialization-only members share one data order, which their infinite ensemble retains.
Each of 10 resampling iterations draws a random ordering of the 20 trained members of each strategy; the ensemble of size $E$ consists of the first $E$ members, and \eqref{eq:main_ensemble_fit} is refitted per iteration. Error bars show the spread over iterations.

\begin{figure}[t]
    \centering
    \includegraphics[width=\linewidth]{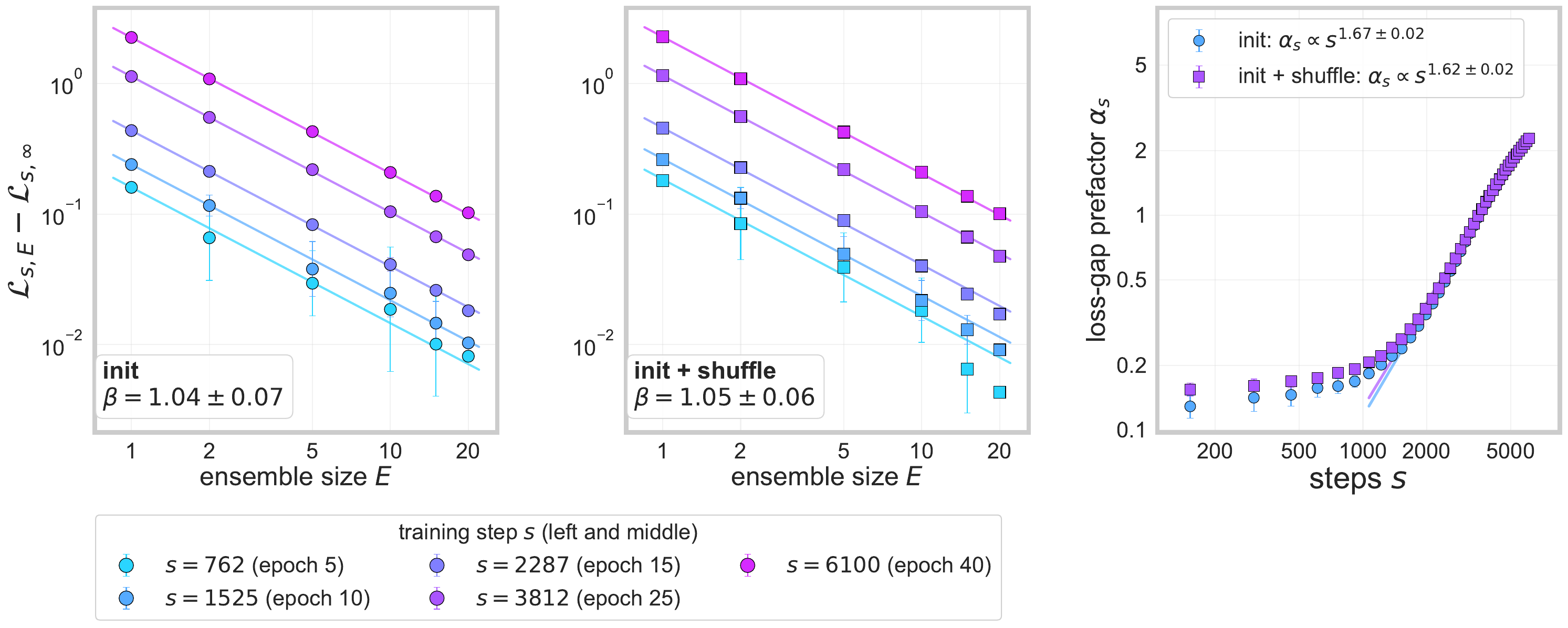}
\caption{\textbf{The reducible ensemble loss decays approximately as $E^{-1}$ under both sources of member diversity.} Left and middle: loss above the fitted infinite-ensemble asymptote against ensemble size, for members with shared or independently shuffled data orders; colour marks the training step and lines are fits of \eqref{eq:main_ensemble_fit}. Right: fitted amplitude $\alpha_s$ against step. $P=20$M, 40 epochs.}
    \label{fig:ensemble_scaling}
\end{figure}

The fitted ensemble-size exponents are close to one
(Figure~\ref{fig:ensemble_scaling}, left and middle), consistent with the
$E^{-1}$ dependence predicted by the fixed-checkpoint cross-entropy
expansion (Appendix~\ref{app:logit_ensemble_ce}).
The amplitude $\alpha_s$ is nearly constant over the first $1{,}000$ steps and grows about tenfold by epoch 40 (right).

\paragraph{Training-budget proxy.}\label{app:compute_matching}
We count the attention and MLP matrices as $N=12LW^2$: each layer has four
$W\times W$ attention matrices and a gated MLP with hidden width $8W/3$
(three matrices totalling $8W^2$). Token embeddings, the output head, and
per-layer scalar gains are excluded. With $D$ training tokens per member,
the compute proxy is $C=END$. For the matched-budget experiments,
$D=40P$: we charge the full 40-epoch run and report the best evaluated
checkpoint within the available evaluation horizon.
Thus four base models and one twice-wider model have the same proxy cost
($4\times85$M $=340$M counted parameters at the same $D$).
The fixed-horizon LM comparison differs from the RF allocation, which
charges training up to each ensemble's own optimal stopping time.
The weight-decay coefficient $\lambda$ is unrelated to the feature-spectrum
eigenvalues $\lambda_k$ in Section~\ref{sec:dmft}.

\FloatBarrier
\subsection{Loss laws and fitted exponents}\label{app:scaling_details}

\paragraph{Shared loss floor.}\label{app:asymptotes}
Loss laws in $N$, $L,W$, and $P$ use the shared floor
$\mathcal L_\infty=1.25$. Table~\ref{tab:fit_conventions} gives its joint-fit
source and the observations used in each subsequent fit.

\begin{table}[ht]
\centering
\caption{Fitting conventions for attainable loss and optimal stopping.}
\label{tab:fit_conventions}
\begin{tabular}{@{}>{\raggedright\arraybackslash}p{0.23\linewidth}>{\raggedright\arraybackslash}p{0.72\linewidth}@{}}
\toprule
Fit & Model and observations \\
\midrule
Joint $N,P$ law & $\mathcal L^*=\mathcal L_\infty+A N^{-\alpha}+B P^{-\alpha}$ on 35 observations: the 18-cell $100$M grid, seven-cell $20$M ladder, and ten data-sweep settings. Shared exponent $\alpha=0.19$, floor $1.25$, $R^2=0.996$. \\
\addlinespace
Model-size laws & One minimum per cell on the 18-cell $100$M grid. Fit $\mathcal L^*-\mathcal L_\infty$ against $N$, or against $L,W$ with separate exponents. Optimal-epoch fits use the same cells. \\
\addlinespace
Data-size laws & Ten corpus sizes at fixed $L=12,W=768$, using the strategy-averaged minima and optimal epochs defined in Table~\ref{tab:measurement_conventions}. \\
\addlinespace
Ensemble/model-size laws & A separate power law in total counted parameters $EN$ for each $E=2,3,4$, using the six replicated cells and the same loss floor. \\
\addlinespace
Ensemble stopping & Fit $\log\mathcal E_j^*=\log c_j+\gamma\log E$ over $E=1,2,3,4$ for six model configurations $j$, with a shared exponent $\gamma=0.27$ and separate prefactors $c_j$. The dashed guide in Figure~\ref{fig:ensemble_vs_size}B uses the geometric mean of these prefactors. \\
\addlinespace
Ensemble-size law & Equation~\eqref{eq:main_ensemble_fit} fits an infinite-ensemble asymptote for each strategy at each checkpoint; this is distinct from the shared $N,P$ loss floor. \\
\bottomrule
\end{tabular}
\end{table}

\begin{table}[ht]
\centering
\caption{Fitted exponents for optimal epochs $\mathcal E^*$ and minimum loss
above the shared floor, $\Delta\mathcal L^*=\mathcal L^*-1.25$.
Table~\ref{tab:fit_conventions} specifies the observations and fitting conventions.}
\label{tab:stopping_exponents}
\setlength{\tabcolsep}{4pt}
\begin{tabular}{@{}llccc@{}}
\toprule
Quantity & Fitted dependence & Exponent(s) & $R^2$ & Points \\
\midrule
$\mathcal E^*$ & $N^{-\alpha}$ & $0.314$ & $0.938$ & 18 \\
$\mathcal E^*$ & $L^{-a_L}W^{-a_W}$ & $a_L=0.336,\ a_W=0.608$ & $0.940$ & 18 \\
$\mathcal E^*$ & $P^{-\alpha}$ & $0.158$ & $0.917$ & 10 \\
\midrule
$\Delta\mathcal L^*$ & $N^{-\beta_N}$ & $0.024$ & $0.943$ & 18 \\
$\Delta\mathcal L^*$ & $L^{-b_L}W^{-b_W}$ & $b_L=0.0304,\ b_W=0.0436$ & $0.969$ & 18 \\
$\Delta\mathcal L^*$ & $P^{-\beta_P}$ & $0.169$ & $0.997$ & 10 \\
\bottomrule
\end{tabular}
\end{table}

\begin{table}[ht]
\centering
\caption{Additional fit diagnostics and quantitative comparisons.}
\label{tab:fit_diagnostics}
\begin{tabular}{@{}>{\raggedright\arraybackslash}p{0.25\linewidth}>{\raggedright\arraybackslash}p{0.70\linewidth}@{}}
\toprule
Comparison & Result on the measured configurations \\
\midrule
Depth/width loss fit & $b_L=0.0304$, $b_W=0.0436$, each stable to $\pm0.002$ under leave-one-out. The fitted sensitivity ratio is $b_W/(2b_L)=0.72$. \\
\addlinespace
Matched-size example & At $N\approx170$M, $L=6,W=1536$ attains loss $3.92$, versus $3.86$ for $L=24,W=768$. \\
\addlinespace
Depth/width stopping fit & $a_L=0.336$, $a_W=0.608$, so $a_W/(2a_L)=0.90$. Its rounded $R^2=0.94$ matches the fit using $N$ alone. Dependence solely on $N\propto LW^2$ would give $a_W=2a_L$. \\
\addlinespace
Resource scaling & Optimal epochs fall by a fitted factor of $1.4$ per decade of data scaling, versus $2.1$ per decade of model scaling. \\
\addlinespace
Ensemble/model-size fits & Exponents $0.022$, $0.021$, $0.022$ for $E=2,3,4$, versus $0.024$ for single models. The $E=4$ fitted law lies about $0.15$ nats below the single-model law over the comparison range. \\
\bottomrule
\end{tabular}
\end{table}

\begin{figure}[t]
    \centering
\includegraphics[width=\linewidth]{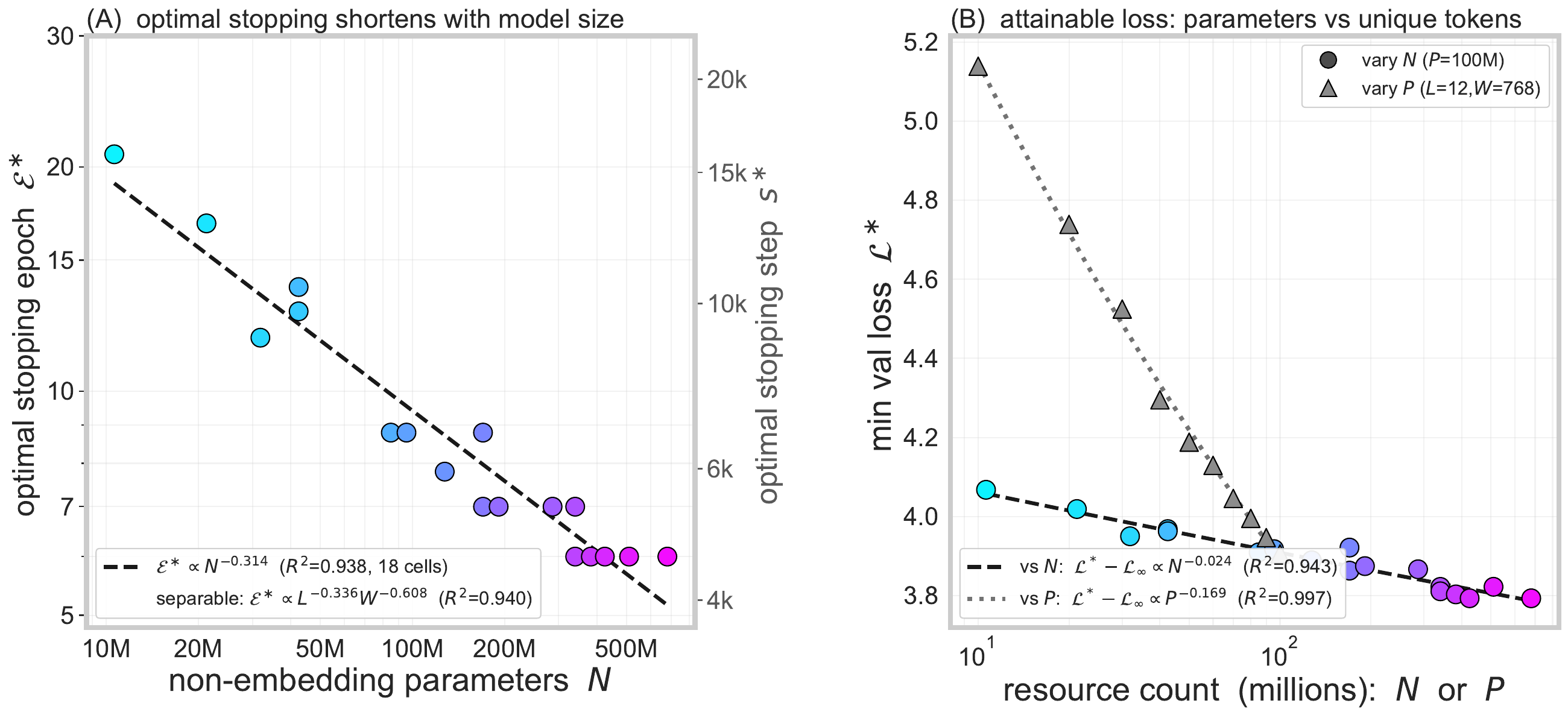}
\caption{\textbf{Model size changes both the best stopping point and the
attainable loss.} (A) Optimal stopping versus model size on the $100$M-token
corpus, shown in epochs and optimizer steps. (B) Minimum validation loss
when varying model size at fixed data (circles) or data size at fixed model
(triangles). Dashed curves show power-law fits over the respective measured
ranges. The loss fits in (B) use the shared floor from
Appendix~\ref{app:asymptotes}. Colour denotes model size.}
\label{fig:optimal_stopping}
\end{figure}

Figure~\ref{fig:optimal_stopping} collects the model-size stopping curve
and the attainable-loss fits along the model and data axes.

\FloatBarrier
\subsection{Additional matched-budget comparisons}
\begin{figure}[H]
    \centering
    \includegraphics[width=0.99\linewidth]{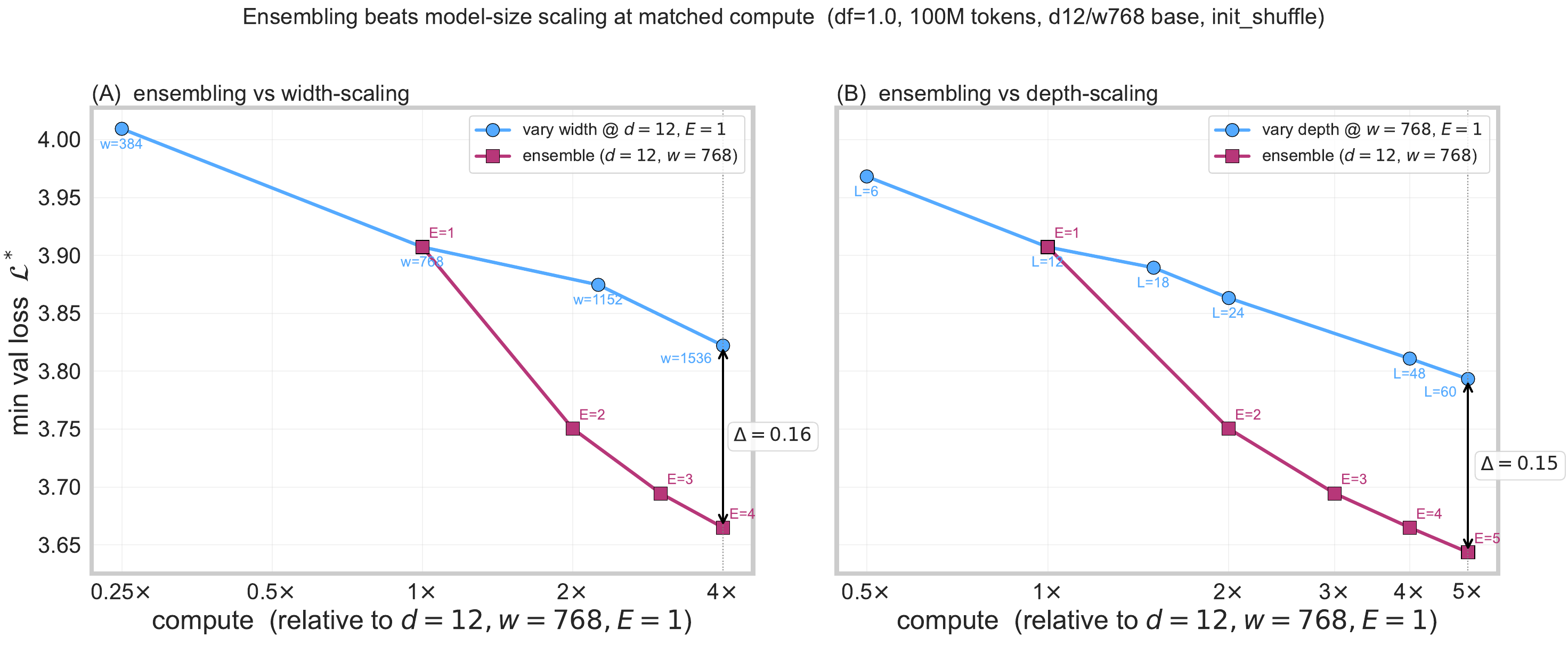}
    \caption{Ensembles compared with wider (left) or deeper (right) single
models at matched parameter-token training budgets. Both comparisons use
$P=100$M, constant learning rates, zero weight decay, and a 40-epoch budget.}
    \label{fig:app_compute_matched}
\end{figure}

\section{Derivation of the Multi-Epoch DMFT Equations}
\label{app:derivations}

We derive the dynamical mean-field theory (DMFT) for fixed-order multi-epoch
training with independent Gaussian population features
\(\bm\psi(\bm x_\mu)\sim\mathcal N(\bm0,\bm\Lambda)\).
The population features, Gaussian random-feature matrix, and label noises
are mutually independent. Feature-space quantities are written in the
eigenbasis of the population covariance,
\[
    \bm \Lambda=\diag(\lambda_1,\lambda_2,\ldots),
    \qquad
    \left\langle \psi_k(\bm x)\psi_\ell(\bm x)\right\rangle
    =
    \lambda_k\delta_{k\ell}.
\]
The target has coefficients \(w_k^\star\), so that
\[
    y(\bm x)
    =
    \bm\psi(\bm x) \cdot \bm w^\star 
    +
    \sigma\epsilon.
\]
The DMFT equations allow an arbitrary spectrum and target. We impose power laws
only in Appendix~\ref{app:scaling_form_interpolation_growth}.

We first state the finite-horizon single-site equations and derive them by
cavity and MSR methods. Later sections treat finite-size error, ensembles,
continuum limits, and the low-frequency scaling near interpolation.

\subsection{Asymptotic settings}
\label{app:asymptotic_settings}

We first derive the closure at finite feature dimension \(M\) and trajectory
length \(S=T\mathcal E\). The population features enter through
\(\bm\Lambda\) and the target coefficients. The closure retains
finite-width and finite-minibatch contributions through its \(N^{-1}\)
and \(B^{-1}\) terms. The loss approximation in
Theorem~\ref{thm:app_finite_horizon_error} keeps \(M,S\) fixed as \(N,B\)
grow. The spectral, small-learning-rate, and long-time limits used for
scaling are specified separately below. We distinguish two settings for
the population spectrum.

\paragraph{Dimension-free power-law setting.}
For the scaling analysis, we let \(M\to\infty\) along a trace-class
population spectrum and assume
\begin{equation}
    \sum_{k=1}^\infty\lambda_k<\infty,
    \qquad
    \sum_{k=1}^\infty
    \lambda_k(w_k^\star)^2<\infty.
    \label{eq:app_trace_class_setting}
\end{equation}
This is the dimension-free setting in which the power laws
\(\lambda_k\sim k^{-b}\) and
\(\lambda_k(w_k^\star)^2\sim k^{-a}\), with \(a,b>1\), are imposed in
Appendix~\ref{app:scaling_form_interpolation_growth}.

\paragraph{Finite-dimensional isotropic specialization.}
For comparison, we also consider a proportional sequence with
\begin{equation}
    M,N,P\to\infty,
    \qquad
    \frac{N}{M}\to\phi\in(0,\infty),
    \qquad
    \frac{P}{M}\to\gamma\in(0,\infty),
    \qquad
    \bm\Lambda_M=\bm I_M.
    \label{eq:app_isotropic_proportional_setting}
\end{equation}
We choose a sequence of targets satisfying
\begin{equation}
    \|\bm w_M^\star\|_2^2\to\rho\in(0,\infty).
    \label{eq:app_isotropic_target_scaling}
\end{equation}
Equivalently, an isotropically oriented target may be generated with
coordinates of order \(M^{-1/2}\), for example
\(\bm w_M^\star=\sqrt{\rho/M}\,\bm g\) with
\(\bm g\sim\mathcal N(\bm0,\bm I_M)\). Since
\[
    \mathbb E_{\bm x}
    \bigl[(\bm w_M^\star\cdot\bm\psi(\bm x))^2\bigr]
    =
    \bm w_M^{\star\top}\bm\Lambda_M\bm w_M^\star
    =
    \|\bm w_M^\star\|_2^2,
\]
this scaling keeps the target signal and loss of order one. This proportional
isotropic sequence is not an instance of the infinite-dimensional trace-class
limit, since \(\operatorname{Tr}(\bm\Lambda_M)=M\).

We use a combined epoch-step index
\[
    \alpha=(n,t),
    \qquad
    n=0,\ldots,\mathcal E-1,
    \qquad
    t=0,\ldots,T-1,
\]
and let \(S=T\mathcal E\) be the total number of update times. Greek indices
\(\alpha,\beta,\gamma,\ldots\) always denote epoch-step indices. We write sums
over epoch-step indices explicitly unless matrix notation is used.
The global update index is \(s=nT+t\), and effective training time is
\(\tau=\eta s\), as in the main text.
Whenever an inequality such as \(\alpha\leq S\) is used, the pairs are ordered
causally through the one-based index \(s+1\in\{1,\ldots,S\}\).

Define the causal matrix \(\bm\Theta\) by
\begin{equation}
    \Theta_{\alpha\beta}
    =
    \Theta_{(n,t),(n',t')}
    =
    \delta_{nn'}\mathbf 1\{t>t'\}
    +
    \mathbf 1\{n>n'\}.
    \label{eq:app_theta_indexed}
\end{equation}
Thus \(\Theta_{\alpha\beta}=1\) when update \(\beta\) precedes state \(\alpha\).
It sums the past gradient updates in the residual recursion.

For any \(S\times S\) matrix \(\bm A\), define \(\mathcal D\) by
\begin{equation}
    [\mathcal D(\bm A)]_{(n,t),(n',t')}
    =
    A_{(n,t),(n',t)}\delta_{tt'}.
    \label{eq:app_D_indexed}
\end{equation}
The map \(\mathcal D\) retains pairs with the same minibatch label \(t\) and
sets all other entries to zero. It appears in covariance and response terms
generated by reusing \(\bm\Psi(t)\) in each epoch.

\subsection{Decomposing the dynamics}
\label{app:decomposing_dynamics}

We begin from the feature-space residual
\begin{equation}
    \bm v^0_n(t)
    =
    \bm w^\star
    -
    \frac1N
    \bm A^\top\bm w_n(t)
    \in\mathbb R^M ,
    \label{eq:app_v0_def_bold}
\end{equation}
where \(\bm A\in\mathbb R^{N\times M}\) is the random feature matrix,
\(\bm w_n(t)\in\mathbb R^N\) is the trainable weight vector, and
\(\bm w^\star\in\mathbb R^M\) is the target coefficient vector. The minibatch
feature matrix at within-epoch step \(t\) is
\[
    \bm\Psi(t)\in\mathbb R^{B\times M}.
\]
In the fixed-order multi-epoch setting, the same \(\bm\Psi(t)\) is reused at
within-epoch step \(t\) of every epoch.

The feature-space residual update, including label noise, is
\begin{equation}
    \bm v^0_n(t+1)
    =
    \bm v^0_n(t)
    -
    \eta
    \left(
        \frac1N\bm A^\top\bm A
    \right)
    \frac1B
    \bm\Psi(t)^\top
    \left(
        \bm\Psi(t)\bm v^0_n(t)
        +
        \sigma\bm\epsilon(t)
    \right),
    \label{eq:app_residual_update_bold}
\end{equation}
with epoch boundary condition
\begin{equation}
    \bm v^0_{n+1}(0)
    =
    \bm v^0_n(T).
    \label{eq:app_epoch_boundary_bold}
\end{equation}
Here \(\bm\epsilon(t)\sim\mathcal N(\bm0,\bm I_B)\) is drawn once with
minibatch \(t\), reused at the same within-epoch step of every epoch, and is
independent across different minibatch labels. Under the normalization
\(f(\bm x)=N^{-1}\bm w^\top\bm A\psi(\bm x)\), the learning rate \(\eta\) in
\eqref{eq:app_residual_update_bold} denotes the \(N\)-rescaled learning rate;
equivalently, the parameter-space learning rate for \(\bm w\) is \(\eta N\).
The noiseless case is recovered by setting \(\sigma=0\).

Following the standard random-feature construction
\cite{bordelon2024dynamical,bordelon2026disordered,paquette20244+}, introduce
four auxiliary fields. Each equation below contains only one of the random
matrices \(\bm\Psi(t)\) and \(\bm A\):
\begin{align}
    \bm v^1_n(t)
    &=
    \bm\Psi(t)\bm v^0_n(t)
    +
    \sigma\bm\epsilon(t)
    \in\mathbb R^B,
    \label{eq:app_v1_bold}
    \\
    \bm v^2_n(t)
    &=
    \frac1B
    \bm\Psi(t)^\top\bm v^1_n(t)
    \in\mathbb R^M,
    \label{eq:app_v2_bold}
    \\
    \bm v^3_n(t)
    &=
    \bm A\bm v^2_n(t)
    \in\mathbb R^N,
    \label{eq:app_v3_bold}
    \\
    \bm v^4_n(t)
    &=
    \frac1N
    \bm A^\top\bm v^3_n(t)
    \in\mathbb R^M.
    \label{eq:app_v4_bold}
\end{align}
The entries of \(\bm v^1_n(t)\) are the residuals on minibatch \(t\), including
label noise. Mapping them back to feature space gives
\[
    \bm v^2_n(t)
    =
    \frac1B
    \sum_{\mu\in\mathrm{batch}(t)}
    \psi(\bm x_\mu)v^1_{\mu,n}(t).
\]
so \(\bm v^2_n(t)\) is the empirical gradient with respect to the feature
coefficient. The trained parameter is \(\bm w_n(t)\in\mathbb R^N\), and
\(\bm v^3_n(t)=\bm A\bm v^2_n(t)\) is its update direction. The induced change
in the feature coefficient is
\[
    \bm v^4_n(t)
    =
    \frac1N\bm A^\top\bm v^3_n(t)
    =
    \left(\frac1N\bm A^\top\bm A\right)\bm v^2_n(t)
\]
Thus \(\bm v^4_n(t)\) is \(\bm v^2_n(t)\) filtered by the random-feature Gram
matrix \(\bm A^\top\bm A/N\), and it is the direction that enters the residual
update.

Equations~\eqref{eq:app_v1_bold}--\eqref{eq:app_v4_bold} reduce
\eqref{eq:app_residual_update_bold} to
\begin{equation}
    \bm v^0_n(t+1)
    =
    \bm v^0_n(t)
    -
    \eta\bm v^4_n(t),
    \qquad
    \bm v^0_{n+1}(0)
    =
    \bm v^0_n(T).
    \label{eq:app_residual_update_v4_bold}
\end{equation}
The label-noise contribution is contained in \(\bm v^4_n(t)\) through
\(\bm v^1_n(t)\).

Define the initial residual
\[
    \bm v^0_{\mathrm{init}}
    =
    \bm w^\star
    -
    \frac1N
    \bm A^\top\bm w_0(0).
\]
The full trajectory is
\begin{equation}
    \bm v^0_\alpha
    =
    \bm v^0_{\mathrm{init}}
    -
    \eta
    \sum_{\beta}
    \Theta_{\alpha\beta}
    \bm v^4_\beta .
    \label{eq:app_v0_causal_bold}
\end{equation}
For zero initialization, \(\bm w_0(0)=0\), this reduces to
\(\bm v^0_{\mathrm{init}}=\bm w^\star\).

Here \(\bm v^i_\alpha=\bm v^i_n(t)\) for \(\alpha=(n,t)\). Stacked notation
collects the \(S\) vectors into one vector; for example,
\[
    \bm v^0
    =
    \begin{pmatrix}
        \bm v^0_1\\
        \vdots\\
        \bm v^0_S
    \end{pmatrix}
    \in\mathbb R^{SM}.
\]
In this notation,
\begin{equation}
    \bm v^0
    =
    \bm 1\otimes\bm v^0_{\mathrm{init}}
    -
    \eta\bm\Theta\bm v^4,
    \label{eq:app_v0_stacked_bold}
\end{equation}
where \(\otimes\) denotes the Kronecker product and
\(\bm 1\otimes\bm v^0_{\mathrm{init}}\) repeats the initial residual across all
\(S\) epoch-step indices. The matrix \(\bm\Theta\) acts only on the epoch-step
indices; equivalently, the last term is
\((\bm\Theta\otimes\bm I_M)\bm v^4\).

\subsection{Single-site stochastic processes}
\label{app:single_site_processes}

The DMFT fixed point is as follows. Appendices~\ref{app:cavity_deriv} and
\ref{app:msr_path_integral} give its cavity and MSR derivations. In the
population eigenbasis, the effective variables are a representative minibatch
coordinate \(v^1\), a representative random-feature coordinate \(v^3\), and a
spectral mode \((v^0_k,v^2_k,v^4_k)\). Each variable retains the epoch-step
index \(\alpha=(n,t)\).

The effective process is
\begin{align}
    v^0_{k,\alpha}
    &=
    v^0_{\mathrm{init},k}
    -
    \eta
    \sum_\beta
    \Theta_{\alpha\beta}
    v^4_{k,\beta},
    \label{eq:app_eff_v0_full}
    \\
    v^1_\alpha
    &=
    u^1_\alpha
    +
    \sigma\epsilon_\alpha
    +
    \frac1B
    \sum_\beta
    [\mathcal D(\bm R_{0,2})]_{\alpha\beta}
    v^1_\beta,
    \label{eq:app_eff_v1_full}
    \\
    v^2_{k,\alpha}
    &=
    u^2_{k,\alpha}
    +
    \lambda_k
    \sum_\beta
    R_{1;\alpha\beta}
    v^0_{k,\beta},
    \label{eq:app_eff_v2_full}
    \\
    v^3_\alpha
    &=
    u^3_\alpha
    +
    \frac1N
    \sum_\beta
    [\bm R_{2,4}]_{\alpha\beta}
    v^3_\beta,
    \label{eq:app_eff_v3_full}
    \\
    v^4_{k,\alpha}
    &=
    u^4_{k,\alpha}
    +
    \sum_\beta
    R_{3;\alpha\beta}v^2_{k,\beta}.
    \label{eq:app_eff_v4_full}
\end{align}
Here \(v^0_{\mathrm{init},k}\) is the \(k\)-th coordinate of the initial
feature-space residual. For zero initialization, \(v^0_{\mathrm{init},k}=w_k^\star\).
For \(\alpha=(n,t)\), the scalar
\(\epsilon_\alpha\equiv\epsilon_t\) is the effective single-site label-noise
variable inherited from \(\bm\epsilon(t)\). It is therefore the same variable
at the same within-epoch label \(t\) in every epoch.

The zero-mean Gaussian fields have covariances
\begin{align}
    \left\langle
    u^1_\alpha u^1_\beta
    \right\rangle
    &=
    [\mathcal D(\bm C_0)]_{\alpha\beta},
    \label{eq:app_u1_cov}
    \\
    \left\langle
    u^2_{k,\alpha}u^2_{\ell,\beta}
    \right\rangle
    &=
    \delta_{k\ell}
    \frac{\lambda_k}{B}
    [\mathcal D(\bm C_1)]_{\alpha\beta},
    \label{eq:app_u2_cov}
    \\
    \left\langle
    u^3_\alpha u^3_\beta
    \right\rangle
    &=
    [\bm C_2]_{\alpha\beta},
    \label{eq:app_u3_cov}
    \\
    \left\langle
    u^4_{k,\alpha}u^4_{\ell,\beta}
    \right\rangle
    &=
    \delta_{k\ell}
    \frac1N
    [\bm C_3]_{\alpha\beta}.
    \label{eq:app_u4_cov}
\end{align}
All omitted cross-covariances among the Gaussian fields and the label-noise
field vanish. The label-noise covariance is
\[
    \left\langle
    \epsilon_\alpha\epsilon_\beta
    \right\rangle
    =
    [\mathcal D(\bm 1\bm 1^\top)]_{\alpha\beta}.
\]
Both \(u^1\) and the label noise are correlated across epochs only when the
minibatch labels agree.

The correlations \(\bm C_i\) and responses \(\bm R_{i,j}\) are determined
self-consistently below.

\subsection{Correlation functions}
\label{app:correlation_definitions}

The Gaussian covariances depend on the following correlation functions:
\begin{align}
    [\bm C_0]_{\alpha\beta}
    &=
    \sum_k
    \lambda_k
    \left\langle
        v^0_{k,\alpha}v^0_{k,\beta}
    \right\rangle,
    \label{eq:app_C0_def_full}
    \\
    [\bm C_1]_{\alpha\beta}
    &=
    \left\langle
        v^1_\alpha v^1_\beta
    \right\rangle,
    \label{eq:app_C1_def_full}
    \\
    [\bm C_2]_{\alpha\beta}
    &=
    \sum_k
    \left\langle
        v^2_{k,\alpha}v^2_{k,\beta}
    \right\rangle,
    \label{eq:app_C2_def_full}
    \\
    [\bm C_3]_{\alpha\beta}
    &=
    \left\langle
        v^3_\alpha v^3_\beta
    \right\rangle.
    \label{eq:app_C3_def_full}
\end{align}
These are uncentered second moments. In particular, \(v^0_k\) generally has a
nonzero mean, so \(\bm C_0\) includes both bias and variance. The test and train
losses are read from the diagonal entries
\begin{equation}
    \mathcal L_{\mathrm{test}}(\alpha)
    =
    [\bm C_0]_{\alpha\alpha}
    +
    \sigma^2,
    \qquad
    \widehat{\mathcal L}_{\mathrm{train}}(\alpha)
    =
    [\bm C_1]_{\alpha\alpha}.
    \label{eq:app_train_test_loss_from_C}
\end{equation}
The additive \(\sigma^2\) is the irreducible label-noise contribution to the
test mean-squared error.

The \(v^1\) and \(v^3\) equations give
\begin{align}
    \bm C_1
    &=
    \bm R_1
    \mathcal D\left(
        \bm C_0
        +
        \sigma^2\bm 1\bm 1^\top
    \right)
    \bm R_1^\top,
    \label{eq:app_C1_closed_def}
    \\
    \bm C_3
    &=
    \bm R_3\bm C_2\bm R_3^\top.
    \label{eq:app_C3_closed_def}
\end{align}
Here \(\bm C_0\) is the full two-time residual correlation. The projection
\(\mathcal D\) enters when this residual is sampled through the fixed
minibatches.

\subsection{Response functions}
\label{app:response_definitions}

The response functions use the standard random-feature DMFT convention.
\(R_{0,2}\) and \(R_{2,4}\) measure the responses of \(v^0\) to \(u^2\) and of
\(v^2\) to \(u^4\); \(R_1\) and \(R_3\) are the site responses to \(u^1\) and
\(u^3\). Label noise changes correlations but not these linear responses.

For each mode \(k\), define
\begin{align}
    R_{0,2;k;\alpha\beta}
    &=
    \left\langle
    \frac{\delta v^0_{k,\alpha}}
         {\delta u^2_{k,\beta}}
    \right\rangle,
    \label{eq:app_R02k_def}
    \\
    R_{2,4;k;\alpha\beta}
    &=
    \left\langle
    \frac{\delta v^2_{k,\alpha}}
         {\delta u^4_{k,\beta}}
    \right\rangle.
    \label{eq:app_R24k_def}
\end{align}
The spectrum-averaged responses are
\begin{align}
    [\bm R_{0,2}]_{\alpha\beta}
    &=
    \sum_k
    \lambda_k
    R_{0,2;k;\alpha\beta},
    \label{eq:app_R02_def}
    \\
    [\bm R_{2,4}]_{\alpha\beta}
    &=
    \sum_k
    R_{2,4;k;\alpha\beta}.
    \label{eq:app_R24_def}
\end{align}
The site responses are
\begin{align}
    R_{1;\alpha\beta}
    &=
    \left\langle
    \frac{\delta v^1_\alpha}
         {\delta u^1_\beta}
    \right\rangle,
    \label{eq:app_R1_def}
    \\
    R_{3;\alpha\beta}
    &=
    \left\langle
    \frac{\delta v^3_\alpha}
         {\delta u^3_\beta}
    \right\rangle.
    \label{eq:app_R3_def}
\end{align}

From \eqref{eq:app_eff_v1_full} and \eqref{eq:app_eff_v3_full}, these responses
obey
\begin{align}
    \bm R_1
    &=
    \left[
        \bm I
        -
        \frac1B
        \mathcal D(\bm R_{0,2})
    \right]^{-1},
    \label{eq:app_R1_R02_relation}
    \\
    \bm R_3
    &=
    \left[
        \bm I
        -
        \frac1N
        \bm R_{2,4}
    \right]^{-1}.
    \label{eq:app_R3_R24_relation}
\end{align}
These equations contain the feedback through the remaining sites.

\subsection{Mode transfer functions}
\label{app:mode_transfer_functions}

Combining \eqref{eq:app_eff_v0_full}, \eqref{eq:app_eff_v2_full}, and
\eqref{eq:app_eff_v4_full} gives the residual equation for each mode:
\begin{equation}
    v^0_{k,\alpha}
    =
    v^0_{\mathrm{init},k}
    -
    \eta
    \sum_\beta
    \Theta_{\alpha\beta}
    \left[
        u^4_{k,\beta}
        +
        \sum_\gamma
        R_{3;\beta\gamma}
        u^2_{k,\gamma}
        +
        \lambda_k
        \sum_{\gamma,\rho}
        R_{3;\beta\gamma}
        R_{1;\gamma\rho}
        v^0_{k,\rho}
    \right].
    \label{eq:app_v0_closed_before_H}
\end{equation}
Solving for \(\bm v^0_k\) gives
\begin{equation}
    \bm v^0_k
    =
    \bm H_k
    \left[
        v^0_{\mathrm{init},k}\bm 1
        -
        \eta\bm\Theta
        \left(
            \bm u^4_k+\bm R_3\bm u^2_k
        \right)
    \right],
    \label{eq:app_v0_solution_H}
\end{equation}
where
\begin{equation}
    \bm H_k
    =
    \left[
        \bm I
        +
        \eta\lambda_k
        \bm\Theta\bm R_3\bm R_1
    \right]^{-1}.
    \label{eq:app_Hk_def}
\end{equation}
We also define the spectrum-averaged transfer operator
\begin{equation}
    \bar{\bm H}
    =
    \sum_k
    \lambda_k\bm H_k.
    \label{eq:app_Hbar_def}
\end{equation}

The auxiliary responses can now be written in terms of the transfer functions:
\begin{align}
    \bm R_{0,2;k}
    &=
    -\eta\bm H_k\bm\Theta\bm R_3,
    \qquad
    \bm R_{0,2}
    =
    -\eta\bar{\bm H}\bm\Theta\bm R_3,
    \label{eq:app_R02_from_H}
    \\
    \bm R_{2,4;k}
    &=
    -\eta\lambda_k
    \bm R_1\bm H_k\bm\Theta,
    \qquad
    \bm R_{2,4}
    =
    -\eta
    \bm R_1\bar{\bm H}\bm\Theta.
    \label{eq:app_R24_from_H}
\end{align}
Substituting these into
\eqref{eq:app_R1_R02_relation}--\eqref{eq:app_R3_R24_relation} gives the closed
response equations
\begin{align}
    \bm R_1
    &=
    \left[
        \bm I
        +
        \frac{\eta}{B}
        \mathcal D(\bar{\bm H}\bm\Theta\bm R_3)
    \right]^{-1},
    \label{eq:app_R1_closed}
    \\
    \bm R_3
    &=
    \left[
        \bm I
        +
        \frac{\eta}{N}
        \bm R_1\bar{\bm H}\bm\Theta
    \right]^{-1}.
    \label{eq:app_R3_closed}
\end{align}

\subsection{Closed Matrix Equations for the Multi-Epoch DMFT}
\label{app:vectorized_dmft}

The equations used for numerical solution are collected here. Every matrix is
indexed by the \(S=T\mathcal E\) epoch-step pairs.
Define
\begin{equation}
    \bm M_k
    =
    \bm I
    -
    \eta\lambda_k
    \bm R_1\bm H_k\bm\Theta\bm R_3.
    \label{eq:app_Mk_def}
\end{equation}
This matrix propagates the Gaussian field \(\bm u^2_k\), generated by the
minibatch features, into \(\bm v^2_k\).

The closed response equations are
\begin{align}
    \bm H_k
    &=
    \left[
        \bm I
        +
        \eta\lambda_k\bm\Theta\bm R_3\bm R_1
    \right]^{-1},
    \qquad
    \bar{\bm H}
    =
    \sum_k\lambda_k\bm H_k,
    \label{eq:app_H_boxed}
    \\
    \bm R_1
    &=
    \left[
        \bm I
        +
        \frac{\eta}{B}
        \mathcal D(\bar{\bm H}\bm\Theta\bm R_3)
    \right]^{-1},
    \label{eq:app_R1_boxed}
    \\
    \bm R_3
    &=
    \left[
        \bm I
        +
        \frac{\eta}{N}
        \bm R_1\bar{\bm H}\bm\Theta
    \right]^{-1}.
    \label{eq:app_R3_boxed}
\end{align}

The closed correlation equations are
\begin{align}
    \bm C_1
    &=
    \bm R_1
    \mathcal D\left(
        \bm C_0
        +
        \sigma^2\bm 1\bm 1^\top
    \right)
    \bm R_1^\top,
    \qquad
    \bm C_3
    =
    \bm R_3\bm C_2\bm R_3^\top,
    \label{eq:app_C1C3_boxed}
    \\
    \bm C_0
    &=
    \sum_k
    \lambda_k
    \bm H_k
    \Bigg[
        (v^0_{\mathrm{init},k})^2\bm 1\bm 1^\top
        +
        \frac{\eta^2}{N}
        \bm\Theta\bm R_3
        \bm C_2
        \bm R_3^\top\bm\Theta^\top
        +
        \frac{\eta^2\lambda_k}{B}
        \bm\Theta\bm R_3
        \mathcal D(\bm C_1)
        \bm R_3^\top\bm\Theta^\top
    \Bigg]
    \bm H_k^\top,
    \label{eq:app_C0_boxed}
    \\
    \bm C_2
    &=
    \sum_k
    \lambda_k^2
    \bm R_1\bm H_k
    \left[
        (v^0_{\mathrm{init},k})^2\bm 1\bm 1^\top
        +
        \frac{\eta^2}{N}
        \bm\Theta\bm R_3
        \bm C_2
        \bm R_3^\top\bm\Theta^\top
    \right]
    \bm H_k^\top\bm R_1^\top
    +
    \frac1B
    \sum_k
    \lambda_k
    \bm M_k
    \mathcal D(\bm C_1)
    \bm M_k^\top .
    \label{eq:app_C2_boxed}
\end{align}
For zero initialization, these equations reduce to the same expressions with
\(v^0_{\mathrm{init},k}=w_k^\star\).

Finally, for \(\alpha=(n,t)\), the test and train losses are
\begin{equation}
    \mathcal L_{\mathrm{test}}(n,t)
    =
    [\bm C_0]_{\alpha\alpha}
    +
    \sigma^2,
    \qquad
    \widehat{\mathcal L}_{\mathrm{train}}(n,t)
    =
    [\bm C_1]_{\alpha\alpha}.
    \label{eq:app_test_train_loss_dmft}
\end{equation}
The additive \(\sigma^2\) is the irreducible label-noise contribution to the
test mean-squared error. In noiseless experiments, set \(\sigma=0\).

\subsection{Cavity Derivation of the DMFT}\label{app:cavity_deriv}

We derive the single-site process in Appendix~\ref{app:single_site_processes} by
adding one coordinate to the system and expanding the old trajectory to first
order in its coupling to that coordinate. The fields \(\bm v^0,\ldots,\bm v^4\)
are defined in Appendix~\ref{app:decomposing_dynamics}.

Throughout, \(\alpha=(n,t)\) and \(\beta=(n',t')\). The operators
\(\bm\Theta\) and \(\mathcal D\) enforce causality and fixed-minibatch reuse.

\paragraph{Adding a spectral mode.}
Add a population feature mode \(k=0\) with eigenvalue \(\lambda_0\).
The added mode has variables \(v^0_{0,\alpha},v^2_{0,\alpha},v^4_{0,\alpha}\).
It couples weakly to many data and random-feature sites, so its leading effect
is given by linear response. Tildes denote the enlarged system. The data field
responds to
\(\psi_{b0}(t_\beta)v^0_{0,\beta}\) as
\begin{equation}
    \widetilde v^1_{b,\alpha}
    =
    v^1_{b,\alpha}
    +
    \sum_\beta
    R_{1;\alpha\beta}
    \psi_{b0}(t_\beta)v^0_{0,\beta}
    +
    o(1).
    \label{eq:app_cavity_v1_tilde_mode}
\end{equation}
Similarly, the old random-feature-site field responds to the perturbation
\(A_{i0}v^2_{0,\beta}\):
\begin{equation}
    \widetilde v^3_{i,\alpha}
    =
    v^3_{i,\alpha}
    +
    \sum_\beta
    R_{3;\alpha\beta}
    A_{i0}v^2_{0,\beta}
    +
    o(1).
    \label{eq:app_cavity_v3_tilde_mode}
\end{equation}

The added \(v^2\) coordinate is
\begin{align}
    v^2_{0,\alpha}
    &=
    \frac1B
    \sum_{b=1}^B
    \psi_{b0}(t_\alpha)
    \widetilde v^1_{b,\alpha}
    \nonumber\\
    &=
    \underbrace{
    \frac1B
    \sum_{b=1}^B
    \psi_{b0}(t_\alpha)v^1_{b,\alpha}
    }_{u^2_{0,\alpha}}
    +
    \frac1B
    \sum_{b=1}^B
    \sum_\beta
    \psi_{b0}(t_\alpha)\psi_{b0}(t_\beta)
    R_{1;\alpha\beta}v^0_{0,\beta}
    +
    o(1).
    \label{eq:app_cavity_v2_mode_expansion}
\end{align}
The first term is asymptotically Gaussian, with covariance
\begin{equation}
    \left\langle
    u^2_{0,\alpha}u^2_{0,\beta}
    \right\rangle
    =
    \frac{\lambda_0}{B}
    [\mathcal D(\bm C_1)]_{\alpha\beta}.
    \label{eq:app_cavity_u2_cov}
\end{equation}
Only equal minibatch labels share the same data, which gives \(\mathcal D\).
The second term self-averages to the Onsager term
\begin{equation}
    v^2_{0,\alpha}
    =
    u^2_{0,\alpha}
    +
    \lambda_0
    \sum_\beta
    R_{1;\alpha\beta}v^0_{0,\beta}.
    \label{eq:app_cavity_v2_mode_process}
\end{equation}

Likewise,
\begin{align}
    v^4_{0,\alpha}
    &=
    \frac1N
    \sum_{i=1}^N
    A_{i0}
    \widetilde v^3_{i,\alpha}
    \nonumber\\
    &=
    \underbrace{
    \frac1N
    \sum_{i=1}^N
    A_{i0}v^3_{i,\alpha}
    }_{u^4_{0,\alpha}}
    +
    \frac1N
    \sum_{i=1}^N
    \sum_\beta
    A_{i0}^2
    R_{3;\alpha\beta}v^2_{0,\beta}
    +
    o(1).
\end{align}
The random term is Gaussian with covariance
\begin{equation}
    \left\langle
    u^4_{0,\alpha}u^4_{0,\beta}
    \right\rangle
    =
    \frac1N[\bm C_3]_{\alpha\beta},
    \label{eq:app_cavity_u4_cov}
\end{equation}
and the deterministic part self-averages to
\begin{equation}
    v^4_{0,\alpha}
    =
    u^4_{0,\alpha}
    +
    \sum_\beta
    R_{3;\alpha\beta}v^2_{0,\beta}.
    \label{eq:app_cavity_v4_mode_process}
\end{equation}
Together with
\[
    v^0_{0,\alpha}
    =
    v^0_{\mathrm{init},0}
    -
    \eta
    \sum_\beta
    \Theta_{\alpha\beta}v^4_{0,\beta},
\]
this gives the mode part of the effective process.

\paragraph{Adding a data site.}
Add one data coordinate at a fixed within-epoch minibatch label. Its
contribution to \(v^2\) perturbs the residual according to
\begin{equation}
    \widetilde v^0_{k,\alpha}
    =
    v^0_{k,\alpha}
    +
    \frac1B
    \sum_\beta
    R_{0,2;k;\alpha\beta}
    \psi_{0k}(t_\beta)v^1_{0,\beta}
    +
    o(1).
    \label{eq:app_cavity_v0_tilde_data}
\end{equation}
The new data-site field is
\begin{align}
    v^1_{0,\alpha}
    &=
    \sum_k
    \psi_{0k}(t_\alpha)
    \widetilde v^0_{k,\alpha}
    +
    \sigma\epsilon_\alpha
    \nonumber\\
    &=
    \underbrace{
    \sum_k
    \psi_{0k}(t_\alpha)v^0_{k,\alpha}
    }_{u^1_\alpha}
    +
    \sigma\epsilon_\alpha
    +
    \frac1B
    \sum_{\beta}
    \sum_k
    \psi_{0k}(t_\alpha)\psi_{0k}(t_\beta)
    R_{0,2;k;\alpha\beta}
    v^1_{0,\beta}
    +
    o(1).
    \label{eq:app_cavity_v1_data_expansion}
\end{align}
The Gaussian source satisfies
\begin{equation}
    \left\langle
    u^1_\alpha u^1_\beta
    \right\rangle
    =
    [\mathcal D(\bm C_0)]_{\alpha\beta}.
    \label{eq:app_cavity_u1_cov}
\end{equation}
Rows from different minibatches are independent, while the row at a given
label is reused across epochs. The second term therefore self-averages to
\begin{equation}
    v^1_{0,\alpha}
    =
    u^1_\alpha
    +
    \sigma\epsilon_\alpha
    +
    \frac1B
    \sum_\beta
    [\mathcal D(\bm R_{0,2})]_{\alpha\beta}
    v^1_{0,\beta}.
    \label{eq:app_cavity_v1_data_process}
\end{equation}
The label-noise field has the same reuse pattern:
\[
    \left\langle
    \epsilon_\alpha\epsilon_\beta
    \right\rangle
    =
    [\mathcal D(\bm 1\bm 1^\top)]_{\alpha\beta}.
\]

\paragraph{Adding a random-feature site.}
Add one random-feature coordinate \(i=0\). Its contribution to \(v^4\)
perturbs \(v^2\) according to
\begin{equation}
    \widetilde v^2_{k,\alpha}
    =
    v^2_{k,\alpha}
    +
    \frac1N
    \sum_\beta
    R_{2,4;k;\alpha\beta}
    A_{0k}v^3_{0,\beta}
    +
    o(1).
    \label{eq:app_cavity_v2_tilde_feature}
\end{equation}
The newly added random-feature-site field is
\begin{align}
    v^3_{0,\alpha}
    &=
    \sum_k
    A_{0k}
    \widetilde v^2_{k,\alpha}
    \nonumber\\
    &=
    \underbrace{
    \sum_k A_{0k}v^2_{k,\alpha}
    }_{u^3_\alpha}
    +
    \frac1N
    \sum_\beta
    \sum_k
    A_{0k}^2
    R_{2,4;k;\alpha\beta}
    v^3_{0,\beta}
    +
    o(1).
\end{align}
The random part is Gaussian with covariance
\begin{equation}
    \left\langle
    u^3_\alpha u^3_\beta
    \right\rangle
    =
    [\bm C_2]_{\alpha\beta},
    \label{eq:app_cavity_u3_cov}
\end{equation}
and the deterministic part gives
\begin{equation}
    v^3_{0,\alpha}
    =
    u^3_\alpha
    +
    \frac1N
    \sum_\beta
    [\bm R_{2,4}]_{\alpha\beta}
    v^3_{0,\beta}.
    \label{eq:app_cavity_v3_feature_process}
\end{equation}

\paragraph{Closure.}
The three additions give the effective process
\eqref{eq:app_eff_v0_full}--\eqref{eq:app_eff_v4_full}. The response functions
are the averaged linear sensitivities appearing in the cavity expansions:
\begin{align}
    R_{1;\alpha\beta}
    &=
    \left\langle
    \frac{\delta v^1_\alpha}{\delta u^1_\beta}
    \right\rangle,
    &
    R_{3;\alpha\beta}
    &=
    \left\langle
    \frac{\delta v^3_\alpha}{\delta u^3_\beta}
    \right\rangle,
    \\
    R_{0,2;k;\alpha\beta}
    &=
    \left\langle
    \frac{\delta v^0_{k,\alpha}}{\delta u^2_{k,\beta}}
    \right\rangle,
    &
    R_{2,4;k;\alpha\beta}
    &=
    \left\langle
    \frac{\delta v^2_{k,\alpha}}{\delta u^4_{k,\beta}}
    \right\rangle.
\end{align}
Combining the equations for \((v^0_k,v^2_k,v^4_k)\) gives
\begin{equation}
    \bm H_k
    =
    \left[
        \bm I
        +
        \eta\lambda_k\bm\Theta\bm R_3\bm R_1
    \right]^{-1}.
\end{equation}
Substitution gives the response closure
\eqref{eq:app_H_boxed}--\eqref{eq:app_R3_boxed}. Taking second moments of the
Gaussian effective process gives the correlation closure
\eqref{eq:app_C1C3_boxed}--\eqref{eq:app_C2_boxed}.

\subsection{MSR Path Integral Derivation of the DMFT}\label{app:msr_path_integral}

We derive the same DMFT with the Martin--Siggia--Rose (MSR) generating
functional. We extend the random-feature derivation of
\citet{bordelon2024dynamical} to multi-epoch SGD by retaining the epoch-step
index \(\alpha=(n,t)\). The operators \(\bm\Theta\) and \(\mathcal D\)
encode causality and reuse of fixed minibatches, respectively.

We use the fields from Appendix~\ref{app:decomposing_dynamics}; their stacked
residual equation is \eqref{eq:app_v0_stacked_bold}. Dot products below include
the epoch-step and site indices.

\paragraph{Generating functional, normalization, and response convention.}
For a finite trajectory, we use the Fourier representation
\[
    \delta(g)
    =
    \int_{\mathbb R}\frac{d\hat v}{2\pi}
    \exp(i\hat v g).
\]
Accordingly, \(D\bm v\) denotes the product of Lebesgue measures over all
ordinary fields, and \(D\hat{\bm v}\) includes one factor \(d\hat v/(2\pi)\)
for every MSR response field. The discrete-time constraints are ordered
causally and have unit coefficients on their ordinary fields. Their Jacobian
is therefore block triangular with unit diagonal and contributes no
trajectory-dependent determinant.

Introduce MSR response fields
\(\hat{\bm v}^0,\hat{\bm v}^1,\hat{\bm v}^2,\hat{\bm v}^3,\hat{\bm v}^4\)
to enforce the constraints. We also add auxiliary kicks
\(\bm h^1,\bm h^2,\bm h^3,\bm h^4\) to the right-hand sides:
\begin{align}
    \bm v^0
    &=
    \bm 1\otimes\bm v^0_{\mathrm{init}}
    -
    \eta\bm\Theta\bm v^4,
    \\
    \bm v^1_\alpha
    &=
    \bm\Psi(t_\alpha)\bm v^0_\alpha
    +
    \sigma\bm\epsilon(t_\alpha)
    +
    \bm h^1_\alpha,
    \\
    \bm v^2_\alpha
    &=
    \frac1B\bm\Psi(t_\alpha)^\top\bm v^1_\alpha
    +
    \bm h^2_\alpha,
    \\
    \bm v^3_\alpha
    &=
    \bm A\bm v^2_\alpha
    +
    \bm h^3_\alpha,
    \\
    \bm v^4_\alpha
    &=
    \frac1N\bm A^\top\bm v^3_\alpha
    +
    \bm h^4_\alpha.
\end{align}
With a source \(J\) coupled to \(\bm v^0\), the generating functional is
\begin{align}
Z[J,\bm h]
&=
\Bigg\langle
\int D\bm v\,D\hat{\bm v}\,
\exp\Bigg\{
    i\hat{\bm v}^0\cdot
    \left(
        \bm v^0-\bm 1\otimes \bm v^0_{\mathrm{init}}
        +\eta\bm\Theta\bm v^4
    \right)
    + \sum_\alpha J_\alpha\cdot\bm v^0_\alpha
\nonumber\\
&\qquad
    +
    i\hat{\bm v}^1\cdot
    \left(
        \bm v^1-\bm\Psi\bm v^0-\sigma\bm\epsilon-\bm h^1
    \right)
    +
    i\hat{\bm v}^2\cdot
    \left(
        \bm v^2-\frac1B\bm\Psi^\top\bm v^1-\bm h^2
    \right)
\nonumber\\
&\qquad
    +
    i\hat{\bm v}^3\cdot
    \left(
        \bm v^3-\bm A\bm v^2-\bm h^3
    \right)
    +
    i\hat{\bm v}^4\cdot
    \left(
        \bm v^4-\frac1N\bm A^\top\bm v^3-\bm h^4
    \right)
\Bigg\}
\Bigg\rangle_{\bm A,\bm\Psi,\epsilon}.
\label{eq:app_msr_Z_full}
\end{align}
For every fixed kick, normalization of the deterministic trajectory measure
gives
\begin{equation}
    Z[0,\bm h]=1.
    \label{eq:app_msr_normalization}
\end{equation}
This identity fixes both the response sign and the pure-response
correlators. Because a kick enters a constraint as \(-\bm h^a\),
\begin{equation}
    \left.
    \frac{\delta\langle v^a_\alpha\rangle}
         {\delta h^b_\beta}
    \right|_{\bm h=0}
    =
    -i
    \left\langle
        v^a_\alpha\hat v^b_\beta
    \right\rangle_{\!c}.
    \label{eq:app_msr_response_sign}
\end{equation}
At zero source, \(\langle\hat v^a_\alpha\rangle=0\), so the connected
correlator can be replaced by the ordinary one. Repeated derivatives of
\eqref{eq:app_msr_normalization} with respect to the kicks also give
\(\langle\hat v^a_\alpha\hat v^b_\beta\rangle=0\)
\citep{crisanti2018path,helias2020statistical}. The additive kicks enter
the same equations as the Gaussian fields \(u^a\) in the effective process;
hence \eqref{eq:app_msr_response_sign} agrees with the functional-derivative
definitions in Appendix~\ref{app:response_definitions}.

\paragraph{Average over the random-feature matrix.}
The entries of \(\bm A\) and the whitened minibatch features are independent
Gaussians with the covariances specified below. The disorder averages in this
step are therefore exact.

The terms depending on \(\bm A\) are
\begin{equation}
    -i
    \sum_{i,k,\alpha}
    A_{ik}
    \left[
        \hat v^3_{i,\alpha}v^2_{k,\alpha}
        +
        \frac1N\hat v^4_{k,\alpha}v^3_{i,\alpha}
    \right].
\end{equation}
Define
\begin{equation}
    Y_{ik}
    =
    \sum_\alpha
    \left[
        \hat v^3_{i,\alpha}v^2_{k,\alpha}
        +
        \frac1N\hat v^4_{k,\alpha}v^3_{i,\alpha}
    \right].
\end{equation}
For Gaussian \(A_{ik}\) with unit variance,
\begin{equation}
    \left\langle
    \exp\left(
        -i\sum_{i,k}A_{ik}Y_{ik}
    \right)
    \right\rangle_{\bm A}
    =
    \exp\left(
        -\frac12\sum_{i,k}Y_{ik}^2
    \right).
\end{equation}
Expanding the square gives three types of contractions:
\begin{align}
    -\frac12\sum_{\alpha,\beta}
    \left(
        \sum_k v^2_{k,\alpha}v^2_{k,\beta}
    \right)
    \left(
        \sum_i \hat v^3_{i,\alpha}\hat v^3_{i,\beta}
    \right),
    \label{eq:app_A_C2_contraction}
    \\
    -\frac1{2N^2}\sum_{\alpha,\beta}
    \left(
        \sum_i v^3_{i,\alpha}v^3_{i,\beta}
    \right)
    \left(
        \sum_k \hat v^4_{k,\alpha}\hat v^4_{k,\beta}
    \right),
    \label{eq:app_A_C3_contraction}
    \\
    -\frac1N\sum_{\alpha,\beta}
    \left(
        \sum_k v^2_{k,\alpha}\hat v^4_{k,\beta}
    \right)
    \left(
        \sum_i \hat v^3_{i,\alpha}v^3_{i,\beta}
    \right).
    \label{eq:app_A_response_contraction}
\end{align}
The first two contractions generate the Gaussian fields \(u^3\) and \(u^4_k\),
with covariances \(\bm C_2\) and \(N^{-1}\bm C_3\). The third contraction is
the Onsager reaction term involving the responses \(\bm R_{2,4}\) and
\(\bm R_3\).

\paragraph{Average over the fixed minibatch matrices.}
The terms depending on the data matrix at within-epoch label \(t\) are
\begin{equation}
    -i
    \sum_{b,k}
    \Psi_{bk}(t)
    \sum_{\alpha:t_\alpha=t}
    \left[
        \hat v^1_{b,\alpha}v^0_{k,\alpha}
        +
        \frac1B
        \hat v^2_{k,\alpha}v^1_{b,\alpha}
    \right].
\end{equation}
Define
\begin{equation}
    X_{bk}(t)
    =
    \sum_{\alpha:t_\alpha=t}
    \left[
        \hat v^1_{b,\alpha}v^0_{k,\alpha}
        +
        \frac1B
        \hat v^2_{k,\alpha}v^1_{b,\alpha}
    \right].
\end{equation}
Since
\(\langle \Psi_{bk}(t)\Psi_{b' \ell}(t')\rangle
=
\delta_{bb'}\delta_{tt'}\delta_{k\ell}\lambda_k\), the average over
\(\bm\Psi(t)\) gives
\begin{equation}
    \left\langle
    \exp\left(
        -i\sum_{t,b,k}\Psi_{bk}(t)X_{bk}(t)
    \right)
    \right\rangle_{\bm\Psi}
    =
    \exp\left(
        -\frac12
        \sum_{t,b,k}
        \lambda_k X_{bk}(t)^2
    \right).
\end{equation}
Expanding this square gives
\begin{align}
    -\frac12
    \sum_{\alpha,\beta}
    \left(
        \sum_k\lambda_k v^0_{k,\alpha}v^0_{k,\beta}
    \right)
    \left(
        \sum_b
        \hat v^1_{b,\alpha}\hat v^1_{b,\beta}
    \right)
    \mathbf 1\{t_\alpha=t_\beta\},
    \label{eq:app_Psi_C0_contraction}
    \\
    -\frac1{2B^2}
    \sum_{\alpha,\beta}
    \left(
        \sum_k
        \lambda_k\hat v^2_{k,\alpha}\hat v^2_{k,\beta}
    \right)
    \left(
        \sum_b
        v^1_{b,\alpha}v^1_{b,\beta}
    \right)
    \mathbf 1\{t_\alpha=t_\beta\},
    \label{eq:app_Psi_C1_contraction}
    \\
    -\frac1B
    \sum_{\alpha,\beta}
    \left(
        \sum_k
        \lambda_k v^0_{k,\alpha}\hat v^2_{k,\beta}
    \right)
    \left(
        \sum_b
        \hat v^1_{b,\alpha}v^1_{b,\beta}
    \right)
    \mathbf 1\{t_\alpha=t_\beta\}.
    \label{eq:app_Psi_response_contraction}
\end{align}
The first contraction generates the \(u^1\) covariance
\(\mathcal D(\bm C_0)\). The second generates the \(u^2_k\) covariance
\[
    \delta_{k\ell}
    \frac{\lambda_k}{B}
    [\mathcal D(\bm C_1)]_{\alpha\beta}.
\]
The third generates the Onsager term associated with the minibatch features,
\(\mathcal D(\bm R_{0,2})\) in the \(v^1\) process and \(\bm R_1\) in the
\(v^2_k\) process. The projection \(\mathcal D\) appears because only
epoch-step pairs with the same within-epoch minibatch label share the same data
matrix.

Averaging over the fixed label-noise vectors similarly produces the quadratic
term
\begin{equation}
    -\frac{\sigma^2}{2}
    \sum_{\alpha,\beta}
    \left(
        \sum_b
        \hat v^1_{b,\alpha}\hat v^1_{b,\beta}
    \right)
    \mathbf 1\{t_\alpha=t_\beta\}.
    \label{eq:app_epsilon_contraction}
\end{equation}
Equivalently, this term can be represented by the explicit Gaussian field
\(\sigma\epsilon_\alpha\) in the effective process, with
\[
    \left\langle
        \sigma\epsilon_\alpha\,\sigma\epsilon_\beta
    \right\rangle
    =
    \sigma^2
    [\mathcal D(\bm1\bm1^\top)]_{\alpha\beta}.
\]

\paragraph{Order parameters and delta-functional insertions.}
Introduce the path-level order parameters below. In the generating functional,
these are integration variables; their saddle-point values give the DMFT
correlations and responses.
\begin{align}
    [\bm C_0]_{\alpha\beta}
    &=
    \sum_k\lambda_k
    v^0_{k,\alpha}v^0_{k,\beta},
    &
    [\bm C_1]_{\alpha\beta}
    &=
    \frac1B\sum_b
    v^1_{b,\alpha}v^1_{b,\beta},
    \nonumber\\
    [\bm C_2]_{\alpha\beta}
    &=
    \sum_k
    v^2_{k,\alpha}v^2_{k,\beta},
    &
    [\bm C_3]_{\alpha\beta}
    &=
    \frac1N\sum_i
    v^3_{i,\alpha}v^3_{i,\beta},
    \label{eq:app_msr_path_correlations}
\end{align}
and the two site responses
\begin{align}
    R_{1;\alpha\beta}
    &=
    -\frac{i}{B}
    \sum_b
    v^1_{b,\alpha}\hat v^1_{b,\beta},
    &
    R_{3;\alpha\beta}
    &=
    -\frac{i}{N}
    \sum_i
    v^3_{i,\alpha}\hat v^3_{i,\beta}.
    \label{eq:app_msr_path_responses}
\end{align}
The factors \(1/B\) and \(1/N\) make these the empirical versions of the
single-site responses. Their conjugate order parameters are denoted by
\(\hat{\bm C}_i,\hat{\bm R}_1,\hat{\bm R}_3\).

For matrices over epoch-step indices, define
\[
    \bm X:\bm Y
    :=
    \sum_{\alpha,\beta}
    X_{\alpha\beta}Y_{\alpha\beta}.
\]
Each definition is imposed by a Fourier representation of a delta functional.
For example, with normalization constants chosen so that each identity equals
one,
\begin{align}
    1
    &=
    \mathcal N_C
    \int D\bm C_1D\hat{\bm C}_1
    \exp\left\{
        \frac B2
        \hat{\bm C}_1:
        \left(
            \bm C_1
            -
            \frac1B\sum_b
            \bm v^1_b\bm v_b^{1\top}
        \right)
    \right\},
    \label{eq:app_msr_delta_C1}
    \\
    1
    &=
    \mathcal N_R
    \int D\bm R_1D\hat{\bm R}_1
    \exp\left\{
        -\bm R_1:\hat{\bm R}_1^\top
        -
        \frac{i}{B}
        \sum_{b,\alpha,\beta}
        \hat R_{1;\alpha\beta}
        \hat v^1_{b,\alpha}v^1_{b,\beta}
    \right\}.
    \label{eq:app_msr_delta_R1}
\end{align}
The conjugate integrations run along the corresponding Fourier contours.
Analogous identities impose \(\bm C_0,\bm C_2,\bm C_3\) and \(\bm R_3\).
The transpose in \eqref{eq:app_msr_delta_R1} is an index-ordering convention
chosen so that the first index of every response is the responding time.

\paragraph{Explicit factorized action.}
Inserting these identities into the disorder-averaged generating functional
factorizes the microscopic variables into a representative data-site path, a
representative random-feature-site path, and one path for each spectral mode:
\begin{equation}
    Z[J,\bm h]
    =
    \int D\mathcal QD\hat{\mathcal Q}\,
    \exp\left\{
        \mathcal S_{\mathrm{op}}
    \right\}
    \mathcal Z_1^B
    \mathcal Z_3^N
    \prod_{k=1}^M
    \mathcal Z_{0,2,4;k}.
    \label{eq:app_msr_factorized_Z}
\end{equation}
Here
\[
    \mathcal Q
    =
    \{
        \bm C_0,\bm C_1,\bm C_2,\bm C_3,
        \bm R_1,\bm R_3
    \},
\]
and \(\hat{\mathcal Q}\) contains the corresponding conjugates. The
order-parameter part of the action is
\begin{align}
    \mathcal S_{\mathrm{op}}
    &=
    \frac12
    \left(
        \hat{\bm C}_0:\bm C_0
        +
        B\hat{\bm C}_1:\bm C_1
        +
        \hat{\bm C}_2:\bm C_2
        +
        N\hat{\bm C}_3:\bm C_3
    \right)
    \nonumber\\
    &\qquad
    -
    \bm R_1:\hat{\bm R}_1^\top
    -
    \bm R_3:\hat{\bm R}_3^\top.
    \label{eq:app_msr_order_action}
\end{align}
The three factors keep the \(B\), \(N\), and spectral-mode contributions
separate.

The data-site factor is
\begin{equation}
    \mathcal Z_1
    =
    \mathbb E_{\epsilon}
    \int D\bm v^1D\hat{\bm v}^1
    \exp\left\{
        \mathcal S_1
    \right\},
    \label{eq:app_msr_Z1}
\end{equation}
where \(\epsilon_{(n,t)}=\epsilon_t\), the \(T\) variables
\(\{\epsilon_t\}\) are independent standard Gaussians, and
\begin{align}
    \mathcal S_1
    &=
    i
    \sum_\alpha
    \hat v^1_\alpha
    \left[
        v^1_\alpha
        -
        h^1_\alpha
        -
        \sigma\epsilon_\alpha
        -
        \frac1B
        \sum_\beta
        \hat R_{1;\alpha\beta}v^1_\beta
    \right]
    \nonumber\\
    &\qquad
    -
    \frac12
    \hat{\bm v}^{1\top}
    \mathcal D(\bm C_0)
    \hat{\bm v}^1
    -
    \frac12
    \hat{\bm C}_1:
    \left(
        \bm v^1\bm v^{1\top}
    \right).
    \label{eq:app_msr_S1}
\end{align}
The path in \(\mathcal Z_1\) spans all epoch-step indices. The operator
\(\mathcal D\) makes its components at different within-epoch labels
independent, while retaining reuse of the same label across epochs.

The random-feature-site factor is
\begin{equation}
    \mathcal Z_3
    =
    \int D\bm v^3D\hat{\bm v}^3
    \exp\left\{
        \mathcal S_3
    \right\},
    \label{eq:app_msr_Z3}
\end{equation}
with
\begin{align}
    \mathcal S_3
    &=
    i
    \sum_\alpha
    \hat v^3_\alpha
    \left[
        v^3_\alpha
        -
        h^3_\alpha
        -
        \frac1N
        \sum_\beta
        \hat R_{3;\alpha\beta}v^3_\beta
    \right]
    \nonumber\\
    &\qquad
    -
    \frac12
    \hat{\bm v}^{3\top}
    \bm C_2
    \hat{\bm v}^3
    -
    \frac12
    \hat{\bm C}_3:
    \left(
        \bm v^3\bm v^{3\top}
    \right).
    \label{eq:app_msr_S3}
\end{align}

Finally, the factor for spectral mode \(k\) is
\begin{equation}
    \mathcal Z_{0,2,4;k}
    =
    \int
    D\bm v^0_kD\hat{\bm v}^0_k
    D\bm v^2_kD\hat{\bm v}^2_k
    D\bm v^4_kD\hat{\bm v}^4_k
    \exp\left\{
        \mathcal S_{0,2,4;k}
    \right\},
    \label{eq:app_msr_Z024}
\end{equation}
where
\begin{align}
    \mathcal S_{0,2,4;k}
    &=
    i\hat{\bm v}^{0\top}_k
    \left(
        \bm v^0_k
        -
        v^0_{\mathrm{init},k}\bm 1
        +
        \eta\bm\Theta\bm v^4_k
    \right)
    +
    \bm J_k^\top\bm v^0_k
    \nonumber\\
    &\quad
    +
    i\hat{\bm v}^{2\top}_k
    \left[
        \bm v^2_k
        -
        \bm h^2_k
        -
        \lambda_k
        \mathcal D(\bm R_1)\bm v^0_k
    \right]
    \nonumber\\
    &\quad
    +
    i\hat{\bm v}^{4\top}_k
    \left[
        \bm v^4_k
        -
        \bm h^4_k
        -
        \bm R_3\bm v^2_k
    \right]
    \nonumber\\
    &\quad
    -
    \frac{\lambda_k}{2B}
    \hat{\bm v}^{2\top}_k
    \mathcal D(\bm C_1)
    \hat{\bm v}^2_k
    -
    \frac1{2N}
    \hat{\bm v}^{4\top}_k
    \bm C_3
    \hat{\bm v}^4_k
    \nonumber\\
    &\quad
    -
    \frac{\lambda_k}{2}
    \hat{\bm C}_0:
    \left(
        \bm v^0_k\bm v_k^{0\top}
    \right)
    -
    \frac12
    \hat{\bm C}_2:
    \left(
        \bm v^2_k\bm v_k^{2\top}
    \right).
    \label{eq:app_msr_S024}
\end{align}
The projection \(\mathcal D\) in the \(\bm v^2_k\) response term follows
directly from \eqref{eq:app_Psi_response_contraction}. At the physical saddle,
\(\bm R_1\) itself has this block support, so
\(\mathcal D(\bm R_1)=\bm R_1\).

\paragraph{Saddle-point equations.}
At fixed training horizon, we obtain the DMFT closure by evaluating the
order-parameter integral in \eqref{eq:app_msr_factorized_Z} in the
saddle-point approximation. The preceding Gaussian disorder averages are
exact; Theorem~\ref{thm:app_finite_horizon_error} establishes the loss
approximation separately at fixed feature dimension and training horizon. Let
\(\langle\cdot\rangle_1\), \(\langle\cdot\rangle_3\), and
\(\langle\cdot\rangle_k\) denote normalized averages under
\(\mathcal Z_1\), \(\mathcal Z_3\), and
\(\mathcal Z_{0,2,4;k}\), respectively. Variation with respect to the
conjugate correlations gives
\begin{align}
    [\bm C_0]_{\alpha\beta}
    &=
    \sum_k\lambda_k
    \left\langle
        v^0_{k,\alpha}v^0_{k,\beta}
    \right\rangle_k,
    &
    [\bm C_1]_{\alpha\beta}
    &=
    \left\langle
        v^1_\alpha v^1_\beta
    \right\rangle_1,
    \nonumber\\
    [\bm C_2]_{\alpha\beta}
    &=
    \sum_k
    \left\langle
        v^2_{k,\alpha}v^2_{k,\beta}
    \right\rangle_k,
    &
    [\bm C_3]_{\alpha\beta}
    &=
    \left\langle
        v^3_\alpha v^3_\beta
    \right\rangle_3.
    \label{eq:app_msr_correlation_saddle}
\end{align}
Variation with respect to the conjugate responses gives the response-sign
relations derived independently in \eqref{eq:app_msr_response_sign}:
\begin{align}
    R_{1;\alpha\beta}
    &=
    -i
    \left\langle
        v^1_\alpha\hat v^1_\beta
    \right\rangle_1,
    &
    R_{3;\alpha\beta}
    &=
    -i
    \left\langle
        v^3_\alpha\hat v^3_\beta
    \right\rangle_3.
    \label{eq:app_msr_site_response_saddle}
\end{align}

The variations with respect to the physical site responses determine their
conjugates. Define
\begin{align}
    R_{0,2;k;\alpha\beta}
    &=
    -i
    \left\langle
        v^0_{k,\alpha}\hat v^2_{k,\beta}
    \right\rangle_k,
    &
    R_{2,4;k;\alpha\beta}
    &=
    -i
    \left\langle
        v^2_{k,\alpha}\hat v^4_{k,\beta}
    \right\rangle_k,
    \nonumber\\
    [\bm R_{0,2}]_{\alpha\beta}
    &=
    \sum_k\lambda_k
    R_{0,2;k;\alpha\beta},
    &
    [\bm R_{2,4}]_{\alpha\beta}
    &=
    \sum_k
    R_{2,4;k;\alpha\beta}.
    \label{eq:app_msr_cross_response_saddle}
\end{align}
The conjugate-response saddle equations are then
\begin{equation}
    \hat{\bm R}_1
    =
    \mathcal D(\bm R_{0,2}),
    \qquad
    \hat{\bm R}_3
    =
    \bm R_{2,4}.
    \label{eq:app_msr_conjugate_response_saddle}
\end{equation}
Thus the two response functions that appear as feedback on representative
sites are the conjugates of the site responses.

Variation with respect to the physical correlations gives
\begin{align}
    \hat{\bm C}_0
    &=
    B\,
    \mathcal D\left(
        \left\langle
            \hat{\bm v}^1\hat{\bm v}^{1\top}
        \right\rangle_1
    \right),
    &
    \hat{\bm C}_1
    &=
    \frac1{B^2}
    \mathcal D\left(
        \sum_k\lambda_k
        \left\langle
            \hat{\bm v}^2_k\hat{\bm v}^{2\top}_k
        \right\rangle_k
    \right),
    \nonumber\\
    \hat{\bm C}_2
    &=
    N
    \left\langle
        \hat{\bm v}^3\hat{\bm v}^{3\top}
    \right\rangle_3,
    &
    \hat{\bm C}_3
    &=
    \frac1{N^2}
    \sum_k
    \left\langle
        \hat{\bm v}^4_k\hat{\bm v}^{4\top}_k
    \right\rangle_k.
    \label{eq:app_msr_conjugate_correlation_saddle}
\end{align}
All pure-response correlators on the right-hand side vanish by
\eqref{eq:app_msr_normalization}. Therefore the physical saddle has
\begin{equation}
    \hat{\bm C}_0
    =
    \hat{\bm C}_1
    =
    \hat{\bm C}_2
    =
    \hat{\bm C}_3
    =
    \bm 0.
    \label{eq:app_msr_chat_zero}
\end{equation}

\paragraph{Hubbard--Stratonovich reduction.}
For any positive semidefinite covariance \(\bm K\), use
\begin{equation}
    \exp\left(
        -\frac12\hat{\bm v}^{\top}\bm K\hat{\bm v}
    \right)
    =
    \int d\mu_{\bm K}(\bm u)\,
    \exp\left(
        -i\hat{\bm v}^{\top}\bm u
    \right),
    \label{eq:app_msr_HS_identity}
\end{equation}
where \(d\mu_{\bm K}\) is the normalized centered Gaussian measure with
covariance \(\bm K\). Singular covariance matrices are understood through
their Gaussian law on the support of \(\bm K\).

Substituting \eqref{eq:app_msr_conjugate_response_saddle} and
\eqref{eq:app_msr_chat_zero} into \(\mathcal Z_1\), and applying
\eqref{eq:app_msr_HS_identity}, gives
\begin{align}
    \mathcal Z_1
    &=
    \mathbb E_{\bm u^1,\epsilon}
    \int D\bm v^1D\hat{\bm v}^1
    \exp\Bigg\{
        i\sum_\alpha
        \hat v^1_\alpha
        \left[
            v^1_\alpha
            -
            h^1_\alpha
            -
            u^1_\alpha
            -
            \sigma\epsilon_\alpha
    \right.\nonumber\\[-2mm]
    &\hspace{6.0cm}\left.
            -
            \frac1B
            \sum_\beta
            [\mathcal D(\bm R_{0,2})]_{\alpha\beta}
            v^1_\beta
        \right]
    \Bigg\},
    \label{eq:app_msr_Z1_HS}
\end{align}
where
\[
    \bm u^1
    \sim
    \mathcal N\left(
        \bm 0,\mathcal D(\bm C_0)
    \right).
\]
The \(\hat{\bm v}^1\) integral is now a delta functional, yielding
\begin{equation}
    v^1_\alpha
    =
    h^1_\alpha
    +
    u^1_\alpha
    +
    \sigma\epsilon_\alpha
    +
    \frac1B
    \sum_\beta
    [\mathcal D(\bm R_{0,2})]_{\alpha\beta}
    v^1_\beta.
    \label{eq:app_msr_v1_process}
\end{equation}

The identical steps for the random-feature site give
\begin{equation}
    v^3_\alpha
    =
    h^3_\alpha
    +
    u^3_\alpha
    +
    \frac1N
    \sum_\beta
    [\bm R_{2,4}]_{\alpha\beta}
    v^3_\beta,
    \qquad
    \bm u^3\sim\mathcal N(\bm0,\bm C_2).
    \label{eq:app_msr_v3_process}
\end{equation}

For each spectral mode, the two quadratic response-field terms introduce
independent Gaussian fields \(\bm u^2_k,\bm u^4_k\). Integrating
\(\hat{\bm v}^2_k,\hat{\bm v}^4_k\), and
\(\hat{\bm v}^0_k\) gives
\begin{align}
    \bm v^0_k
    &=
    v^0_{\mathrm{init},k}\bm 1
    -
    \eta\bm\Theta\bm v^4_k,
    \nonumber\\
    \bm v^2_k
    &=
    \bm h^2_k
    +
    \bm u^2_k
    +
    \lambda_k\bm R_1\bm v^0_k,
    \nonumber\\
    \bm v^4_k
    &=
    \bm h^4_k
    +
    \bm u^4_k
    +
    \bm R_3\bm v^2_k,
    \label{eq:app_msr_mode_process}
\end{align}
where we used \(\mathcal D(\bm R_1)=\bm R_1\). The Gaussian covariances are
\begin{align}
    \left\langle
        u^2_{k,\alpha}u^2_{\ell,\beta}
    \right\rangle
    &=
    \delta_{k\ell}
    \frac{\lambda_k}{B}
    [\mathcal D(\bm C_1)]_{\alpha\beta},
    &
    \left\langle
        u^4_{k,\alpha}u^4_{\ell,\beta}
    \right\rangle
    &=
    \delta_{k\ell}
    \frac1N
    [\bm C_3]_{\alpha\beta}.
    \label{eq:app_msr_mode_covariances}
\end{align}
The support statement \(\mathcal D(\bm R_1)=\bm R_1\) follows directly from
\eqref{eq:app_msr_v1_process}: its forcing and feedback connect only equal
within-epoch labels, so the response to \(h^1_\beta\) vanishes whenever
\(t_\alpha\neq t_\beta\).

Setting the kicks to zero in
\eqref{eq:app_msr_v1_process}--\eqref{eq:app_msr_mode_process} recovers
\eqref{eq:app_eff_v0_full}--\eqref{eq:app_eff_v4_full}, including the
fixed-order label-noise covariance. Differentiating these linear equations
recovers the response closure, while taking second moments recovers the
correlation closure in Appendix~\ref{app:vectorized_dmft}.

\subsection{Finite-Horizon Error Estimate}
\label{app:error_estimates}

We prove a loss approximation for Gaussian population features at fixed
feature dimension \(M\) and horizon \(S=T\mathcal E\). The comparison uses
population gradient descent as a common reference for the finite system
and the closed DMFT equations.

Write \(\bm C_0^{N,B}\) for the empirical residual correlation and
\(\bm C_0\) for the solution of Appendix~\ref{app:vectorized_dmft} at the
same \(N,B,M\):
\begin{equation}
    [\bm C_0^{N,B}]_{\alpha\beta}
    :=\sum_{k=1}^M\lambda_k
    v^{0,N,B}_{k,\alpha}v^{0,N,B}_{k,\beta}.
    \label{eq:app_finite_empirical_C0}
\end{equation}
Their test losses are
\begin{equation}
    \mathcal L_{N,B}(\alpha)
    =[\bm C_0^{N,B}]_{\alpha\alpha}+\sigma^2,
    \qquad
    \mathcal L_{\mathrm{DMFT}}(\alpha)
    =[\bm C_0]_{\alpha\alpha}+\sigma^2.
    \label{eq:app_finite_dmft_losses}
\end{equation}
The population trajectory and loss are
\begin{equation}
    \bar{\bm v}_s=(\bm I-\eta\bm\Lambda)^s\bm w^\star,
    \qquad
    \mathcal L_{\mathrm{pop}}(s)
    =\bar{\bm v}_s^\top\bm\Lambda\bar{\bm v}_s+\sigma^2,
    \quad 0\leq s<S.
    \label{eq:app_error_population_reference}
\end{equation}
Here \(s=nT+t\) corresponds to the one-based index \(\alpha=s+1\).

\begin{theorem}[Fixed-horizon DMFT approximation]
\label{thm:app_finite_horizon_error}
Fix \(M,T,\mathcal E\), a finite spectrum \(\lambda_k>0\), a deterministic
target \(\bm w^\star\), and \(\sigma\geq0\). Let the entries of \(\bm A\)
be independent standard Gaussians, the training features be independent
\(\mathcal N(\bm0,\bm\Lambda)\) vectors, and the label noises be independent
standard Gaussians, with these three collections mutually independent.
Use zero initialization, fixed-order minibatch reuse, and a fixed learning
rate satisfying \(0<\eta\|\bm\Lambda\|_{\mathrm{op}}\leq1-\rho\) for some
\(\rho\in(0,1)\). For \(\delta\in(0,1)\), set
\begin{equation}
    h=\log\frac{8(T+1)}\delta,
    \qquad
    \varepsilon=\sqrt{\frac{M+h}{N}}+
                \sqrt{\frac{M+h}{B}}.
    \label{eq:app_fixed_horizon_error_rate}
\end{equation}
There are constants \(K_{M,S}<\infty\) and \(c_0>0\), depending on the
fixed parameters but independent of \(N,B,\delta\), such that, whenever
\(\varepsilon\leq c_0\), with probability at least \(1-\delta\),
\begin{equation}
    \max_{\alpha\leq S}
    \left|\mathcal L_{N,B}(\alpha)
          -\mathcal L_{\mathrm{DMFT}}(\alpha)\right|
    \leq K_{M,S}\varepsilon.
    \label{eq:app_dmft_error_estimate}
\end{equation}
If
\begin{equation}
    \ell_S:=\min_{0\leq s<S}\mathcal L_{\mathrm{pop}}(s)>0,
    \label{eq:app_positive_loss_floor}
\end{equation}
the constants can be chosen so that, on the same event,
\begin{equation}
    \max_{\alpha\leq S}
    \frac{\left|\mathcal L_{N,B}(\alpha)
          -\mathcal L_{\mathrm{DMFT}}(\alpha)\right|}
         {\mathcal L_{\mathrm{DMFT}}(\alpha)}
    \leq \frac{2K_{M,S}}{\ell_S}\varepsilon.
    \label{eq:app_dmft_error_simple}
\end{equation}
\end{theorem}

The condition \(\ell_S>0\) holds when \(\sigma>0\), or when
\(\bm w^\star\neq\bm0\) under the stated positive spectrum and learning
rate. The constants depend on the fixed spectrum and horizon; extending
the bounds to growing cutoffs or horizons requires control of this dependence.

\paragraph{Proof outline.}
Concentration of the fixed feature and minibatch covariance matrices gives
a uniform comparison of the finite trajectory with population gradient
descent. Causality makes the finite-horizon DMFT solution polynomial in
\(1/N\) and \(1/B\); at \(1/N=1/B=0\), it equals the same population
trajectory. The triangle inequality gives the absolute loss bound.
The relative bound uses only the positive minimum of the reference loss.
Appendix~\ref{app:error_estimate_proof} gives the proof.

\paragraph{Stability of the causal equations.}
Let
\(\mathfrak Q=(\bm C_0,\bm C_1,\bm C_2,\bm C_3,\bm R_1,\bm R_3)\), and
define the absolute product norm by
\begin{equation}
    \|\mathfrak Q\|_{\max}
    :=\max\left\{\max_{0\leq j\leq3}\|\bm C_j\|_{\max},
                        \|\bm R_1\|_{\max},\|\bm R_3\|_{\max}\right\}.
    \label{eq:app_closure_absolute_norm}
\end{equation}
Each matrix norm on the right is the maximum absolute entry.

\begin{lemma}[Causal closure at a fixed horizon]
\label{lem:app_fixed_horizon_stability}
Fix \(M,S\) and the model parameters. Add residuals \(\bm r\) to the six
correlation and response equations in Appendix~\ref{app:vectorized_dmft},
keeping \(\bm H_k\) and \(\bm M_k\) defined by
\eqref{eq:app_H_boxed} and \eqref{eq:app_Mk_def}. Take the response
residuals to be strictly lower triangular and the correlation residuals
to be symmetric, with \(r_{0,11}=0\) for the fixed initial residual.
The perturbed equations have a unique causal solution, polynomial in
the residual entries and in \(1/N,1/B\). For sufficiently small residuals,
\begin{equation}
    \|\mathfrak Q(\bm r)-\mathfrak Q(\bm0)\|_{\max}
    \leq K_{M,S}\|\bm r\|_{\max}.
    \label{eq:app_causal_stability_bound}
\end{equation}
The local constant can be chosen uniformly for \(N,B\geq1\), with the
remaining parameters fixed.
\end{lemma}

The lemma establishes algebraic stability of the closed equations at
each fixed horizon through their triangular time dependence.

\section{DMFT for Ensembling}
\label{app:ensembling_theory}

Consider \(E\) models indexed by \(e=1,\ldots,E\). Their random feature matrices
\(\bm A^{(e)}\) are independent, while the dataset and fixed minibatch order are
shared.

For ensemble member \(e\), the residual field is
\begin{equation}
    \bm v^{0,(e)}_\alpha
    =
    \bm w^\star
    -
    \frac1N
    \bm A^{(e)\top}\bm w^{(e)}_\alpha .
    \label{eq:app_replica_residual}
\end{equation}
The ensemble residual is the average across members,
\begin{equation}
    \bar{\bm v}^0
    =
    \frac1E
    \sum_{e=1}^E
    \bm v^{0,(e)}.
    \label{eq:app_ensemble_residual}
\end{equation}
Its residual correlation is therefore
\begin{equation}
    \bar{\bm C}_0
    =
    \frac1{E^2}
    \sum_{e,e'}
    \bm C_0^{(e,e')},
    \qquad
    [\bm C_0^{(e,e')}]_{\alpha\beta}
    =
    \sum_k
    \lambda_k
    \left\langle
    v^{0,(e)}_{k,\alpha}
    v^{0,(e')}_{k,\beta}
    \right\rangle .
    \label{eq:app_ensemble_cov_def}
\end{equation}

\subsection{Within-Model and Cross-Model Correlations}
\label{app:replica_symmetry}

The replicated generating functional contains \(E^2\) correlation matrices.
Independence of the random feature matrices makes the response diagonal in the
ensemble indices, \(R_{e,e'}=\delta_{e,e'}R\). For the disorder-averaged
correlations, exchangeability of ensemble members gives the exact form
\begin{equation}
    \bm C_0^{(e,e')}
    =
    \begin{cases}
        \bm C_0^{\mathrm{self}}, & e=e',\\
        \bm C_0^{\mathrm{cross}}, & e\neq e',
    \end{cases}
    \label{eq:app_replica_symmetric_C0}
\end{equation}
where ``self'' denotes a within-model correlation and ``cross'' a correlation
between two distinct models. Averaging over the ensemble indices gives
\begin{equation}
    \bar{\bm C}_0
    =
    \frac1E\bm C_0^{\mathrm{self}}
    +
    \frac{E-1}{E}\bm C_0^{\mathrm{cross}}
    =
    \bm C_0^{\mathrm{cross}}
    +
    \frac1E
    \left(
        \bm C_0^{\mathrm{self}}
        -
        \bm C_0^{\mathrm{cross}}
    \right).
    \label{eq:app_ensemble_cov_decomp}
\end{equation}
The difference between within-model and cross-model correlations is reduced
by \(1/E\). For models that are independent conditional on the dataset, this
difference is the conditional covariance averaged over datasets. The
cross-model term is the second moment of the conditional mean residual.

\subsection{Effective Process for the Ensemble}
\label{app:replicated_effective_process}

Each ensemble member obeys the same effective single-site process as in
Appendix~\ref{app:vectorized_dmft}:
\begin{equation}
    \bm v^{0,(e)}_k
    =
    \bm H_k
    \left[
        w_k^\star\bm 1
        -
        \eta\bm\Theta
        \left(
            \bm u^{4,(e)}_k
            +
            \bm R_3\bm u^{2,(e)}_k
        \right)
    \right],
    \qquad
    \bm H_k
    =
    \left[
        \bm I
        +
        \eta\lambda_k\bm\Theta\bm R_3\bm R_1
    \right]^{-1}.
    \label{eq:app_replica_effective_process}
\end{equation}
The response operators \(\bm H_k,\bm R_1,\bm R_3\) are determined by the
single-model dynamics and therefore do not depend on \(E\).

The Gaussian fields generated by \(\bm A^{(e)}\) are independent across models:
\begin{equation}
    \left\langle
    \bm u^{4,(e)}_k
    \bm u^{4,(e')\top}_\ell
    \right\rangle
    =
    \delta_{k\ell}\delta_{ee'}
    \frac1N
    \bm C_3^{\mathrm{self}}.
    \label{eq:app_u4_replica_cov}
\end{equation}
The fields generated by the minibatch features remain correlated because the
data and their order are shared:
\begin{equation}
    \left\langle
    \bm u^{2,(e)}_k
    \bm u^{2,(e')\top}_\ell
    \right\rangle
    =
    \delta_{k\ell}
    \frac{\lambda_k}{B}
    \mathcal D(\bm C_1^{(e,e')}).
    \label{eq:app_u2_replica_cov}
\end{equation}
Since label noise is also part of the shared dataset, it contributes equally to
the within-model and cross-model correlations:
\begin{equation}
    \bm C_1^{\mathrm{self}}
    =
    \bm R_1
    \mathcal D\left(
        \bm C_0^{\mathrm{self}}
        +
        \sigma^2\bm 1\bm 1^\top
    \right)
    \bm R_1^\top,
    \qquad
    \bm C_1^{\mathrm{cross}}
    =
    \bm R_1
    \mathcal D\left(
        \bm C_0^{\mathrm{cross}}
        +
        \sigma^2\bm 1\bm 1^\top
    \right)
    \bm R_1^\top.
    \label{eq:app_C1_self_cross}
\end{equation}
If label noise were independently resampled across ensemble members, the
\(\sigma^2\)-term would appear only in the within-model correlation.

\subsection{Within-Model Correlations}
\label{app:self_sector}

The within-model equations are the single-model DMFT equations with
\(\bm C_i\) replaced by \(\bm C_i^{\mathrm{self}}\):
\begin{align}
    \bm C_0^{\mathrm{self}}
    &=
    \sum_k
    \lambda_k
    \bm H_k
    \left[
        (w_k^\star)^2\bm 1\bm 1^\top
        +
        \frac{\eta^2}{N}
        \bm\Theta\bm R_3\bm C_2^{\mathrm{self}}
        \bm R_3^\top\bm\Theta^\top
    \right]
    \bm H_k^\top
    \nonumber\\
    &\quad+
    \frac{\eta^2}{B}
    \sum_k\lambda_k^2
    \bm H_k\bm\Theta\bm R_3
    \mathcal D(\bm C_1^{\mathrm{self}})
    \bm R_3^\top\bm\Theta^\top\bm H_k^\top,
    \label{eq:app_C0_self}
    \\
    \bm C_2^{\mathrm{self}}
    &=
    \sum_k
    \lambda_k^2
    \bm R_1\bm H_k
    (w_k^\star)^2\bm 1\bm 1^\top
    \bm H_k^\top\bm R_1^\top
    \nonumber\\
    &\quad+
    \frac{\eta^2}{N}
    \sum_k\lambda_k^2
    \bm R_1\bm H_k\bm\Theta\bm R_3
    \bm C_2^{\mathrm{self}}
    \bm R_3^\top\bm\Theta^\top\bm H_k^\top\bm R_1^\top
    \nonumber\\
    &\quad+
    \frac1B
    \sum_k
    \lambda_k
    \bm M_k
    \mathcal D(\bm C_1^{\mathrm{self}})
    \bm M_k^\top .
    \label{eq:app_C2_self}
\end{align}

\subsection{Cross-Model Correlations}
\label{app:cross_sector}

For \(e\neq e'\), the Gaussian fields generated by the random feature matrices
have zero cross-model covariance:
\[
\left\langle
\bm u^{4,(e)}_k
\bm u^{4,(e')\top}_\ell
\right\rangle=0.
\]
Therefore the explicit \(1/N\) variance term is absent from the cross-model
correlation equation:
\begin{equation}
    \bm C_0^{\mathrm{cross}}
    =
    \sum_k
    \lambda_k
    \bm H_k
    \left[
        (w_k^\star)^2\bm 1\bm 1^\top
        +
        \frac{\eta^2\lambda_k}{B}
        \bm\Theta\bm R_3
        \mathcal D(\bm C_1^{\mathrm{cross}})
        \bm R_3^\top\bm\Theta^\top
    \right]
    \bm H_k^\top .
    \label{eq:app_C0_cross}
\end{equation}
The corresponding auxiliary cross-model correlation is
\begin{align}
    \bm C_2^{\mathrm{cross}}
    &=
    \sum_k
    \lambda_k^2
    \bm R_1\bm H_k
    (w_k^\star)^2\bm 1\bm 1^\top
    \bm H_k^\top\bm R_1^\top
    +
    \frac1B
    \sum_k
    \lambda_k
    \bm M_k
    \mathcal D(\bm C_1^{\mathrm{cross}})
    \bm M_k^\top .
    \label{eq:app_C2_cross}
\end{align}
In the present closure, \(\bm C_2^{\mathrm{cross}}\) does not feed back into
\(\bm C_0^{\mathrm{cross}}\), because fluctuations generated independently by
the random feature matrices have zero cross-model covariance.

\subsection{Variance Reduction by Ensembling}
\label{app:self_minus_cross}

Define
\[
    \Delta\bm C_0
    =
    \bm C_0^{\mathrm{self}}-\bm C_0^{\mathrm{cross}}.
\]
Subtracting the cross equation from the self equation gives
\begin{align}
    \Delta\bm C_0
    &=
    \sum_k
    \lambda_k
    \bm H_k
    \Bigg[
        \frac{\eta^2}{N}
        \bm\Theta\bm R_3\bm C_2^{\mathrm{self}}\bm R_3^\top\bm\Theta^\top
        +
        \frac{\eta^2\lambda_k}{B}
        \bm\Theta\bm R_3
        \mathcal D\!\left(
            \bm R_1\Delta\bm C_0\bm R_1^\top
        \right)
        \bm R_3^\top\bm\Theta^\top
    \Bigg]
    \bm H_k^\top.
    \label{eq:app_delta_C0}
\end{align}
The explicit shared label-noise term cancels on subtraction. Random-feature
fluctuations generate the conditional covariance \(\Delta\bm C_0\), and
minibatch feedback propagates it through training. Label noise can still
affect this covariance through \(\bm C_2^{\mathrm{self}}\).

\subsection{Ensemble loss}
\label{app:ensemble_loss_subsection}

Using \(\Delta\bm C_0\) in \eqref{eq:app_ensemble_cov_decomp} gives
\begin{equation}
    \bar{\bm C}_0
    =
    \bm C_0^{\mathrm{cross}}
    +
    \frac1E
    \Delta\bm C_0
    =
    \frac1E\bm C_0^{\mathrm{self}}
    +
    \frac{E-1}{E}\bm C_0^{\mathrm{cross}} .
    \label{eq:app_ensemble_C0}
\end{equation}
Thus the ensemble test loss at epoch-step \(\alpha=(n,t)\) is
\begin{equation}
    \mathcal L_{\mathrm{ens}}(n,t)
    =
    [\bar{\bm C}_0]_{\alpha\alpha}
    +
    \sigma^2.
    \label{eq:app_ensemble_loss}
\end{equation}
In effective time, this is \(\mathcal L(\tau,N,P,E)\) with
\(\tau=\eta(nT+t)\).

For \(E=1\),
\(\bar{\bm C}_0=\bm C_0^{\mathrm{self}}\), recovering the single-model DMFT. As
\(E\to\infty\), \(\bar{\bm C}_0\to\bm C_0^{\mathrm{cross}}\): variance
across random-feature realizations vanishes. The remaining
correlation includes the bias of the mean predictor and its dependence on
the shared data and labels.

Equation~\eqref{eq:app_ensemble_C0} is exact at the symmetric saddle.
Suppressing late-time growth by \(1/E\) further requires that the growing
contribution lie in \(\Delta\bm C_0\), as specified in
Appendix~\ref{app:double_laplace_late_growth}.

\paragraph{Independent shuffles.}
The equations above correspond to fixed-order training with the same minibatch
order for every ensemble member. If each member uses an independent data
shuffle, \(\mathcal D\) in \eqref{eq:app_C0_cross} should be replaced by
\(\mathcal D_{\mathrm{cross}}\), defined to measure the probability that two
members see the same example at the same update time.
For independent random shuffles, this overlap is heuristically \(O(1/T)\), so
the shared-minibatch feedback is weaker.

\subsection{Cross-Entropy of Averaged Logits}
\label{app:logit_ensemble_ce}

Fix a training checkpoint and condition on the training data, any shared
training randomness, and a validation example with label \(y\). Let
\(\bm z_1,\ldots,\bm z_E\in\mathbb R^K\) be conditionally independent,
identically distributed member logits, with mean \(\bm m\), covariance
\(\bm\Sigma\), and finite third centered moment. For
\(\bar{\bm z}_E=E^{-1}\sum_e\bm z_e\) and
\(\ell(\bm z,y)=\log\sum_{j=1}^K e^{z_j}-z_y\), Taylor expansion gives
\begin{equation}
 \mathbb E\ell(\bar{\bm z}_E,y)
 =\ell(\bm m,y)+\frac{1}{2E}
 \operatorname{tr}\!\left[
 (\diag(\bm p)-\bm p\bm p^\top)\bm\Sigma\right]
 +O(E^{-3/2}),
 \qquad \bm p=\operatorname{softmax}(\bm m).
 \label{eq:app_logit_ensemble_ce}
\end{equation}
Indeed, the centered average has mean zero and covariance
\(\bm\Sigma/E\). The third directional derivative of log-sum-exp obeys
\(\lvert D^3\ell(\bm z,y)[\bm u,\bm u,\bm u]\rvert
\le8\|\bm u\|_2^3\), since it is the third centered moment of
\(u_j\) under the softmax probabilities. The finite third moment and
independence give
\(\mathbb E\|\bar{\bm z}_E-\bm m\|_2^3=O(E^{-3/2})\), which controls
the Taylor remainder. Averaging over validation examples gives the same
expansion when the conditional third moments and leading loss are integrable.
The \(1/E\) coefficient is nonnegative and measures the effect of member
logit variation at a fixed checkpoint; it vanishes when that variation is
only a common additive shift of all logits.

\section{Batch Size in Finite-Step SGD}
\label{app:finite_step_batches}

Fix a realization of the dataset, random-feature matrix, and label noise,
and let the integer batch size \(B\) divide \(P\), with \(T=P/B\).
The same ordered partition is reused every epoch. The following identities
hold at any width \(N\). Define
\begin{equation}
\begin{aligned}
 \bm W&=\frac{\bm A^\top\bm A}{N},&
 \widehat{\bm\Sigma}_t&=\frac{\bm\Psi(t)^\top\bm\Psi(t)}B,\\
 \bm K_t&=\bm W\widehat{\bm\Sigma}_t,&
 \bm g_t&=\frac{\bm\Psi(t)^\top\bm\epsilon(t)}B.
\end{aligned}
 \label{eq:app_batch_update_matrices}
\end{equation}
Thus one update is
\(\bm v_n^0(t+1)=(\bm I-\eta\bm K_t)\bm v_n^0(t)
-\eta\sigma\bm W\bm g_t\).

\paragraph{Exact update over one epoch.}
Composing the \(T\) updates gives
\begin{equation}
\begin{aligned}
 \bm v_{n+1}^0(0)&=\bm U_\eta\bm v_n^0(0)+\bm d_\eta,\\
 \bm U_\eta&=(\bm I-\eta\bm K_{T-1})\cdots(\bm I-\eta\bm K_0),\\
 \bm d_\eta&=-\eta\sigma\sum_{t=0}^{T-1}
 (\bm I-\eta\bm K_{T-1})\cdots(\bm I-\eta\bm K_{t+1})
 \bm W\bm g_t.
\end{aligned}
 \label{eq:app_batch_epoch_map}
\end{equation}
The product in the last line is the identity when \(t=T-1\).
Since the matrices and noisy labels are fixed, the same affine update
applies every epoch, and
\[
 \bm v_n^0(0)=\bm U_\eta^n\bm w^\star
             +\sum_{j=0}^{n-1}\bm U_\eta^j\bm d_\eta.
\]
This expression also permits direct evaluation at late checkpoints by
composing epoch updates. A checkpoint after \(n\) complete epochs and
\(t\) further steps applies the first \(t\) batch updates to this state.

\paragraph{Comparison with gradient flow.}
For the noiseless compute analysis, set \(\sigma=0\). Write
\[
 h=\eta T,\qquad
 \bar{\bm K}=\frac1T\sum_{t=0}^{T-1}\bm K_t
            =\bm W\frac{\bm\Psi^\top\bm\Psi}{P},\qquad
 L=\max_{t<T}\|\bm K_t\|_{\mathrm{op}},
\]
where \(\bm\Psi\) contains all \(P\) training examples and \(h\) is the
effective duration of one epoch. The GF update over this duration is
\(e^{-h\bar{\bm K}}\). Expanding both updates to second order yields
\begin{equation}
\begin{aligned}
 \bm U_\eta-e^{-h\bar{\bm K}}
 &=\frac{\eta^2}{2}
   \left(\sum_{0\le i<j<T}[\bm K_j,\bm K_i]
                -\sum_{t=0}^{T-1}\bm K_t^2\right)+\bm R_\eta,\\
 \|\bm R_\eta\|_{\mathrm{op}}&\le\frac{(hL)^3}{3}e^{hL},
 \qquad [\bm K_j,\bm K_i]=\bm K_j\bm K_i-\bm K_i\bm K_j.
\end{aligned}
 \label{eq:app_batch_gf_correction}
\end{equation}
Indeed, the product has quadratic term
\(\eta^2\sum_{i<j}\bm K_j\bm K_i\), whereas the exponential has
\(\eta^2(\sum_t\bm K_t)^2/2\). Their difference gives the displayed
commutators and squared matrices. In operator norm, each tail from degree
three onward is bounded by
\((hL)^3e^{hL}/6\). This proves the remainder estimate without requiring
\(\bm W\) to be invertible or the \(\bm K_t\) to be symmetric.

The squared-matrix term is the finite-step correction already present in
full-batch gradient descent. The commutators record the order of distinct
batch updates; they vanish when the \(\bm K_t\) commute. At \(B=P\),
\(T=1\) and the leading difference is
\(-\eta^2\bar{\bm K}^2/2\). For a fixed dataset, changing its batch
partition leaves \(\bar{\bm K}\) unchanged but changes the two correction
terms. Equation~\eqref{eq:app_batch_gf_correction} is a small-step expansion
when \(hL\ll1\).

The comparison also controls a fixed effective training horizon.
The first-order expansions give
\(\|\bm U_\eta-e^{-h\bar{\bm K}}\|_{\mathrm{op}}
\le(hL)^2e^{hL}\), while both update norms are at most \(e^{hL}\).
Telescoping their \(n\)th powers therefore gives, at \(\tau=nh\),
\begin{equation}
 \|\bm U_\eta^n-e^{-\tau\bar{\bm K}}\|_{\mathrm{op}}
 \le n(hL)^2e^{nhL}
 =\tau hL^2e^{\tau L}.
 \label{eq:app_batch_finite_time_comparison}
\end{equation}
Multiplying by \(\bm w^\star\) bounds the residual difference. The
corresponding loss difference is controlled by
\[
 \left|\|\bm v\|_{\bm\Lambda}^2-\|\bm u\|_{\bm\Lambda}^2\right|
 \le\|\bm\Lambda\|_{\mathrm{op}}
 (\|\bm v\|_2+\|\bm u\|_2)\|\bm v-\bm u\|_2.
\]
At fixed \(P,N,M,B\), a fixed realization, and fixed \(\tau\),
the noiseless SGD loss at epoch boundaries approaches GF with error
\(O(h)\) as \(\eta\to0\). The displayed bound specifies the dependence
on the training horizon and batch matrices.

\paragraph{Batch size under a training budget.}
At \(s\) updates, the training cost of \(E\) members is proportional to
\(ENBs\), and the effective time is \(\tau=\eta s\). At fixed
\(N,E,B,\eta\), spending an unnormalized budget \(C_{\mathrm{SGD}}\)
therefore gives \(\tau=(\eta/B)C_{\mathrm{SGD}}/(EN)\), up to integer
checkpoint rounding. At fixed epoch count, the same relation is expressed
by \(h=\eta P/B\). Holding \(\eta/B\) fixed matches the GF duration per
epoch, while the finite-step corrections still depend on the partition
and learning rate. A finite-step compute frontier evaluates
\eqref{eq:app_batch_epoch_map} at all affordable checkpoints, with the
learning rate specified or optimized over the allowed choices.

For noiseless targets, independent feature draws give the exact ensemble decomposition
\(\mathcal L_E=C_{\mathrm{cross}}
+(C_{\mathrm{self}}-C_{\mathrm{cross}})/E\) at any common checkpoint.
Here changing \(B\) changes the self and cross correlations through the
epoch update, while averaging predictions reduces the conditional variance.

At fixed learning rate, a smaller batch therefore allows a longer effective
training time at the same compute. Increasing the learning rate with batch
size can compensate for this change in time, subject to the stability of
the discrete updates. Appendix~\ref{app:rf_batch} compares both learning-rate
conventions. It evaluates the full epoch update, including the finite-step
terms, over the tested batch sizes and learning rates.

\section{Formal Continuum Reductions: Noisy SGD and Gradient Flow}
\label{app:continuum_limits}

We use two small-learning-rate reductions. The first keeps the epoch index
discrete and takes a continuum limit within each epoch. The second averages
over infinitesimal passes and gives gradient flow on the empirical loss. These
are different formal limits of the discrete equations.
Appendix~\ref{app:finite_step_batches} gives a direct finite-step comparison
with GF for each integer batch size and fixed dataset.

\subsection{SGD limit: integer epochs, continuum within-epoch time}
\label{app:sgd_continuum_limit}

Consider the scaling
\begin{equation}
    \eta=\frac1T\to0,
    \qquad
    B=\frac{P}{T},
    \qquad
    P=BT \ \text{fixed},
    \qquad
    n=0,1,\ldots.
    \label{eq:app_sgd_continuum_scaling}
\end{equation}
Define the within-epoch time
\[
    u=\frac{t}{T}\in[0,1].
\]
The global effective time remains \(\tau=\eta(nT+t)=\eta T(n+u)\).
Under \(\eta=1/T\), this gives \(\tau=n+u\); the coordinate \(u\) resets
at each epoch.
The relation \(B=P/T\) is a continuum idealization, since minibatch size is
integer in the original dynamics. One epoch still processes \(P\) examples.
The drift is \(O(1)\) per epoch and the data-noise covariance is \(O(P^{-1})\),
so this limit retains SGD noise.

We use the combined index
\[
    \alpha=(n,u),
    \qquad
    n=0,1,\ldots,
    \qquad
    u\in[0,1].
\]
The continuum causal operator is the integral operator
\begin{equation}
    [\bm\Theta \bm f]_n(u)
    =
    \sum_{m<n}\int_0^1 du'\, f_m(u')
    +
    \int_0^u du'\, f_n(u').
    \label{eq:app_continuum_epoch_theta}
\end{equation}
Equivalently, its kernel is
\begin{equation}
    \Theta_{nm}(u,u')
    =
    \delta_{nm}\mathbf 1\{u>u'\}
    +
    \mathbf 1\{n>m\}.
    \label{eq:app_continuum_epoch_theta_kernel}
\end{equation}
The fixed-order reuse of the same within-epoch minibatch label is encoded by
\begin{equation}
    [\mathcal D(\bm C)]_{nm}(u,u')
    =
    C_{nm}(u,u')\,\delta(u-u').
    \label{eq:app_continuum_D_operator}
\end{equation}
Thus \(\mathcal D\) is local in \(u\) but can couple different epochs.

The corresponding single-site stochastic process is
\begin{align}
    v^0_{k,n}(u)
    &=
    w_k^\star
    -
    [\bm\Theta \bm v^4_k]_n(u),
    \label{eq:app_sgd_cont_v0}
    \\
    v^1_n(u)
    &=
    u^1_n(u)
    +
    \sigma\epsilon_n(u)
    +
    \frac1P
    [\mathcal D(\bm R_{0,2})\bm v^1]_n(u),
    \label{eq:app_sgd_cont_v1}
    \\
    v^2_{k,n}(u)
    &=
    u^2_{k,n}(u)
    +
    \lambda_k
    [\bm R_1\bm v^0_k]_n(u),
    \label{eq:app_sgd_cont_v2}
    \\
    v^3_n(u)
    &=
    u^3_n(u)
    +
    \frac1N
    [\bm R_{2,4}\bm v^3]_n(u),
    \label{eq:app_sgd_cont_v3}
    \\
    v^4_{k,n}(u)
    &=
    u^4_{k,n}(u)
    +
    [\bm R_3\bm v^2_k]_n(u).
    \label{eq:app_sgd_cont_v4}
\end{align}
Here all products denote integral-operator products over \((n,u)\). The
label field is reused across epochs,
\(\epsilon_n(u)\equiv\epsilon(u)\), and has covariance
\(\langle\epsilon_n(u)\epsilon_m(u')\rangle=\delta(u-u')\).
The
Gaussian fields have covariances
\begin{align}
    \left\langle
    u^1_n(u)u^1_m(u')
    \right\rangle
    &=
    [\mathcal D(\bm C_0)]_{nm}(u,u'),
    \label{eq:app_sgd_cont_u1}
    \\
    \left\langle
    u^2_{k,n}(u)u^2_{\ell,m}(u')
    \right\rangle
    &=
    \delta_{k\ell}
    \frac{\lambda_k}{P}
    [\mathcal D(\bm C_1)]_{nm}(u,u'),
    \label{eq:app_sgd_cont_u2}
    \\
    \left\langle
    u^3_n(u)u^3_m(u')
    \right\rangle
    &=
    C_{2;nm}(u,u'),
    \label{eq:app_sgd_cont_u3}
    \\
    \left\langle
    u^4_{k,n}(u)u^4_{\ell,m}(u')
    \right\rangle
    &=
    \delta_{k\ell}
    \frac1N
    C_{3;nm}(u,u').
    \label{eq:app_sgd_cont_u4}
\end{align}
Compared to the discrete-time equations, \(\eta\bm\Theta\) has become the
continuum integral operator \(\bm\Theta\), and the minibatch-variance prefactor
becomes \(1/P\).

The closed response equations are
\begin{align}
    \bm H_k
    &=
    \left[
        \bm I
        +
        \lambda_k\bm\Theta\bm R_3\bm R_1
    \right]^{-1},
    \qquad
    \bar{\bm H}
    =
    \sum_k\lambda_k\bm H_k,
    \label{eq:app_sgd_cont_H}
    \\
    \bm R_1
    &=
    \left[
        \bm I
        +
        \frac1P
        \mathcal D(\bar{\bm H}\bm\Theta\bm R_3)
    \right]^{-1},
    \label{eq:app_sgd_cont_R1}
    \\
    \bm R_3
    &=
    \left[
        \bm I
        +
        \frac1N
        \bm R_1\bar{\bm H}\bm\Theta
    \right]^{-1}.
    \label{eq:app_sgd_cont_R3}
\end{align}
Define
\begin{equation}
    \bm M_k
    =
    \bm I
    -
    \lambda_k
    \bm R_1\bm H_k\bm\Theta\bm R_3.
    \label{eq:app_sgd_cont_Mk}
\end{equation}
Then the closed correlation equations are
\begin{align}
    \bm C_1
    &=
    \bm R_1
    \mathcal D\left(
        \bm C_0+\sigma^2\bm 1\bm 1^\top
    \right)
    \bm R_1^\top,
    \qquad
    \bm C_3
    =
    \bm R_3\bm C_2\bm R_3^\top,
    \label{eq:app_sgd_cont_C1C3}
    \\
    \bm C_0
    &=
    \sum_k
    \lambda_k
    \bm H_k
    \Bigg[
        (w_k^\star)^2\bm 1\bm 1^\top
        +
        \frac1N
        \bm\Theta\bm R_3
        \bm C_2
        \bm R_3^\top\bm\Theta^\top
        \nonumber\\
    &\hspace{3.4cm}
        +
        \frac{\lambda_k}{P}
        \bm\Theta\bm R_3
        \mathcal D(\bm C_1)
        \bm R_3^\top\bm\Theta^\top
    \Bigg]
    \bm H_k^\top,
    \label{eq:app_sgd_cont_C0}
    \\
    \bm C_2
    &=
    \sum_k
    \lambda_k^2
    \bm R_1\bm H_k
    \left[
        (w_k^\star)^2\bm 1\bm 1^\top
        +
        \frac1N
        \bm\Theta\bm R_3
        \bm C_2
        \bm R_3^\top\bm\Theta^\top
    \right]
    \bm H_k^\top\bm R_1^\top
    \nonumber\\
    &\quad
    +
    \frac1P
    \sum_k
    \lambda_k
    \bm M_k
    \mathcal D(\bm C_1)
    \bm M_k^\top .
    \label{eq:app_sgd_cont_C2}
\end{align}

The epoch index remains explicit, so the response and correlation functions
need not be time-translation invariant. The projection \(\mathcal D\) retains
the same within-epoch label across epochs.

\subsection{Gradient-flow limit}
\label{app:gradient_flow_limit}

For the gradient-flow limit, let many infinitesimal passes occur over an
\(O(1)\) interval of macroscopic time. The auxiliary parameter
\(\mathcal N\) counts passes per unit time and is distinct from the model
width \(N\); both \(N\) and \(P\) remain fixed in this limit:
\begin{equation}
    \tau
    =
    \eta(nT+t),
    \qquad
    \eta=\frac{1}{T\mathcal N},
    \qquad
    N,P \ \text{fixed},
    \qquad
    \mathcal N\to\infty.
    \label{eq:app_gf_scaling}
\end{equation}
One epoch has duration
\[
    \Delta \tau
    =
    \eta T
    =
    \frac1{\mathcal N}.
\]
Over an \(O(1)\) interval, the algorithm makes \(\mathcal N\) infinitesimal
passes. The accumulated minibatch variance is \(O((P\mathcal N)^{-1})\) and
vanishes, while the label-noise vector remains fixed. The limit is gradient
flow on the noisy empirical loss:
\begin{align}
    \partial_\tau\bm v^0(\tau)
    &=
    -
    \left(
        \frac1N\bm A^\top\bm A
    \right)
    \frac1P
    \bm\Psi^\top
    \left(
        \bm\Psi\bm v^0(\tau)
        +
        \sigma\bm\epsilon
    \right)
    \nonumber\\
    &=
    -
    \left(
        \frac1N\bm A^\top\bm A
    \right)
    \left[
        \frac1P\bm\Psi^\top\bm\Psi\bm v^0(\tau)
        +
        \frac{\sigma}{P}\bm\Psi^\top\bm\epsilon
    \right].
    \label{eq:app_gf_microscopic_ode}
\end{align}
Here \(\bm\epsilon\in\mathbb R^P\) collects the label-noise variables drawn
once with the \(P\) training examples. It is fixed throughout gradient flow,
whereas the stochasticity from minibatch sampling vanishes in the limit.

The auxiliary fields are now defined using the full dataset:
\begin{align}
    \bm v^1(\tau)
    &=
    \bm\Psi\bm v^0(\tau)
    +
    \sigma\bm\epsilon,
    \label{eq:app_gf_v1_micro}
    \\
    \bm v^2(\tau)
    &=
    \frac1P
    \bm\Psi^\top\bm v^1(\tau),
    \label{eq:app_gf_v2_micro}
    \\
    \bm v^3(\tau)
    &=
    \bm A\bm v^2(\tau),
    \label{eq:app_gf_v3_micro}
    \\
    \bm v^4(\tau)
    &=
    \frac1N
    \bm A^\top\bm v^3(\tau),
    \label{eq:app_gf_v4_micro}
\end{align}
with
\[
    \partial_\tau\bm v^0(\tau)
    =
    -\bm v^4(\tau).
\]

The corresponding single-site stochastic process is
\begin{align}
    \partial_\tau v^0_k(\tau)
    &=
    -v^4_k(\tau),
    \label{eq:app_gf_v0_process}
    \\
    v^1(\tau)
    &=
    u^1(\tau)
    +
    \sigma\epsilon
    +
    \frac1P
    \int_0^\tau d\bar \tau\,
    R_{0,2}(\tau,\bar \tau)v^1(\bar \tau),
    \label{eq:app_gf_v1_process}
    \\
    v^2_k(\tau)
    &=
    u^2_k(\tau)
    +
    \lambda_k
    \int_0^\tau d\bar \tau\,
    R_1(\tau,\bar \tau)v^0_k(\bar \tau),
    \label{eq:app_gf_v2_process}
    \\
    v^3(\tau)
    &=
    u^3(\tau)
    +
    \frac1N
    \int_0^\tau d\bar \tau\,
    R_{2,4}(\tau,\bar \tau)v^3(\bar \tau),
    \label{eq:app_gf_v3_process}
    \\
    v^4_k(\tau)
    &=
    u^4_k(\tau)
    +
    \int_0^\tau d\bar \tau\,
    R_3(\tau,\bar \tau)v^2_k(\bar \tau).
    \label{eq:app_gf_v4_process}
\end{align}
The Gaussian fields have covariances
\begin{align}
    \left\langle u^1(\tau)u^1(\tau')\right\rangle
    &=
    C_0(\tau,\tau'),
    \label{eq:app_gf_u1_cov}
    \\
    \left\langle u^2_k(\tau)u^2_\ell(\tau')\right\rangle
    &=
    \delta_{k\ell}
    \frac{\lambda_k}{P}
    C_1(\tau,\tau'),
    \label{eq:app_gf_u2_cov}
    \\
    \left\langle u^3(\tau)u^3(\tau')\right\rangle
    &=
    C_2(\tau,\tau'),
    \label{eq:app_gf_u3_cov}
    \\
    \left\langle u^4_k(\tau)u^4_\ell(\tau')\right\rangle
    &=
    \delta_{k\ell}
    \frac1N
    C_3(\tau,\tau').
    \label{eq:app_gf_u4_cov}
\end{align}
The projection \(\mathcal D\) is absent because every infinitesimal update
averages over the full dataset. The data term therefore contains the full
two-time correlation \(C_1(\tau,\tau')\).

Let
\[
    \Theta(\tau,\tau')
    =
    \mathbf 1\{\tau>\tau'\}.
\]
In operator notation, the response equations are
\begin{align}
    \bm H_k
    &=
    \left[
        \bm I
        +
        \lambda_k\bm\Theta\bm R_3\bm R_1
    \right]^{-1},
    \qquad
    \bar{\bm H}
    =
    \sum_k\lambda_k\bm H_k,
    \label{eq:app_gf_H_time}
    \\
    \bm R_1
    &=
    \left[
        \bm I
        +
        \frac1P
        \bar{\bm H}\bm\Theta\bm R_3
    \right]^{-1},
    \label{eq:app_gf_R1_time}
    \\
    \bm R_3
    &=
    \left[
        \bm I
        +
        \frac1N
        \bm R_1\bar{\bm H}\bm\Theta
    \right]^{-1}.
    \label{eq:app_gf_R3_time}
\end{align}
The correlation equations are
\begin{align}
    \bm C_1
    &=
    \bm R_1
    \left(
        \bm C_0+\sigma^2\bm 1\bm 1^\top
    \right)
    \bm R_1^\top,
    \qquad
    \bm C_3
    =
    \bm R_3\bm C_2\bm R_3^\top,
    \label{eq:app_gf_C1C3_time}
    \\
    \bm C_0
    &=
    \sum_k
    \lambda_k
    \bm H_k
    \Bigg[
        (w_k^\star)^2\bm 1\bm 1^\top
        +
        \frac1N
        \bm\Theta\bm R_3
        \bm C_2
        \bm R_3^\top\bm\Theta^\top
        \nonumber\\
    &\hspace{3.3cm}
        +
        \frac{\lambda_k}{P}
        \bm\Theta\bm R_3
        \bm C_1
        \bm R_3^\top\bm\Theta^\top
    \Bigg]
    \bm H_k^\top,
    \label{eq:app_gf_C0_time}
    \\
    \bm C_2
    &=
    \sum_k
    \lambda_k^2
    \bm R_1\bm H_k
    \left[
        (w_k^\star)^2\bm 1\bm 1^\top
        +
        \frac1N
        \bm\Theta\bm R_3
        \bm C_2
        \bm R_3^\top\bm\Theta^\top
    \right]
    \bm H_k^\top\bm R_1^\top
    \nonumber\\
    &\quad
    +
    \frac1P
    \sum_k
    \lambda_k
    \bm M_k
    \bm C_1
    \bm M_k^\top,
    \label{eq:app_gf_C2_time}
\end{align}
where
\begin{equation}
    \bm M_k
    =
    \bm I
    -
    \lambda_k
    \bm R_1\bm H_k\bm\Theta\bm R_3.
    \label{eq:app_gf_Mk_time}
\end{equation}

For the stationary response solution,
\[
    R(\tau,\tau')
    =
    R(\tau-\tau').
\]
Using the Fourier convention
\[
    R(\tau-\tau')
    =
    \int\frac{d\omega}{2\pi}
    e^{i\omega(\tau-\tau')}
    R(\omega),
\]
we introduce the causal mode resolvent
\[
    H_k(\omega)
    :=
    [\bm\Theta\bm H_k](\omega).
\]
Thus \(H_k(\omega)\) below is not the literal Fourier transform of the
transfer operator \(\bm H_k\); the extra causal integration contributes a
factor \((i\omega)^{-1}\). With this convention, the response equations reduce
to
\begin{align}
    H_k(\omega)
    &=
    \frac{1}{
        i\omega+\lambda_kR_1(\omega)R_3(\omega)
    },
    \label{eq:app_gf_H_fourier}
    \\
    R_1(\omega)
    &=
    1
    -
    \frac1P
    \sum_k
    \frac{
        \lambda_kR_1(\omega)R_3(\omega)
    }{
        i\omega+\lambda_kR_1(\omega)R_3(\omega)
    },
    \label{eq:app_gf_R1_fourier}
    \\
    R_3(\omega)
    &=
    1
    -
    \frac1N
    \sum_k
    \frac{
        \lambda_kR_1(\omega)R_3(\omega)
    }{
        i\omega+\lambda_kR_1(\omega)R_3(\omega)
    }.
    \label{eq:app_gf_R3_fourier}
\end{align}

The correlations still depend on both times through the initial signal. Define
their two-frequency Fourier transform by
\begin{equation}
    C(\omega,\omega')
    =
    \int d\tau\,d\tau'\,
    e^{-i\omega \tau-i\omega'\tau'}
    C(\tau,\tau').
    \label{eq:app_gf_two_freq_transform}
\end{equation}
For \(\sigma=0\), the Fourier-domain correlation equations are
\begin{align}
    C_0(\omega,\omega')
    &=
    \sum_k
    \lambda_k
    H_k(\omega)H_k(\omega')
    \Bigg[
        (w_k^\star)^2
        +
        \frac1N C_3(\omega,\omega')
    \nonumber\\
    &\hspace{2.8cm}
        +
        \frac{\lambda_k}{P}
        R_3(\omega)R_3(\omega')C_1(\omega,\omega')
    \Bigg],
    \label{eq:app_gf_C0_fourier}
    \\
    C_1(\omega,\omega')
    &=
    R_1(\omega)R_1(\omega')
    C_0(\omega,\omega'),
    \label{eq:app_gf_C1_fourier}
    \\
    C_2(\omega,\omega')
    &=
    \frac{(i\omega)(i\omega')}{P}
    C_1(\omega,\omega')
    \sum_k
    \lambda_k
    H_k(\omega)H_k(\omega')
    \nonumber\\
    &\quad+
    R_1(\omega)R_1(\omega')
    \sum_k\lambda_k^2H_k(\omega)H_k(\omega')
    (w_k^\star)^2
    \nonumber\\
    &\quad+
    \frac{R_1(\omega)R_1(\omega')}{N}
    C_3(\omega,\omega')
    \sum_k\lambda_k^2H_k(\omega)H_k(\omega'),
    \label{eq:app_gf_C2_fourier}
    \\
    C_3(\omega,\omega')
    &=
    R_3(\omega)R_3(\omega')
    C_2(\omega,\omega').
    \label{eq:app_gf_C3_fourier}
\end{align}
These Fourier-domain equations use \(\sigma=0\). For \(\sigma>0\), the fixed
label-noise vector adds a zero-frequency source; the noisy analysis uses the
time-domain closure in
\eqref{eq:app_gf_C1C3_time}--\eqref{eq:app_gf_C2_time}.

The test and train losses are obtained from the diagonal:
\begin{equation}
    \mathcal L_{\mathrm{test}}(\tau)
    =
    C_0(\tau,\tau)+\sigma^2,
    \qquad
    \widehat{\mathcal L}_{\mathrm{train}}(\tau)
    =
    C_1(\tau,\tau).
    \label{eq:app_gf_losses}
\end{equation}

\section{Scaling Form for Interpolation-Induced Late-Time Growth}
\label{app:scaling_form_interpolation_growth}

We analyze the low-frequency equations underlying
\eqref{eq:main_late_growth}. An equal-frequency calculation first identifies
the relevant scales. The cross asymptotics and stopping laws are derived
separately in Appendices~\ref{app:finite_constant_target}
and~\ref{app:continuum_cross_stopping}.
Appendix~\ref{app:double_laplace_late_growth} then gives the equal-time result
under joint double-Laplace assumptions. We set \(\sigma=0\) throughout this
analysis, except in Appendix~\ref{app:label_noise_growth}, which derives the
additional contribution from label noise.

At the interpolation threshold \(N=P\), the finite-data and finite-width
responses agree. Write
\[
    R_1(\omega)\simeq R_3(\omega)=R(\omega),
    \qquad
    H_k(\omega)
    =
    \frac{1}{i\omega+\lambda_k R(\omega)^2}.
\]
The corresponding response equation is
\begin{equation}
    R(\omega)
    =
    1
    -
    \frac1P
    R(\omega)^2
    \sum_k
    \lambda_k H_k(\omega).
    \label{eq:interp_response_R}
\end{equation}
The frequency \(\omega\) is conjugate to the global effective time \(\tau\)
defined in Appendix~\ref{app:gradient_flow_limit}. At fixed learning rate, replacing effective
time by update count changes prefactors but leaves the exponents unchanged.

\paragraph{Fourier-domain correlation equations.}
Let \(C_0(\omega,\omega')\) and \(C_2(\omega,\omega')\) be the double Fourier
transforms of the residual and feature-gradient correlations.

Their coupled equations are
\begin{align}
    C_0(\omega,\omega')
    &=
    B_0(\omega,\omega')
    +
    \frac1P
    T_2(\omega,\omega')
    R(\omega)^2R(\omega')^2
    C_0(\omega,\omega')
    \nonumber\\
    &\quad+
    \frac1P
    R(\omega)R(\omega')
    T_1(\omega,\omega')
    C_2(\omega,\omega'),
    \label{eq:C0_lowfreq_system}
    \\
    C_2(\omega,\omega')
    &=
    B_2(\omega,\omega')
    +
    \frac1P
    (i\omega)(i\omega')
    T_1(\omega,\omega')
    R(\omega)R(\omega')
    C_0(\omega,\omega')
    \nonumber\\
    &\quad+
    \frac1P
    T_2(\omega,\omega')
    R(\omega)^2R(\omega')^2
    C_2(\omega,\omega').
    \label{eq:C2_lowfreq_system}
\end{align}
Here
\begin{align}
    B_0(\omega,\omega')
    &=
    \sum_k
    \lambda_k(w_k^\star)^2
    H_k(\omega)H_k(\omega'),
    \label{eq:B0_def}
    \\
    B_2(\omega,\omega')
    &=
    R(\omega)R(\omega')
    \sum_k
    \lambda_k^2(w_k^\star)^2
    H_k(\omega)H_k(\omega'),
    \label{eq:B2_def}
\end{align}
and
\begin{equation}
    T_q(\omega,\omega')
    =
    \sum_k
    \lambda_k^q
    H_k(\omega)H_k(\omega'),
    \qquad q\in\{1,2\}.
    \label{eq:Tq_def}
\end{equation}
Here \(B_0,B_2\) contain the target coefficients. The terms containing
\(T_2\) couple each correlation to itself, while those containing \(T_1\)
couple \(C_0\) and \(C_2\).

Equations~\eqref{eq:C0_lowfreq_system}--\eqref{eq:C2_lowfreq_system} can be
written as
\begin{equation}
    \bm G(\omega,\omega')
    \begin{pmatrix}
        C_0(\omega,\omega')\\
        C_2(\omega,\omega')
    \end{pmatrix}
    =
    \begin{pmatrix}
        B_0(\omega,\omega')\\
        B_2(\omega,\omega')
    \end{pmatrix},
    \label{eq:G_system}
\end{equation}
where
\begin{equation}
    \bm G(\omega,\omega')
    =
    \begin{pmatrix}
        1-\frac1P T_2 R(\omega)^2R(\omega')^2
        &
        -\frac1P R(\omega)R(\omega')T_1
        \\
        -\frac1P (i\omega)(i\omega')R(\omega)R(\omega')T_1
        &
        1-\frac1P T_2 R(\omega)^2R(\omega')^2
    \end{pmatrix}.
    \label{eq:G_matrix}
\end{equation}
For compactness, \(T_1,T_2\) in \eqref{eq:G_matrix} denote
\(T_1(\omega,\omega')\) and \(T_2(\omega,\omega')\). The factor
\((i\omega)(i\omega')\) appears only in the feedback from \(C_0\) to \(C_2\).

Solving \eqref{eq:G_system} for \(C_0\) introduces a denominator
\begin{equation}
    1-\Gamma(\omega,\omega')
    :=
    \det \bm G(\omega,\omega'),
    \label{eq:Gamma_det_def}
\end{equation}
The low-frequency behavior of this determinant controls the variance
amplification.

Equivalently, the residual correlation can be written as
\[
    C_0(\omega,\omega')
    =
    \frac{
        G_{22}(\omega,\omega')B_0(\omega,\omega')
        -
        G_{12}(\omega,\omega')B_2(\omega,\omega')
    }{
        1-\Gamma(\omega,\omega')
    }.
\]
The source terms contain the target spectrum, and \(1-\Gamma\) contains the
finite-data and finite-width feedback.

\subsection{Low-Frequency Scaling Heuristic}
\label{app:frequency_diagonal_heuristic}

\paragraph{Low-frequency analysis at equal frequencies.}
We first examine
\[
    \omega\sim\omega'\to0.
\]
After deriving the two-frequency system, set \(\omega'=\omega\) and define
\[
    T_q(\omega):=T_q(\omega,\omega),
\]
and
\begin{equation}
    \det\bm G(\omega)
    =
    \left(
        1-\frac1P T_2(\omega)R(\omega)^4
    \right)^2
    -
    \frac1{P^2}
    (i\omega)^2
    R(\omega)^4
    T_1(\omega)^2 .
    \label{eq:G_det_diag}
\end{equation}
Using
\[
    H_k(\omega)=\frac{1}{i\omega+\lambda_kR(\omega)^2},
\]
we have
\begin{equation}
    \lambda_kR(\omega)^2H_k(\omega)
    =
    1-i\omega H_k(\omega).
    \label{eq:lambda_RH_identity}
\end{equation}
Therefore,
\begin{align}
    \frac1P R(\omega)^4T_2(\omega)
    &=
    \frac1P
    R(\omega)^2
    \sum_k
    \lambda_kH_k(\omega)
    \left[
        \lambda_kR(\omega)^2H_k(\omega)
    \right]
    \nonumber\\
    &=
    \frac1P
    R(\omega)^2
    \sum_k
    \lambda_kH_k(\omega)
    -
    \frac1P
    i\omega R(\omega)^2
    \sum_k
    \lambda_kH_k(\omega)^2.
    \label{eq:T2_reduction_1}
\end{align}
By the response equation~\eqref{eq:interp_response_R},
\[
    \frac1P
    R(\omega)^2
    \sum_k
    \lambda_kH_k(\omega)
    =
    1-R(\omega).
\]
Hence
\begin{equation}
    \frac1P R(\omega)^4T_2(\omega)
    =
    1-R(\omega)
    -
    \frac1P i\omega R(\omega)^2T_1(\omega).
    \label{eq:T2_reduction}
\end{equation}
Substituting \eqref{eq:T2_reduction} into \eqref{eq:G_det_diag} gives
\begin{align}
    1-\Gamma(\omega,\omega)
    &=
    \left[
        R(\omega)
        +
        \frac1P i\omega R(\omega)^2T_1(\omega)
    \right]^2
    -
    \frac1{P^2}
    (i\omega)^2
    R(\omega)^4
    T_1(\omega)^2
    \nonumber\\
    &=
    R(\omega)^2
    +
    \frac2P
    i\omega R(\omega)^3T_1(\omega).
    \label{eq:Gamma_reduced}
\end{align}

\subsubsection{Power-law bias and variance scaling}

\paragraph{Power-law source and capacity estimates.}
The DMFT equations above hold for arbitrary spectra \(\{\lambda_k\}\) and target
coefficients \(\{w_k^\star\}\). To obtain explicit exponents, we specialize to the
standard power-law setting used in random-feature learning-curve analyses
\citep{bordelon2020spectrum,canatar2021spectral,simon2021eigenlearning,bordelon2024dynamical}:
\begin{equation}
    \lambda_k\sim k^{-b},
    \qquad
    \lambda_k(w_k^\star)^2\sim k^{-a},
    \qquad
    b>1,\quad a>1.
    \label{eq:power_law_spectrum_target}
\end{equation}
Here \(b\) controls spectral decay and \(a\) controls target power. The
conditions \(a,b>1\) make the relevant tails summable. We use \(\asymp\)
when multiplicative constants are suppressed; frequency-domain order
estimates below refer to magnitudes.

\paragraph{Scaling away from the interpolation threshold.}
Let
\[
    m_-:=\min\{N,P\},
    \qquad
    m_+:=\max\{N,P\},
    \qquad
    Q(\omega):=R_1(\omega)R_3(\omega).
\]
The regular low-frequency branch of
\eqref{eq:app_gf_R1_fourier}--\eqref{eq:app_gf_R3_fourier} satisfies
\begin{equation}
    Q(\omega)
    \sim
    i\omega r^2,
    \qquad
    1
    =
    \frac1{m_-}
    \sum_k
    \frac{\lambda_k r^2}{1+\lambda_k r^2}.
    \label{eq:R_lowfreq_noneq}
\end{equation}
Under the power-law assumption, the spectral sum is of order
\((r^2)^{1/b}\), and hence
\begin{equation}
    r^2
    \asymp
    m_-^b.
    \label{eq:r_noneq_scaling}
\end{equation}
For \(a-1<2b\), substitution into the asymptotic bias
gives the residual-bias scale
\begin{equation}
    \mathcal B_\infty(N,P)
    :=
    \sum_k
    \frac{
        \lambda_k(w_k^\star)^2
    }{
        (1+\lambda_k r^2)^2
    }
    \asymp
    m_-^{-(a-1)}.
    \label{eq:bias_noneq_scaling}
\end{equation}

For the off-threshold variance, write
\[
    H_k(\omega)
    =
    \frac1{i\omega+\lambda_kQ(\omega)}
\]
and \(T_q\) defined in \eqref{eq:Tq_def}, the two-frequency feedback matrix is
\begin{equation}
    \bm G_{N,P}(\omega,\omega')
    =
    \begin{pmatrix}
        1-\dfrac{Q(\omega)Q(\omega')}{P}T_2
        &
        -\dfrac{R_3(\omega)R_3(\omega')}{N}T_1
        \\[8pt]
        -\dfrac{(i\omega)(i\omega')
        R_1(\omega)R_1(\omega')}{P}T_1
        &
        1-\dfrac{Q(\omega)Q(\omega')}{N}T_2
    \end{pmatrix},
    \label{eq:G_matrix_noneq}
\end{equation}
where \(T_q=T_q(\omega,\omega')\). Thus
\begin{align}
    1-\Gamma_{N,P}(\omega,\omega')
    &:=
    \det\bm G_{N,P}(\omega,\omega')
    \nonumber\\
    &=
    \left(
        1-\frac{Q(\omega)Q(\omega')}{P}T_2
    \right)
    \left(
        1-\frac{Q(\omega)Q(\omega')}{N}T_2
    \right)
    \nonumber\\
    &\quad-
    \frac{
        (i\omega)(i\omega')Q(\omega)Q(\omega')
    }{
        NP
    }
    T_1^2.
    \label{eq:Gamma_noneq_det}
\end{align}

To evaluate the low-frequency limit, define
\[
    \chi_k
    :=
    \frac{\lambda_k r^2}{1+\lambda_k r^2},
    \qquad
    A_2
    :=
    \sum_k\chi_k^2.
\]
Equation~\eqref{eq:R_lowfreq_noneq} implies
\(\sum_k\chi_k=m_-\). Moreover,
\[
    Q(\omega)Q(\omega')T_2(\omega,\omega')
    \longrightarrow
    A_2,
\]
and
\[
    (i\omega)(i\omega')Q(\omega)Q(\omega')
    T_1(\omega,\omega')^2
    \longrightarrow
    (m_- - A_2)^2.
\]
Substituting these limits into \eqref{eq:Gamma_noneq_det} gives
\begin{equation}
    \lim_{\omega,\omega'\to0}
    \left[
        1-\Gamma_{N,P}(\omega,\omega')
    \right]
    =
    \frac{m_+ - m_-}{m_+}
    \left(
        1-\frac{A_2}{m_-}
    \right).
    \label{eq:Gamma_noneq_limit}
\end{equation}
For a nondegenerate spectrum, the right-hand side is positive when \(N\neq P\).
The determinant therefore vanishes at zero frequency only at \(N=P\).

\paragraph{Low-frequency response at the interpolation threshold.}
At \(N=P\), \(R_1=R_3=R\), and the low-frequency solution has the form
\begin{equation}
    R(\omega)^2
    =
    i\omega r^2+o(\omega),
    \qquad \omega\to0,
    \label{eq:R_lowfreq}
\end{equation}
where \(r\) is determined by
\begin{equation}
    1
    =
    \frac1P
    \sum_k
    \frac{\lambda_kr^2}{1+\lambda_kr^2}.
    \label{eq:r_equation}
\end{equation}
Substitution into \eqref{eq:interp_response_R} gives
\eqref{eq:r_equation} at leading order.

Let \(k_\star\) be the cutoff satisfying
\[
    \lambda_{k_\star}r^2\asymp1.
\]
Using \(\lambda_k\sim k^{-b}\) from \eqref{eq:power_law_spectrum_target}, this gives
\[
    k_\star\asymp r^{2/b}.
\]
The sum in \eqref{eq:r_equation} is of order \(k_\star\), so \(k_\star\asymp P\).
Therefore
\begin{equation}
    r^2\asymp P^b,
    \qquad
    r\asymp P^{b/2}.
    \label{eq:r_scaling}
\end{equation}

Next,
\begin{align}
    T_1(\omega)
    &=
    \sum_k
    \frac{\lambda_k}
    {
    \left(
        i\omega+\lambda_kR(\omega)^2
    \right)^2
    }
    =
    \frac1{(i\omega)^2}
    \sum_k
    \frac{\lambda_k}
    {
    \left(
        1+\lambda_kr^2
    \right)^2
    }
    =
    \frac{1}{(i\omega)^2}\mathcal T_1(P).
    \label{eq:T1_scaling_start}
\end{align}
Using the same cutoff \(k_\star\asymp P\),
\begin{align}
    \mathcal T_1(P)
    =
    \sum_k
    \frac{\lambda_k}
    {(1+\lambda_kr^2)^2}
    \nonumber&\asymp
    r^{-4}\sum_{k\le P}\lambda_k^{-1}
    +
    \sum_{k>P}\lambda_k
    \nonumber\\
    &\asymp
    P^{-2b}
    \sum_{k\le P}k^b
    +
    \sum_{k>P}k^{-b}
    \nonumber\\
    &\asymp
    P^{-2b}\frac{P^{b+1}}{b+1}
    +
    \frac{P^{1-b}}{b-1}
    \asymp
    P^{1-b},
    \qquad b>1.
    \label{eq:T1_prefactor_scaling}
\end{align}
Thus
\begin{equation}
    T_1(\omega)
    \asymp
    (i\omega)^{-2}P^{1-b}.
    \label{eq:T1_final_scaling}
\end{equation}

Substituting \eqref{eq:R_lowfreq}, \eqref{eq:r_scaling}, and
\eqref{eq:T1_final_scaling} into \eqref{eq:Gamma_reduced} gives
\begin{align}
    1-\Gamma(\omega,\omega)
    &=
    R(\omega)^2
    +
    \frac2P
    i\omega R(\omega)^3T_1(\omega)
    \nonumber\\
    &\asymp
    P^b(i\omega)
    +
    \frac2P
    (i\omega)
    (i\omega)^{3/2}P^{3b/2}
    (i\omega)^{-2}P^{1-b}
    \nonumber\\
    &\asymp
    P^b(i\omega)
    +
    2P^{b/2}(i\omega)^{1/2}.
    \label{eq:Gamma_scaling}
\end{align}

The ratio of the analytic and square-root terms is
\[
    \frac{P^b|\omega|}
    {P^{b/2}|\omega|^{1/2}}
    =
    P^{b/2}|\omega|^{1/2}.
\]
Thus the square-root term dominates when
\begin{equation}
    P^{b/2}|\omega|^{1/2}\ll1,
    \qquad\text{equivalently}\qquad
    |\omega|\ll P^{-b}.
    \label{eq:onset_frequency_condition}
\end{equation}
Since \(|\omega|\asymp \tau^{-1}\) at time \(\tau\), the square-root term appears for
\begin{equation}
    \tau\gtrsim P^b.
    \label{eq:onset_time_condition}
\end{equation}
For \(\tau\ll P^b\), the analytic term \(P^b(i\omega)\) dominates and gives the
usual mode-wise relaxation.

\paragraph{Heuristic equal-time inversion.}
For a generic two-time correlation, the equal-time inverse transform is
\begin{equation}
    C(\tau,\tau)
    =
    \int\frac{d\omega\,d\omega'}{(2\pi)^2}
    e^{i\tau(\omega+\omega')}
    \mathcal C(\omega,\omega').
    \label{eq:equal_time_double_inverse}
\end{equation}
Introduce the total and relative frequencies
\[
    \Omega=\omega+\omega',
    \qquad
    \nu=\frac{\omega-\omega'}{2}.
\]
Then
\begin{equation}
    C(\tau,\tau)
    =
    \int\frac{d\Omega}{2\pi}e^{i\tau\Omega}
    \overline{\mathcal C}(\Omega),
    \qquad
    \overline{\mathcal C}(\Omega)
    =
    \int\frac{d\nu}{2\pi}
    \mathcal C\left(
        \frac{\Omega}{2}+\nu,
        \frac{\Omega}{2}-\nu
    \right).
    \label{eq:equal_time_total_relative_frequency}
\end{equation}
Large time selects small total frequency \(\Omega\), but does not set
\(\nu=0\). The equal-frequency approximation therefore requires an additional
assumption. Suppose the contours can be deformed to a unique dominant saddle of
\[
    \Phi_\tau(\omega,\omega')
    =
    i\tau(\omega+\omega')+\log\mathcal C(\omega,\omega').
\]
Its saddle equations are
\begin{equation}
    i\tau+\partial_\omega\log\mathcal C(\omega,\omega')=0,
    \qquad
    i\tau+\partial_{\omega'}\log\mathcal C(\omega,\omega')=0.
    \label{eq:frequency_saddle_equations}
\end{equation}
Exchange symmetry maps an off-diagonal saddle to a second saddle with the same
value. Uniqueness therefore places the saddle at
\(\omega=\omega'=\omega_\star\). The diagonal exponent then determines the
large-time exponent only if the rescaled \(\nu\)-integral is finite and nonzero,
including its Jacobian.

The joint inversion assumptions in
Appendix~\ref{app:double_laplace_late_growth} provide the control away from
equal frequencies that this heuristic requires.

\paragraph{Time bottleneck.}
For \(\tau\ll P^b\), each mode relaxes independently to leading order. Since
\[
    \lambda_k\sim k^{-b},
\]
the \(k\)-th mode relaxes on the timescale
\[
    \tau_k\asymp \lambda_k^{-1}\asymp k^b.
\]
Hence, by time \(\tau\), only modes with
\[
    k\lesssim \tau^{1/b}
\]
have relaxed appreciably. The higher modes retain most of their initial error.

Using the power-law task assumption
\[
    \lambda_k(w_k^\star)^2\sim k^{-a},
\]
the residual bias is dominated by these higher modes:
\[
    \mathcal L_{\mathrm{bias}}(\tau)
    \asymp
    \sum_{k\gtrsim \tau^{1/b}}
    \lambda_k(w_k^\star)^2
    \asymp
    \sum_{k\gtrsim \tau^{1/b}}
    k^{-a}.
\]
Approximating the sum by an integral gives
\[
    \mathcal L_{\mathrm{bias}}(\tau)
    \asymp
    \int_{\tau^{1/b}}^\infty k^{-a}\,dk
    \asymp
    \tau^{-(a-1)/b}.
\]

For \(\tau\gtrsim P^b\), the square-root term instead gives
\begin{equation}
    1-\Gamma(\omega,\omega)
    \asymp
    P^{b/2}(i\omega)^{1/2}.
    \label{eq:Gamma_sqrt_growth}
\end{equation}
Under the equal-frequency heuristic, the inverse determinant contributes
\(\tau^{1/2}P^{-b/2}\).

\paragraph{Signal source.}
The source \(B_0\) alone has the low-frequency form
\begin{equation}
    B_0(\omega,\omega)
    \sim
    \frac{\mathcal B_0(P)}{(i\omega)^2},
    \qquad
    \mathcal B_0(P)
    :=\sum_k
    \frac{\lambda_k(w_k^\star)^2}{(1+\lambda_k r^2)^2}.
    \label{eq:B0_lowfreq}
\end{equation}
Splitting the sum at \(k\asymp P\) gives
\begin{equation}
    \mathcal B_0(P)
    \asymp
    P^{-2b}\sum_{k\le P}k^{2b-a}
    +\sum_{k>P}k^{-a}.
    \label{eq:B0_source_decomp}
\end{equation}
Consequently,
\begin{equation}
    \mathcal B_0(P)\asymp P^{-(a-1)},
    \qquad 1<a<2b+1.
    \label{eq:B0_source_scaling}
\end{equation}
At \(a=2b+1\), this estimate is \(P^{-2b}\log P\); for larger \(a\), it
is \(P^{-2b}\). These are estimates of \(B_0\), not of the full numerator
\(G_{22}B_0-G_{12}B_2\).

Write \(r_P=r\) to display the dependence on \(P\), and define
\begin{equation}
    q_P
    :=\frac{r_P^2}{P}\sum_k
    \frac{\lambda_k}{(1+\lambda_k r_P^2)^2},
    \qquad
    \mathcal B(P)
    :=\sum_k\frac{\lambda_k(w_k^\star)^2}{1+\lambda_k r_P^2}.
    \label{eq:app_full_signal_source}
\end{equation}
The response equation implies
\[
    q_P=1-\frac1P\sum_k
    \left(\frac{\lambda_k r_P^2}{1+\lambda_k r_P^2}\right)^2,
\]
and the power-law estimates give \(q_P\asymp1\). Under the low-frequency
response expansion, \(G_{22}\sim q_P\) and
\(-G_{12}\sim q_P/(i\omega)\) at equal frequencies. Thus
\begin{align}
    G_{22}B_0-G_{12}B_2
    &\sim\frac{q_P}{(i\omega)^2}
    \sum_k\frac{\lambda_k(w_k^\star)^2
        +r_P^2\lambda_k^2(w_k^\star)^2}
        {(1+\lambda_k r_P^2)^2}
    \nonumber\\
    &=\frac{q_P \mathcal B(P)}{(i\omega)^2}.
    \label{eq:app_full_signal_numerator_diagonal}
\end{align}
The two source terms add and leave a single power of the denominator.
Its dependence on \(P\) is
\begin{equation}
    \mathcal B(P)\asymp
    \begin{cases}
        P^{-(a-1)}, & 1<a<b+1,\\
        P^{-b}\log P, & a=b+1,\\
        P^{-b}, & a>b+1.
    \end{cases}
    \label{eq:app_full_signal_source_scaling}
\end{equation}
Indeed, its low-index contribution is
\(P^{-b}\sum_{k\le P}k^{b-a}\), while its tail is
\(\sum_{k>P}k^{-a}\). The late-time and stopping laws below use the first
branch, \(1<a<b+1\).

\paragraph{Resulting scaling form.}
Combining \eqref{eq:Gamma_sqrt_growth} and
\eqref{eq:app_full_signal_source_scaling}, for \(N=P\), \(1<a<b+1\), and
\(\tau\gg P^b\), the equal-frequency estimate suggests a test-loss contribution
of order
\begin{equation}
    \frac{1}{E}\,
    P^{-b/2}\,\mathcal B(P)\,\tau^{1/2}
    \;\asymp\;
    \frac{1}{E}\,
    P^{-(a-1+b/2)}\,
    \tau^{1/2}.
    \label{eq:interp_growth_source_scaled}
\end{equation}
The factor \(E^{-1}\) assumes that the growing contribution lies in
\(\bm C_0^{\mathrm{self}}-\bm C_0^{\mathrm{cross}}\), with the regularity
condition on the cross-model correlation stated below.

The early-time contribution \(\tau^{-(a-1)/b}\) above applies for
\(\tau\ll P^b\). To determine a late-time minimum, one instead needs the
cross-model correlation and its derivative in that window. The exact
noiseless ensemble loss is
\begin{equation}
    \mathcal L(\tau,N,P,E)
    =C_{\mathrm{cross}}(\tau)
    +\frac{C_{\mathrm{self}}(\tau)-C_{\mathrm{cross}}(\tau)}E,
    \qquad N=P.
    \label{eq:interp_local_scaling_law}
\end{equation}
Appendix~\ref{app:continuum_cross_stopping} derives the cross and self
curves in the pure-power continuum model and uses their derivatives to
determine optimal stopping. Its self-growth coefficient agrees with
the calculation above.

With \(T=P/B\) minibatches per epoch and \(\mathcal E\) epochs, the
available effective time is \(\eta T\mathcal E\). Observing the late-time
regime requires \(\eta T\mathcal E\gtrsim P^b\); this kinematic condition
is separate from the ensemble-dependent stopping balance.

\subsection{Double-Laplace Analysis of Equal-Time Correlations}
\label{app:double_laplace_late_growth}

To recover the equal-time correlation, we analyze both frequency variables
using a one-sided double Laplace transform.
For \(\Re z,\Re z'>0\), define
\begin{equation}
    \widehat C_0(z,z')
    =
    \int_0^\infty\int_0^\infty
    e^{-z\tau-z'\tau'}C_0(\tau,\tau')\,d\tau\,d\tau'.
    \label{eq:app_double_laplace_C0}
\end{equation}
The response and correlation equations are obtained from
\eqref{eq:app_gf_H_fourier}--\eqref{eq:app_gf_C3_fourier} by replacing
\(i\omega\) and \(i\omega'\) with \(z\) and \(z'\), respectively. Let
\begin{equation}
    D_P(z,z')
    :=
    1-\Gamma_P(z,z')
    =
    \det\bm G_P(z,z')
    \label{eq:app_laplace_denominator}
\end{equation}
and define the full source numerator
\begin{equation}
    \mathcal N_P(z,z')
    :=
    G_{22,P}(z,z')B_{0,P}(z,z')
    -
    G_{12,P}(z,z')B_{2,P}(z,z').
    \label{eq:app_laplace_numerator}
\end{equation}
Then the exact transformed correlation is
\begin{equation}
    \widehat C_0(z,z')
    =
    \frac{\mathcal N_P(z,z')}{D_P(z,z')}.
    \label{eq:app_laplace_C0_ratio}
\end{equation}
Both terms in the full numerator contribute at leading order. Their sum is
proportional to \(\mathcal B(P)\) in \eqref{eq:app_full_signal_source}; its
denominator has one power of \(1+\lambda_k r_P^2\).

\paragraph{Joint low-frequency estimate.}
The next lemma states the required joint expansion in \(z,z'\).

\begin{lemma}[Joint low-frequency expansion]
\label{lem:app_joint_critical_expansion}
Suppose \(N=P\), \(\sigma=0\),
\(\lambda_k\sim c_\lambda k^{-b}\), and
\(\lambda_k(w_k^\star)^2\sim c_w k^{-a}\), where
\(b>1\) and \(1<a<b+1\). Let the square root denote the principal branch in
\(\Re z>0\). In the joint low-frequency region relevant to the inverse Laplace
contours, suppose the stationary DMFT solution satisfies, uniformly as
\(P\to\infty\),
\begin{align}
    R_P(z)^2
    &=
    z r_P^2\bigl(1+o(1)\bigr),
    &
    r_P^2
    &\asymp
    P^b,
    \label{eq:app_joint_response_expansion}
    \\
    D_P(z,z')
    &=
    d_P\bigl(\sqrt z+\sqrt{z'}\bigr)
    \bigl(1+o(1)\bigr),
    &
    d_P
    &\asymp
    P^{b/2},
    \label{eq:app_joint_denominator_expansion}
    \\
    \mathcal N_P(z,z')
    &=
    \frac{n_P}{zz'}\bigl(1+o(1)\bigr),
    &
    n_P
    &:=q_P \mathcal B(P)\asymp P^{-(a-1)}.
    \label{eq:app_joint_numerator_expansion}
\end{align}
Assume in addition that the remainders in
\eqref{eq:app_joint_denominator_expansion}--\eqref{eq:app_joint_numerator_expansion}
are dominated on the inversion contours by an integrable majorant and that the
complementary-frequency contribution is
\(o(P^{-(a-1+b/2)}\tau^{1/2})\) in the late-time scaling window. Then
\begin{equation}
    \widehat C_{0,\mathrm{grow}}(z,z')
    =
    K_P
    \frac{1+o(1)}
    {zz'(\sqrt z+\sqrt{z'})},
    \qquad
    K_P:=\frac{n_P}{d_P}
    \asymp P^{-(a-1+b/2)}.
    \label{eq:app_joint_C0_singularity}
\end{equation}
\end{lemma}

To obtain the coefficient in \eqref{eq:app_joint_numerator_expansion}, insert
the response expansion into the spectral sums. In the same joint region,
\[
    G_{22,P}\sim q_P,
    \qquad
    -G_{12,P}\sim\frac{q_P}{\sqrt{zz'}},
    \qquad
    B_{2,P}\sim\frac{r_P^2}{\sqrt{zz'}}
    \sum_k\frac{\lambda_k^2(w_k^\star)^2}{(1+\lambda_k r_P^2)^2}.
\]
Combining these terms with \(B_{0,P}\) gives
\(\mathcal N_P\sim q_P \mathcal B(P)/(zz')\). The spectral estimates therefore give
\(r_P^2\asymp P^b\) and \(n_P\asymp P^{-(a-1)}\) in the stated range.
The uniform joint expansions, including the determinant expansion after
its constant term cancels at \(N=P\), and the inversion bounds remain
assumptions of the lemma. Establishing them for the full DMFT requires
control on both Bromwich contours beyond the equal-frequency calculation.

\begin{theorem}[Conditional late-time critical growth]
\label{thm:app_late_time_critical_growth}
Under the conditions of Lemma~\ref{lem:app_joint_critical_expansion}, the
singular contribution to the equal-time residual correlation satisfies
\begin{equation}
    C_{0,\mathrm{grow}}(\tau,\tau)
    =
    \frac{2(2-\sqrt 2)}{\sqrt\pi}
    K_P \tau^{1/2}
    +
    o\!\left(K_P\tau^{1/2}\right).
    \label{eq:app_late_time_critical_theorem}
\end{equation}
Consequently,
\begin{equation}
    C_{0,\mathrm{grow}}(\tau,\tau)
    \asymp
    P^{-(a-1+b/2)}\tau^{1/2}.
    \label{eq:app_late_time_critical_scaling}
\end{equation}
\end{theorem}

\begin{proof}
For \(\Re z,\Re z'>0\), use the integral representation
\begin{equation}
    \frac1{\sqrt z+\sqrt{z'}}
    =
    \int_0^\infty
    e^{-u\sqrt z}e^{-u\sqrt{z'}}\,du.
    \label{eq:app_sqrt_kernel_representation}
\end{equation}
The one-dimensional inverse Laplace identity
\begin{equation}
    \mathcal L^{-1}_{z\to \tau}
    \left[
        \frac{e^{-u\sqrt z}}{z}
    \right]
    =
    \operatorname{erfc}
    \left(
        \frac{u}{2\sqrt \tau}
    \right)
    \label{eq:app_erfc_inverse_laplace}
\end{equation}
therefore gives
\begin{align}
    &\mathcal L^{-1}_{z,z'\to \tau,\tau'}
    \left[
        \frac1{zz'(\sqrt z+\sqrt{z'})}
    \right]
    \nonumber\\
    &\qquad=
    \int_0^\infty
    \operatorname{erfc}
    \left(
        \frac{u}{2\sqrt \tau}
    \right)
    \operatorname{erfc}
    \left(
        \frac{u}{2\sqrt{\tau'}}
    \right)
    du.
    \label{eq:app_double_inverse_erfc}
\end{align}
At equal times \(\tau'=\tau\), the change of variables
\(x=u/(2\sqrt \tau)\) yields
\begin{align}
    \mathcal L^{-1}_{z,z'\to \tau,\tau}
    \left[
        \frac1{zz'(\sqrt z+\sqrt{z'})}
    \right]
    &=
    2\sqrt \tau
    \int_0^\infty\operatorname{erfc}(x)^2\,dx
    \nonumber\\
    &=
    \frac{2(2-\sqrt2)}{\sqrt\pi}\sqrt \tau.
    \label{eq:app_double_inverse_equal_time}
\end{align}
Applying this identity to the leading term in
\eqref{eq:app_joint_C0_singularity} gives the first term in
\eqref{eq:app_late_time_critical_theorem}. The uniform remainder and
complementary-frequency bounds in
Lemma~\ref{lem:app_joint_critical_expansion} allow dominated contour inversion
and give the stated \(o(K_P\tau^{1/2})\) remainder.
\end{proof}

\paragraph{Ensemble consequence.}
For \(E\) ensemble members,
\[
    \bar C_0
    =
    C_0^{\mathrm{cross}}
    +
    \frac1E
    \left(
        C_0^{\mathrm{self}}
        -
        C_0^{\mathrm{cross}}
    \right).
\]
Suppose the singular contribution in
\eqref{eq:app_joint_C0_singularity} belongs to
\(C_0^{\mathrm{self}}-C_0^{\mathrm{cross}}\), while the cross-model correlation
has no growing equal-time contribution. Under the uniform remainder and
complementary-frequency bounds for contour inversion in
Lemma~\ref{lem:app_joint_critical_expansion},
Theorem~\ref{thm:app_late_time_critical_growth} implies
\begin{equation}
    \mathcal L_{\mathrm{grow}}^{(E)}(\tau)
    \asymp
    \frac1E
    P^{-(a-1+b/2)}\tau^{1/2}.
    \label{eq:app_late_time_ensemble_corollary}
\end{equation}
The ensemble identity preserves the \(1/E\) factor at a fixed time.
Optimal stopping additionally depends on the decreasing cross contribution.
Appendices~\ref{app:finite_constant_target} and~\ref{app:continuum_cross_stopping}
establish its boundedness and approach to the limit for the noiseless
Gaussian specializations studied there and use the resulting derivatives
to derive the stopping laws.

\subsection{Effect of Label Noise}
\label{app:label_noise_growth}

We now allow \(\sigma>0\) and average over the fixed training-label noise,
which is shared by all ensemble members. The response equations are
unchanged because the model is linear and the noise is independent of the
features and target. The correlation equations acquire an additional
source through \eqref{eq:app_gf_C1C3_time}. All statements below use the
same gradient-flow limit as the preceding analysis.

\paragraph{Additional source in the single-model correlation.}
At \(N=P\), let \(\Delta\widehat C_i\) denote the difference between the
correlations with noise variance \(\sigma^2\) and those with \(\sigma=0\),
with all other parameters fixed. The double Laplace transform of the
constant noise covariance is \(\sigma^2/(zz')\). Since
\(G_{11}=G_{22}\), the two correlation equations give
\begin{equation}
    \bm G_P(z,z')
    \begin{pmatrix}\Delta\widehat C_0\\\Delta\widehat C_2\end{pmatrix}
    =\frac{\sigma^2}{zz'}
    \begin{pmatrix}1-G_{22,P}\\-G_{21,P}\end{pmatrix}.
    \label{eq:app_label_noise_source}
\end{equation}
The source vector is the first column of \(\bm I-\bm G_P\), so inversion
gives the exact identity within this closure
\begin{equation}
    \Delta\widehat C_0(z,z')
    =\frac{\sigma^2}{zz'}
    \left[\frac{G_{22,P}(z,z')}{D_P(z,z')}-1\right].
    \label{eq:app_label_noise_residual_exact}
\end{equation}
Thus label noise changes the fitted residual as well as adding
\(\sigma^2\) to the test loss.

\paragraph{Conditional late-time growth.}
Assume the joint response and determinant expansions in
Lemma~\ref{lem:app_joint_critical_expansion}. Assume also that they imply
\(G_{22,P}(z,z')=q_P(1+o(1))\) uniformly in the joint inversion region,
and that the remainder and complementary-frequency bounds extend to the
noise source, with error
\(o(\sigma^2q_Pd_P^{-1}\tau^{1/2})\) after inversion. Equation
\eqref{eq:app_label_noise_residual_exact} then has the leading growing term
\[
    \Delta\widehat C_{0,\mathrm{grow}}(z,z')
    =\frac{\sigma^2q_P}{d_P}
    \frac{1+o(1)}{zz'(\sqrt z+\sqrt{z'})}.
\]
The inverse transform in \eqref{eq:app_double_inverse_equal_time} gives
\begin{equation}
    \Delta C_0(\tau,\tau)
    =\frac{2(2-\sqrt2)}{\sqrt\pi}
    \frac{\sigma^2q_P}{d_P}\sqrt\tau
    +o\!\left(\frac{\sigma^2q_P}{d_P}\sqrt\tau\right),
    \qquad \tau/P^b\longrightarrow\infty.
    \label{eq:app_label_noise_growth}
\end{equation}
The explicit \(-\sigma^2/(zz')\) in
\eqref{eq:app_label_noise_residual_exact} and any other terms bounded in time
are lower order in this limit. Since \(q_P\asymp1\) and
\(d_P\asymp P^{b/2}\), the noise-induced growth has scale
\(\sigma^2P^{-b/2}\sqrt\tau\). This part of the calculation depends on
\(b>1\), but not on the target exponent \(a\). Comparing it with the signal
term in the range \(1<a<b+1\) gives the ratio
\[
    \frac{\sigma^2}{\mathcal B(P)}\asymp\sigma^2P^{a-1}.
\]
A fixed nonzero noise level therefore need not remain a small correction
as the dataset size increases.

\paragraph{Label noise and ensembling.}
The cross-model closure retains the shared labels. In the gradient-flow
limit of \eqref{eq:app_C0_cross}, its scalar feedback coefficient is
\(1-G_{22,P}\); the additional \(1/N\) feedback from independent random
features is absent. Consequently,
\begin{equation}
    \Delta\widehat C_0^{\mathrm{cross}}(z,z')
    =\frac{\sigma^2}{zz'}
    \left[\frac{1}{G_{22,P}(z,z')}-1\right].
    \label{eq:app_label_noise_cross_exact}
\end{equation}
Under \(G_{22,P}\to q_P>0\), the leading low-frequency term is
\(\sigma^2(q_P^{-1}-1)/(zz')\). If the inverse transforms of the cross-model
remainders and complementary frequencies stay bounded in time, the contribution
of label noise to the cross-model correlation is also bounded in time.
The transform itself is not regular at zero: the factor
\(1/(zz')\) represents a constant equal-time contribution.

Under these conditions and those of \eqref{eq:app_label_noise_growth},
the ensemble identity in Appendix~\ref{app:self_minus_cross} gives
\begin{equation}
    \Delta\mathcal L_{\mathrm{grow}}^{(E)}(\tau)
    \sim\frac1E\frac{2(2-\sqrt2)}{\sqrt\pi}
    \frac{\sigma^2q_P}{d_P}\sqrt\tau.
    \label{eq:app_label_noise_ensemble_growth}
\end{equation}
This leading growth is reduced by \(1/E\). The contribution of label noise
to the cross-model correlation and the irreducible test-label variance
\(\sigma^2\) remain.
For ensemble sizes that also grow in the limit, uniform control of the
remainders is required before using this leading term.

\paragraph{Comparison of signal and noise sources.}
For \(1<a<b+1\), adding the signal and noise sources replaces
\(\mathcal B(P)\) by \(\mathcal B(P)+\sigma^2\) in the leading growing
coefficient, under the assumptions above. Thus
\(\sigma^2\ll\mathcal B(P)\asymp P^{-(a-1)}\) makes the noise correction
to this coefficient small. Optimal stopping also depends on the noise
contribution to cross and its time derivative.
Appendix~\ref{app:weak_label_noise_stopping} describes the corresponding
conditions for stability of the stopping time and compute allocation.

\section{Constant Squared Teacher Coefficients in the Finite Gaussian Model}
\label{app:finite_constant_target}

We first establish the late-time result directly for a finite Gaussian
model with an arbitrary positive spectrum, $N=P$, $\sigma=0$, zero
initialization, and gradient flow. Let
$X\in\mathbb R^{P\times M}$ have independent $\mathcal N(0,\Lambda)$ rows, with
$\Lambda=\operatorname{diag}(\lambda_1,\ldots,\lambda_M)>0$, and let
$Q\in\mathbb R^{P\times M}$ have independent $\mathcal N(0,1/P)$ entries.
The predictor has coefficient vector $Q^\top\theta$, with gradient flow
on $\|XQ^\top\theta-Xw^\star\|^2/(2P)$. This is the normalization of
\eqref{eq:main_model}: $Q=A/\sqrt N$ and $\theta=w/\sqrt N$, with time
$\tau$ defined using the $N$-rescaled learning rate.
For $v_Q(\tau)=w^\star-Q^\top\theta(\tau)$, write
\begin{align}
 C_{\mathrm{self}}(\tau)
 &=\mathbb E_{X,Q}[v_Q(\tau)^\top\Lambda v_Q(\tau)],\nonumber\\
 C_{\mathrm{cross}}(\tau)
 &=\mathbb E_X[m_\tau(X)^\top\Lambda m_\tau(X)],
 \qquad m_\tau(X)=\mathbb E_Q[v_Q(\tau)\mid X].
 \label{eq:app_new_finite_correlations}
\end{align}
Independent ensemble members share the same dataset. Consequently,
\begin{equation}
 \mathcal L_E(\tau)=C_{\mathrm{cross}}(\tau)
 +\frac{C_{\mathrm{self}}(\tau)-C_{\mathrm{cross}}(\tau)}E,
 \qquad C_{\mathrm{self}}(\tau)-C_{\mathrm{cross}}(\tau)\ge0.
 \label{eq:app_new_ensemble_identity}
\end{equation}

\subsection{Conditional mean and the cross correlation}
Fix a full-row-rank dataset and define
$G=XX^\top$, $U=X^\top G^{-1/2}$, and $\Pi=UU^\top$.
Complete $U$ to an orthogonal basis $(U,V)$ of $\mathbb R^M$.
The feature blocks $QU$ and $QV$ are independent. With
$Z=XQ^\top$ and $K_X=ZZ^\top/P$, the fitted training predictions are
$(I-e^{-\tau K_X})Xw^\star$. At every finite $\tau$, the singular-value
gain $(1-e^{-\tau s^2/P})/s$ is bounded by $\sqrt{\tau/P}$, so the
conditional feature averages exist. In particular,
$\mathbb E[(QV)^\top\theta(\tau)\mid QU,X]=0$.
Set
\begin{equation}
 F_\tau(G)=\mathbb E_Q[e^{-\tau K_X}\mid X],\qquad
 M_\tau=I-\Pi+UF_\tau(G)U^\top.
 \label{eq:app_new_mean_response}
\end{equation}
Gaussian sign symmetry in an eigenbasis of $G$ implies that $F_\tau$
commutes with $G$. Hence $m_\tau=M_\tau w^\star$ and
$0\preceq M_\tau\preceq I$. This proves boundedness of the cross
correlation for any fixed finite target. Moreover,
$F_\tau\to0$ by bounded convergence, so
$m_\tau\to(I-\Pi)w^\star$.

Suppose now that $(w_k^\star)^2=q>0$ for every coordinate.
The dataset distribution is invariant under coordinate sign changes,
which make $\mathbb E_X[M_\tau\Lambda M_\tau]$ diagonal. Its quadratic
form at $w^\star$ therefore equals $q$ times its trace. Since
$M_\tau^2=I-\Pi+UF_\tau^2U^\top$,
\begin{equation}
 C_{\mathrm{cross}}(\tau)
 =C_\infty+q\mathbb E_X\operatorname{tr}
       [\Lambda U F_\tau(G)^2U^\top],\qquad
 C_\infty=q\mathbb E_X\operatorname{tr}[\Lambda(I-\Pi)].
 \label{eq:app_new_cross_trace}
\end{equation}
All $F_\tau(G)$ are simultaneously diagonal in an eigenbasis of $G$,
and decrease in positive-semidefinite order with $\tau$. Thus
$C_{\mathrm{cross}}$ is strictly above $C_\infty$ at every finite time,
is nonincreasing, and is at most $q\operatorname{tr}\Lambda$.
For a general target, the mixed term between $(I-\Pi)w^\star$ and
$UF_\tau U^\top w^\star$ need not vanish. The cancellation in
\eqref{eq:app_new_cross_trace} uses constant squared target coefficients,
not just equality of two asymptotic power-law exponents.

\subsection{The spectral edge and integrable bounds}
Write $K_X=G^{1/2}H^\top H G^{1/2}/P^2$ for a standard square Gaussian
matrix $H$. At fixed $G>0$, the density of $K_X$ is proportional to
\[
 (\det G)^{-P/2}(\det K_X)^{-1/2}
 \exp[-(P^2/2)\operatorname{tr}(G^{-1}K_X)].
\]
Let one eigenvalue $s$ approach zero with eigenvector $u$. The
Vandermonde factors involving $s$ supply one determinant of the positive
matrix on $u^\perp$. Integrating that matrix gives the factor
\[
 \det(U_\perp^\top G^{-1}U_\perp)^{-(P+1)/2}
 =[(\det G)^{-1}u^\top Gu]^{-(P+1)/2}.
\]
Thus the leading matrix spectral density is proportional to
$s^{-1/2}\sqrt{\det G}\,uu^\top/(u^\top Gu)^{(P+1)/2}$.
The remaining eigenvalue integrals are dominated at zero by integrable
negative half-powers times polynomials and at infinity by an exponential.
The scalar coefficient for $H^\top H$ is
$c_P=\Gamma((P+1)/2)/(\sqrt2\Gamma(P/2))$. It follows, for example,
from the ratio $PZ_{P-1,1/2}/Z_{P,-1/2}$, where
\[
 Z_{d,\alpha}=2^{d(\alpha+1)+d(d-1)/2}
 \prod_{j=1}^d
 \frac{\Gamma(1+j/2)\Gamma(\alpha+1+(j-1)/2)}{\Gamma(3/2)}.
\]
Rescaling by $P^2$ and integrating $s^{-1/2}e^{-\tau s}$ gives
\begin{align}
 \sqrt\tau F_\tau(G)&\longrightarrow W(G),&
 \tau^{3/2}\partial_\tau F_\tau(G)&\longrightarrow-\tfrac12W(G),
 \label{eq:app_new_edge_limits}\\
 W(G)&=P\beta_P\sqrt{\det G}
 \int_{S^{P-1}}
 \frac{uu^\top}{(u^\top Gu)^{(P+1)/2}}\,d\sigma(u),&
 \beta_P&=\frac{\sqrt\pi\Gamma((P+1)/2)}{\sqrt2\Gamma(P/2)}.
 \label{eq:app_new_edge_coefficient}
\end{align}
Here $d\sigma$ is normalized surface measure; $W(G)\succ0$.
The differentiated limit follows by including the factor $-s$ in the
same integral. To control the subsequent data average, the density of
the smallest eigenvalue of $H^\top H$ satisfies
$f_{\min}(s)\le c_Ps^{-1/2}$: shift the other eigenvalues by $s$ in
the joint density and bound their resulting negative half-powers.
Eigenvalue comparison and
$s e^{-\tau s}\le2e^{-\tau s/2}/(e\tau)$ then give
\begin{equation}
 \|F_\tau(G)\|\le\frac{P\beta_P}{\sqrt{\tau\lambda_{\min}(G)}},
 \qquad
 \|\partial_\tau F_\tau(G)\|
 \le\frac{2\sqrt2P\beta_P}
 {e\tau^{3/2}\sqrt{\lambda_{\min}(G)}}.
 \label{eq:app_new_edge_domination}
\end{equation}
For $M>P+1$, the inverse-Wishart moment bound is
\[
 \mathbb E\lambda_{\min}(G)^{-1}
 \le\frac{P}{\lambda_{\min}(\Lambda)(M-P-1)}<\infty.
\]
Dominated convergence in \eqref{eq:app_new_cross_trace} now proves
\begin{align}
 C_{\mathrm{cross}}(\tau)&=C_\infty+D\tau^{-1}+o(\tau^{-1}),
 \qquad C_{\mathrm{cross}}'(\tau)=-D\tau^{-2}+o(\tau^{-2}),\nonumber\\
 D&=q\mathbb E_X\operatorname{tr}[\Lambda U W(G)^2U^\top]>0.
 \label{eq:app_new_finite_cross_tail}
\end{align}

\subsection{Self correlation and optimal stopping}
Put $T_X=\operatorname{tr}[\Lambda(I-\Pi)]$,
$\mathsf E_\tau=e^{-\tau K_X}$, and
$\Psi_\tau(K_X)=(I-e^{-\tau K_X})^2K_X^{-1}$.
Conditional on $QU$, the independent block $QV$ contributes
$T_X\|\theta(\tau)\|^2/P$ to the risk. Applying the same data-coordinate
sign average as above gives
\begin{align}
 C_{\mathrm{self}}(\tau)&=C_\infty+R_0(\tau)+V_0(\tau),\nonumber\\
 R_0(\tau)&=q\mathbb E_{X,Q}\operatorname{tr}
 [\Lambda UG^{-1/2}\mathsf E_\tau G\mathsf E_\tau G^{-1/2}U^\top],
 \nonumber\\
 V_0(\tau)&=\frac q{P^2}\mathbb E_X
 [T_X\operatorname{tr}(G\mathbb E_Q[\Psi_\tau(K_X)\mid X])].
 \label{eq:app_new_finite_self_decomposition}
\end{align}
The row-space term tends to zero and obeys
\[
 R_0(\tau)\le q\lambda_{\max}(\Lambda)
 \mathbb E[\operatorname{tr}G\operatorname{tr}G^{-1}],\qquad
 |R_0'(\tau)|\le\frac{2q\lambda_{\max}(\Lambda)}{e\tau}
 \mathbb E[\operatorname{tr}G\operatorname{tr}G^{-1}].
\]
This moment is finite: for $W_0=ZZ^\top$ with standard Gaussian
$Z\in\mathbb R^{P\times M}$, radial independence gives
$\mathbb E[\operatorname{tr}W_0\operatorname{tr}W_0^{-1}]
=(PM-2)P/(M-P-1)$, and eigenvalue comparison transfers the bound to $G$.
For the other term, the identity
\[
 \partial_\tau\mathbb E_Q[\Psi_\tau(K_X)\mid X]
 =2(F_\tau(G)-F_{2\tau}(G))
\]
and \eqref{eq:app_new_edge_limits} yield limits
$2(2-\sqrt2)W(G)$ and $(2-\sqrt2)W(G)$ after multiplying
$\mathbb E_Q\Psi_\tau$ by $\tau^{-1/2}$ and its derivative by
$\sqrt\tau$, respectively. The bounds in
\eqref{eq:app_new_edge_domination} dominate these quantities by
$4P\beta_P/\sqrt{\lambda_{\min}(G)}$ and
$2P\beta_P/\sqrt{\lambda_{\min}(G)}$.
Since $T_X\le\operatorname{tr}\Lambda$,
$\mathbb E(\operatorname{tr}G)^2<\infty$, and the first inverse moment
is finite, Cauchy--Schwarz permits the data averages. Therefore
\begin{align}
 C_{\mathrm{self}}(\tau)&=H\sqrt\tau+o(\sqrt\tau),\qquad
 C_{\mathrm{self}}'(\tau)=\frac H{2\sqrt\tau}+o(\tau^{-1/2}),\nonumber\\
 H&=\frac{2q(2-\sqrt2)}{P^2}
 \mathbb E_X[T_X\operatorname{tr}(GW(G))]>0.
 \label{eq:app_new_finite_self_tail}
\end{align}
These statements concern the expectation over data and random features.
Every fixed full-rank realization has a finite limit. Nearly singular
feature matrices prevent uniform integrability of the squared error.

\begin{theorem}[Stopping with constant squared target coefficients]
\label{thm:app_new_finite_stopping}
Fix $N=P$, $M>P+1$, $\Lambda>0$, and $(w_k^\star)^2=q>0$, with the
noiseless Gaussian model and initialization specified above. Every
sequence of global minimizers $\tau_\star(P,E)$ of the expected loss
satisfies, as $E\to\infty$,
\begin{equation}
 \tau_\star(P,E)\sim\left(\frac{2D}{H}E\right)^{2/3},\qquad
 \min_\tau\mathcal L_E(\tau)-C_\infty
 \sim3D^{1/3}(H/2)^{2/3}E^{-2/3}.
 \label{eq:app_new_finite_stopping}
\end{equation}
The positive coefficients $D,H$ are given in
\eqref{eq:app_new_finite_cross_tail} and
\eqref{eq:app_new_finite_self_tail}.
\end{theorem}
\begin{proof}
For finite $E$, continuity and self growth guarantee a minimizer.
For every fixed time, the ensemble loss tends to the cross correlation
as $E\to\infty$; the infimum therefore tends to $C_\infty$.
On each bounded time interval the cross correlation is strictly above
$C_\infty$, and \eqref{eq:app_new_ensemble_identity} bounds the ensemble
loss from below by cross. All minimizers consequently diverge and are
eventually interior. At a minimizer, the differentiated tail estimates
give $D(1-E^{-1})\tau_\star^{-2}(1+o(1))
=H(2E)^{-1}\tau_\star^{-1/2}(1+o(1))$.
This proves the time formula; substitution proves the loss formula.
\end{proof}
The theorem gives the ensemble exponent at fixed $P,M$, with coefficient
determined by $D/H$.
Trace-class infinite spectra can also be treated at fixed $P$ by using
a fixed positive block with more than $P+3$ modes to dominate inverse
moments before taking the feature cutoff to infinity. Uniform estimates
for growing $P$ or stopping times require control of these bounds in
the joint limit.

\section{Continuum Cross Correlations and General Stopping Laws}
\label{app:continuum_cross_stopping}

We now evaluate the GF closure for the pure-power continuum spectrum
$\lambda_k=k^{-b}$ and $(w_k^\star)^2=k^{b-a}$, with
$b>1$ and $1<a<b+1$. Mode sums are replaced by integrals over
$k\in(0,\infty)$. This describes the scaling limit with a large feature
cutoff relative to $P$; it differs from the ultimate finite-system
spectral edge in Appendix~\ref{app:finite_constant_target}.
The assumptions $N=P$, $\sigma=0$, and zero initialization remain in force.
Define, locally in this section,
\begin{equation}
 r=\frac1b,\quad p=\frac{a-1}{b},\quad
 c_b=\frac{\pi}{b\sin(\pi/b)},\quad
 K=\frac{\pi}{b\sin(\pi p)},\quad
 \kappa=(c_b/P)^b,\quad g=\frac{2(2-\sqrt2)}{\sqrt\pi}.
 \label{eq:app_new_continuum_constants}
\end{equation}
Both $p$ and $r$ lie in $(0,1)$.

\subsection{Cross transform, expansion, and inversion}
Set $h(z)=z/R(z)^2$, on the physical branch $h(z)>\kappa$ for $z>0$.
The response equation becomes
\begin{equation}
 R(z)=1-(\kappa/h(z))^r,\qquad
 z=h(z)[1-(\kappa/h(z))^r]^2.
 \label{eq:app_new_continuum_response}
\end{equation}
Writing $h=h(z)$ and $h'=h(z')$, integration of the source and feedback
spectra gives
\begin{align}
 U_p(h,h')&=\int_0^\infty
 \frac{k^{-a}\,dk}{(k^{-b}+h)(k^{-b}+h')}
 =K\frac{h^{p-1}-(h')^{p-1}}{h'-h},\nonumber\\
 V(h,h')&=\int_0^\infty
 \frac{k^{-2b}\,dk}{(k^{-b}+h)(k^{-b}+h')}
 =c_b\frac{(h')^{1-r}-h^{1-r}}{h'-h}.
 \label{eq:app_new_continuum_integrals}
\end{align}
Divided differences use their continuous value at coincident arguments.
The cross equation is consequently
\begin{equation}
 \widehat C_{\mathrm{cross}}(z,z')=\frac{F(h,h')}{zz'},\qquad
 F(h,h')=\frac{hh'U_p(h,h')}{1-V(h,h')/P}
 =Khh'\frac{h^{p-1}-(h')^{p-1}}{h'R(z')-hR(z)}.
 \label{eq:app_new_cross_transform}
\end{equation}
At zero frequency the denominator $1-V/P$ equals $r>0$, and
\begin{equation}
 C_\infty=F(\kappa,\kappa)
 =K(b-a+1)\kappa^p\asymp P^{-(a-1)}.
 \label{eq:app_new_continuum_plateau}
\end{equation}
The response has expansion
$h(z)=\kappa+(\sqrt\kappa/r)\sqrt z+O(z)$. Direct differentiation gives
\begin{equation}
 \frac{F_h(\kappa,\kappa)}{C_\infty}
 =\frac{p+r-1}{2\kappa},\qquad
 \frac{F_{hh'}(\kappa,\kappa)}{C_\infty}
 =\frac{p^2-2p+3pr+4-6r+2r^2}{6\kappa^2}.
 \label{eq:app_new_cross_partials}
\end{equation}
Inverting these terms, with the justification below, yields
\begin{align}
 C_{\mathrm{cross}}(\tau)
 &=C_\infty+J\tau^{-1/2}+D\tau^{-1}+O(\tau^{-3/2}),\nonumber\\
 C_{\mathrm{cross}}'(\tau)
 &=-\tfrac12J\tau^{-3/2}-D\tau^{-2}+O(\tau^{-5/2}),\nonumber\\
 J&=\frac{C_\infty(a-b)}{\sqrt{\pi\kappa}},\qquad
 D=\frac{C_\infty[(a-b)^2+(a-b)+3b(b-1)]}{6\pi\kappa}>0.
 \label{eq:app_new_continuum_cross_tail}
\end{align}
When $a=b$, $F(h,\kappa)=P\kappa$ for all $h$, so every pure-variable
correction vanishes. The cross remainder then improves to $O(\tau^{-2})$,
and $D=Pb(b-1)/(2\pi)$.

To justify the diagonal inversion, rescale $h=\kappa u$, $z=\kappa Z$.
The upper boundary of the physical inverse of $Z=u(1-u^{-r})^2$ is
\[
 u=\rho(\vartheta)e^{i\vartheta},\qquad
 \rho(\vartheta)=
 \left[\frac{\cos((r-1/2)\vartheta)}{\cos(\vartheta/2)}\right]^{1/r},
 \qquad0<\vartheta<\pi.
\]
It maps to the negative real $Z$ axis; the other boundary $u\ge1$ maps
to the positive axis. The derivative
$(1-u^{-r})[1+(2r-1)u^{-r}]$ has no zero in the intervening upper
domain, and $Z\sim u$ at infinity. The argument principle therefore
gives a conformal inverse mapping the upper half-plane to itself.
Thus $h$ is a complete Bernstein function: it is positive on the positive
axis, analytic off the negative axis, and maps the upper half-plane into itself.
On the cut boundary, direct integration gives
\[
 \frac{V(h,\overline h)}P
 =\frac{\sin((1-r)\vartheta)}
 {\sin(r\vartheta)+\sin((1-r)\vartheta)}
 \le\max(1-r,1/2)<1.
\]
The inequality follows from monotonicity of $x\cot x$ on $(0,\pi)$;
the radial factor decreases this bound inside the domain.
Cauchy--Schwarz in the spectral integral therefore proves the uniform gap
\begin{equation}
 |1-V(h(z),h(z'))/P|\ge\min(r,1/2)>0
 \label{eq:app_new_cross_gap}
\end{equation}
on all physical branches. There are no additional cross-denominator poles.

Subtract $C_\infty$, the boundary term $F(h,\kappa)-C_\infty$, and its
primed counterpart from $F$. The remainder vanishes if either argument is
$\kappa$, and its transform divided by $zz'$ has a double-cut inverse.
The boundary terms have ordinary single-cut inverses. Their endpoint
densities are respectively
\[
 \frac{F_h\sqrt\kappa}{\pi r}s^{-1/2}+O(s^{1/2}),\qquad
 \frac{\kappa F_{hh'}}{\pi^2r^2}(ss')^{-1/2}
 +\text{higher endpoint powers}.
\]
These follow from $h(-s\pm i0)-\kappa
=\pm i(\sqrt\kappa/r)\sqrt s+O(s)$.
Analytic integer powers have no cut discontinuity in that variable.
The exact integrals and \eqref{eq:app_new_cross_gap} bound the densities
polynomially away from the endpoints, including regions where only one
frequency is small. Exponential factors in the inverse transforms then
justify termwise endpoint integration and time differentiation. Integrating
$s^{-1/2}$ against $e^{-\tau s}$ proves
\eqref{eq:app_new_continuum_cross_tail}, including its derivative.

\subsection{A scalar inverse for self growth}
Solving the coupled self equations gives
\begin{equation}
 \widehat C_{\mathrm{self}}(z,z')
 =Khh'\frac{R(z')h^{p-1}-R(z)(h')^{p-1}}{zz'(z'-z)}.
 \label{eq:app_new_self_transform}
\end{equation}
The determinant of the coupled system is $(z'-z)/(h'-h)$; the apparent
singularity at $z=z'$ is removable. Define
$A(z)=h(z)^p/z$ and $B(z)=\sqrt{h(z)/z}=1/R(z)$.
Both are Stieltjes functions (positive on the positive axis and mapping
the upper half-plane into the lower half-plane). Write
$a_0(\tau)=\mathcal L^{-1}[A](\tau)$ and
$\beta(\tau)=\mathcal L^{-1}[B-1](\tau)$, separating the constant
part of $B$ from its ordinary inverse. Then
\[
 \widehat C_{\mathrm{self}}(z,z')
 =K\frac{A(z)B(z')-B(z)A(z')}{z'-z}.
\]
Its initial slice is $C_{\mathrm{self}}(\tau,0)=Ka_0(\tau)$.
Inverting $(z'-z)\widehat C_{\mathrm{self}}$ and including the initial
slice yields
$(\partial_{\tau'}-\partial_\tau)C_{\mathrm{self}}
=K[a_0(\tau)\beta(\tau')-\beta(\tau)a_0(\tau')]$.
Integration along $(2\tau-v,v)$ therefore gives the exact identity
\begin{equation}
 C_{\mathrm{self}}(\tau,\tau)=Ka_0(2\tau)
 +K\int_0^\tau
 [a_0(2\tau-v)\beta(v)-\beta(2\tau-v)a_0(v)]\,dv.
 \label{eq:app_new_self_scalar_inverse}
\end{equation}
Single-cut inversion gives
\begin{align*}
 a_0(\tau)&=\kappa^p+O(\tau^{-1/2}),&
 a_0'(\tau)&=O(\tau^{-3/2}),\\
 \beta(\tau)&=\frac{\sqrt\kappa}{\sqrt{\pi\tau}}+O(\tau^{-3/2}),&
 \beta'(\tau)&=-\frac{\sqrt\kappa}{2\sqrt\pi\tau^{3/2}}
 +O(\tau^{-5/2}).
\end{align*}
At zero, $a_0(\tau)\sim\tau^{-p}/\Gamma(1-p)$ and
$\beta(\tau)\sim\kappa^r\tau^{r-1}/\Gamma(r)$, so both are locally
integrable. Replacing $a_0$ by $\kappa^p$ in the integral gives
$gK\kappa^{p+1/2}\sqrt\tau+O(1)$.
For each remaining term, the late argument lies in $[\tau,2\tau]$,
and its $O(\tau^{-1/2})$ bound multiplies an $O(\sqrt\tau)$ integral.
The remainder is therefore $O(1)$. Differentiation has no moving-endpoint
term, since the integrand vanishes at $v=\tau$; each differentiated
remainder is $O(\tau^{-3/2})O(\sqrt\tau)=O(\tau^{-1})$. Hence
\begin{equation}
 C_{\mathrm{self}}(\tau)=H\sqrt\tau+O(1),\qquad
 C_{\mathrm{self}}'(\tau)=\frac H{2\sqrt\tau}+O(\tau^{-1}),\qquad
 H=gK\kappa^{p+1/2}\asymp P^{-(a-1+b/2)}.
 \label{eq:app_new_continuum_self_tail}
\end{equation}

\subsection{Global minima and the three stopping laws}
A positive late cross correction alone would not exclude an earlier
minimum below $C_\infty$. Here this can be ruled out for $a\ge b$.
Let $\alpha=b-a+1$ and $v(z)=(\kappa/h(z))^r$. The boundary transform is
\[
 F(h(z),\kappa)=K\kappa^p\frac{1-v(z)^\alpha}{1-v(z)},\qquad
 \frac{1-v^\alpha}{1-v}
 =\alpha\int_0^1(1-s+sv)^{\alpha-1}\,ds.
\]
For $a>b$, $0<\alpha<1$, and $v$ is Stieltjes. The expression in
brackets after subtraction of $\alpha$ is a bounded complete Bernstein
function vanishing at zero. Its quotient by $z$ is Stieltjes, with a
strictly positive inverse. Thus
$f(\tau)=C_{\mathrm{cross}}(\tau,\infty)-C_\infty>0$ at every finite
positive time. For $a=b$, $f$ is identically zero.

To control the remaining term, expand $1/(1-V/P)$ in
\eqref{eq:app_new_cross_transform}. The scalar factors
\[
 a_\lambda(z)=\frac{h(z)}{z(\lambda+h(z))},\qquad
 b_\lambda(z)=\frac{\lambda}{\lambda+h(z)}
\]
have nonnegative time inverses. Each term in the expansion is an integral
of outer products of $a_\lambda$ convolved with the $b_\lambda$ kernels.
The mass per feedback step is $1-r<1$, so the expansion converges;
one may first truncate the source and then remove the truncation by
positivity and the positive-time inverse above. This Gram representation
implies
\[
 C_{\mathrm{cross}}(\tau,\tau)
 -2C_{\mathrm{cross}}(\tau,\infty)+C_\infty\ge0.
\]
It is strictly positive at finite positive time, since already its
zeroth term contains
$\int\lambda(w_\lambda^\star)^2
[a_\lambda(\tau)-a_\lambda(\infty)]^2\,d\rho(\lambda)>0$.
Consequently $C_{\mathrm{cross}}(\tau)>C_\infty$ for $a\ge b$.
The continuum curve diverges at zero, as follows from its positive source
term with short-time power $\tau^{-p}$, while self growth controls infinity.
Global ensemble minimizers exist and, for $a\ge b$, diverge as
$E\to\infty$. Evaluating the differentiated expansions there proves
\begin{equation}
 \begin{aligned}
 a>b:\quad &
 \tau_\star\sim\frac JH E
 =\frac{(b-a+1)(a-b)}{2(2-\sqrt2)}\kappa^{-1}E,
 \\[-2pt]
 &\mathcal L_E(\tau_\star)-C_\infty
 \sim2\sqrt{JH}\,E^{-1/2},\\[4pt]
 a=b:\quad &
 \tau_\star\sim\left(\frac{2D}{H}E\right)^{2/3}
 =\kappa^{-1}\left[\frac{b(b-1)}{\pi g}E\right]^{2/3},
 \\[-2pt]
 &\mathcal L_E(\tau_\star)-C_\infty
 \sim3D^{1/3}(H/2)^{2/3}E^{-2/3}.
 \end{aligned}
 \label{eq:app_new_continuum_stopping}
\end{equation}

For $a<b$, $J<0$, so the cross curve has a value strictly below its
long-time limit. Its minimizing set
$\mathcal S(P)=\operatorname*{arg\,min}_{\tau>0}C_{\mathrm{cross}}(\tau)$
is a compact subset of $(0,\infty)$.
Since $\mathcal L_E\ge C_{\mathrm{cross}}$ and converges uniformly to
cross on compact intervals, all large-$E$ minimizers lie in one such
interval, and every accumulation point belongs to $\mathcal S(P)$.
With $m=\min_\tau C_{\mathrm{cross}}(\tau)<C_\infty$ and
$V(\tau)=C_{\mathrm{self}}(\tau)-C_{\mathrm{cross}}(\tau)$, compactness
and continuity also give
\begin{equation}
 \min_\tau\mathcal L_E(\tau)
 =m+\frac{\min_{\tau\in\mathcal S(P)}V(\tau)}E+o(E^{-1}).
 \label{eq:app_new_below_cancellation}
\end{equation}
Indeed, evaluating at a minimizer of $V$ on $\mathcal S(P)$ gives the upper
bound; the lower bound follows by compactness of the ensemble minimizers
and $C_{\mathrm{cross}}\ge m$. The $E^{-1}$ scaling has a nonzero
coefficient when $\min_{\mathcal S(P)}V>0$.
Thus the stopping times remain of order $P^b$, rather than growing as
a power of $E$; their limiting location requires the complete cross curve.

Keeping both cross corrections gives a useful late-time balance. With
$u_\star=\sqrt{\kappa\tau_\star}$,
$j=(a-b)/\sqrt\pi$, and
$d=[(a-b)^2+(a-b)+3b(b-1)]/(6\pi)$,
\begin{equation}
 \frac{g}{b-a+1}u_\star^3
 \simeq(E-1)(ju_\star+2d).
 \label{eq:app_new_common_stopping_balance}
\end{equation}
This requires a late stopping time and is not an exact finite-$E$
formula. Near $a=b$, the two derivative terms become comparable at
$E\asymp|a-b|^{-3}$. For $a<b$ close to this line it estimates
$u_\star\simeq2d/|j|$; far from the line this time need not be late.

\subsection{Explicit curves for \texorpdfstring{$a=b=2$}{a=b=2}}
Here $\kappa=(\pi/(2P))^2$ and
$h(z)=(\sqrt\kappa+\sqrt z)^2$.
Substitution in \eqref{eq:app_new_cross_transform} gives
\[
 \widehat C_{\mathrm{cross}}(z,z')
 =\frac{P\kappa}{zz'}+
 \frac{P\sqrt\kappa}
 {\sqrt z\sqrt{z'}(\sqrt\kappa+\sqrt z+\sqrt{z'})}.
\]
Using $(\sqrt\kappa+\sqrt z+\sqrt{z'})^{-1}
=\int_0^\infty e^{-s(\sqrt\kappa+\sqrt z+\sqrt{z'})}\,ds$ and
$\mathcal L^{-1}[e^{-s\sqrt z}/\sqrt z](\tau)
=e^{-s^2/(4\tau)}/\sqrt{\pi\tau}$ gives a positive integral inverse.
The corresponding self transform is
\[
 \widehat C_{\mathrm{self}}(z,z')
 =\frac{P\kappa^{3/2}}{zz'(\sqrt z+\sqrt{z'})}
 +\frac{P\kappa}{zz'}
 +\frac{P\sqrt\kappa}{\sqrt z\sqrt{z'}(\sqrt z+\sqrt{z'})}.
\]
Consequently, writing $x=\tau/P^2$,
$\operatorname{erfcx}(s)=e^{s^2}\operatorname{erfc}(s)$, and
$A=(\pi^3/8)g$,
\begin{align}
 PC_{\mathrm{cross}}(P^2x)
 &=\frac{\pi^2}{4}+\sqrt{\frac\pi8}x^{-1/2}
 \operatorname{erfcx}\!\left(\frac{\pi\sqrt x}{2\sqrt2}\right),
 \nonumber\\
 PC_{\mathrm{self}}(P^2x)
 &=\frac{\pi^2}{4}+\sqrt{\frac\pi8}x^{-1/2}+A\sqrt x,
 \label{eq:app_new_exact_flat_curves}\\
 P\mathcal L_E(P^2x)
 &=\frac{\pi^2}{4}+\sqrt{\frac\pi8}x^{-1/2}
 \left[\frac1E+\left(1-\frac1E\right)
 \operatorname{erfcx}\!\left(\frac{\pi\sqrt x}{2\sqrt2}\right)\right]
 +\frac{A\sqrt x}{E}.
 \label{eq:app_new_exact_flat_ensemble}
\end{align}
Expanding the cross transient gives
$PC_{\mathrm{cross}}-\pi^2/4
=1/(\pi x)-4/(\pi^3x^2)+O(x^{-3})$.
The remaining $x^{-1/2}$ term in the ensemble loss carries $1/E$.
Differentiating \eqref{eq:app_new_exact_flat_ensemble}, the stopping equation is
\[
 A x^{3/2}-\sqrt{\pi/8}\,x^{1/2}
 -\frac{2(E-1)}\pi+\frac{16(E-1)}{\pi^3x}
 +O(Ex^{-2})=0.
\]
Expanding around $x=[2E/(\pi A)]^{2/3}$ gives
\begin{equation}
 \frac{\tau_\star}{P^2}
 =\left(\frac{2E}{\pi A}\right)^{2/3}
 +\frac{2\sqrt{\pi/8}}{3A}-\frac{16}{3\pi^2}
 +O(E^{-1/3}).
 \label{eq:app_finite_E_stopping_correction}
\end{equation}
The constant correction is negative, so the large-$E$ log--log slope
approaches $2/3$ from above.
These are complete positive-time curves of the noiseless continuum
closure with zero initialization. The discrete model has a finite initial
loss; comparisons as $P$, the cutoff, and $E$ grow together require
uniform control of the continuum approximation over the stopping window.
Appendix~\ref{app:width_ensemble_allocation} develops the corresponding
width comparison below interpolation.

\section{Compute Allocation with Fixed Data}
\label{app:compute_optimal_scaling}

We use \(\mathcal L(\tau,N,P,E)\) for the ensemble test loss, with training
time first. The main allocation problem holds the dataset and member size
fixed at \(N=P\). Each ensemble is trained to its own optimal stopping
time, and only the number of members is chosen under the compute budget.
We first optimize width and training time for a single model with abundant
data, then turn to ensemble allocation at fixed \(N=P\).

The stopping results below concern noiseless GF with zero initialization.
The explicit dependence on \(a,b,P\) uses the pure-power continuum
spectrum of Appendix~\ref{app:continuum_cross_stopping}, with
\(b>1\) and \(1<a<b+1\). The cancellation at \(a=b\) corresponds to
constant squared teacher coefficients. Appendix~\ref{app:finite_constant_target}
gives the corresponding finite Gaussian-model argument for that target.

\paragraph{Time and compute conventions.}
With \(s\) updates and \(\mathcal E\) epochs, an ensemble of \(E\) models
uses a number of random-feature multiply-adds proportional to
\[
    ENBs=EN\frac{B}{\eta}\tau=ENP\mathcal E,
    \qquad \tau=\eta s,\quad Bs=P\mathcal E.
\]
We fix \(B\) and \(\eta\), absorb the common factor \(B/\eta\) into the
compute units, and write
\begin{equation}
    C=EN\tau.
    \label{eq:app_compute_budget}
\end{equation}
This normalization also permits comparisons with varying \(B\) and \(\eta\)
if \(B/\eta\) is held fixed. Otherwise the factor must be restored. At
fixed \(N,P,\tau\), the GF equations do not depend separately on batch size.
The finite-step comparison in Appendix~\ref{app:finite_step_batches}
shows how changing the batch partition and learning rate changes the
epoch update. Batch-size allocation uses these discrete dynamics and the
unnormalized cost \(ENBs\).

\paragraph{Single-model allocation with abundant data.}
For \(E=1\), when the leading loss is insensitive to data reuse, the time
and width terms give
\begin{equation}
    \mathcal L(\tau,N)
    \asymp
    A_\tau\tau^{-(a-1)/b}+A_NN^{-(a-1)}.
    \label{eq:app_compute_bound_loss}
\end{equation}
This approximation underlies the usual compute-constrained calculation
\citep{hoffmann2022training,bordelon2024dynamical}. Substituting
\(\tau=C/N\) and balancing the two terms gives
\begin{equation}
    N_\star(C)\asymp C^{1/(b+1)},
    \qquad
    \tau_\star(C)\asymp C^{b/(b+1)},
    \label{eq:app_compute_optimal_N_s}
\end{equation}
and hence
\begin{equation}
    \mathcal L^*(C)\asymp C^{-(a-1)/(b+1)}.
    \label{eq:app_compute_optimal_loss}
\end{equation}

\paragraph{Loss at optimal stopping with fixed data.}
Hold \(N=P\) fixed and abbreviate
\(\mathcal L(\tau,P,E)=\mathcal L(\tau,P,P,E)\). For independent members
sharing one dataset,
\begin{equation}
    \mathcal L(\tau,P,E)
    =\left(1-\frac1E\right)C_{\mathrm{cross}}(\tau)
      +\frac1E C_{\mathrm{self}}(\tau).
    \label{eq:app_local_resource_scaling}
\end{equation}
Appendix~\ref{app:continuum_cross_stopping} derives both the late-time
expansions and the derivative estimates needed to optimize this loss:
\begin{equation}
\begin{aligned}
    C_{\mathrm{cross}}(\tau)
      &=C_\infty+\frac{J}{\sqrt\tau}+\frac{D}{\tau}
          +O(\tau^{-3/2}),\\
    C_{\mathrm{self}}(\tau)&=H\sqrt\tau+O(1),
\end{aligned}
    \label{eq:app_stopping_surrogate}
\end{equation}
where \(D,H>0\). Put
\[
    c_b=\frac{\pi}{b\sin(\pi/b)},\qquad
    \kappa=(c_b/P)^b,\qquad
    g=\frac{2(2-\sqrt2)}{\sqrt\pi}.
\]
In the continuum model, \(C_\infty\asymp P^{-(a-1)}\) and
\begin{equation}
\begin{aligned}
    J=\frac{C_\infty j}{\sqrt\kappa},\qquad
    D=\frac{C_\infty d}{\kappa},\qquad
    H=\frac{gC_\infty\sqrt\kappa}{b-a+1},\\
    j=\frac{a-b}{\sqrt\pi},\quad
    d=\frac{(a-b)^2+(a-b)+3b(b-1)}{6\pi}>0.
\end{aligned}
    \label{eq:app_stopped_loss_prefactor}
\end{equation}
The time-dependent terms at an interior late minimum obey
\begin{equation}
    H\tau_\star^{3/2}
       \simeq(E-1)\bigl(J\sqrt{\tau_\star}+2D\bigr).
    \label{eq:app_stopping_prefactor}
\end{equation}
The bounded and decreasing terms from self carry a factor \(1/E\).
Their derivatives are lower order on the diverging stopping scales below.
Equation~\eqref{eq:app_stopping_prefactor} is an asymptotic balance;
small ensembles, including \(E=1\), require the full loss curve.

For \(a>b\), \(J>0\), and the cross curve lies above \(C_\infty\) at
every finite time. For \(a=b\), \(J=0\), and the positive \(D/\tau\)
term controls the approach to \(C_\infty\). The global-minimum argument
in Appendix~\ref{app:continuum_cross_stopping} then gives
\begin{equation}
\begin{aligned}
 a>b:\quad
 &\tau_\star(P,E)\sim\frac{J}{H}E\asymp P^bE,\\
 &\mathcal L^*(P,E)-C_\infty
       \sim2\sqrt{JH}\,E^{-1/2},\\
 a=b:\quad
 &\tau_\star(P,E)\sim\left(\frac{2D}{H}E\right)^{2/3}
           \asymp P^bE^{2/3},\\
 &\mathcal L^*(P,E)-C_\infty
       \sim3D^{1/3}(H/2)^{2/3}E^{-2/3},
\end{aligned}
    \label{eq:app_stopped_budget_loss}
\end{equation}
where \(\mathcal L^*(P,E)=\min_{\tau>0}\mathcal L(\tau,P,E)\).
At \(a=b=2\), the decreasing term has size \(P/\tau\) and the growing
term has size \(\sqrt\tau/(EP^2)\). Their balance gives
\(\tau_\star\asymp P^2E^{2/3}\).

For \(a<b\), \(J<0\), so cross is below its limiting value at some
finite times. Define
\[
    m(P)=\min_{\tau>0}C_{\mathrm{cross}}(\tau),\qquad
    \mathcal S(P)=\operatorname*{arg\,min}_{\tau>0}C_{\mathrm{cross}}(\tau),
    \qquad
    V(\tau)=C_{\mathrm{self}}(\tau)-C_{\mathrm{cross}}(\tau).
\]
Then \(m(P)<C_\infty\), and optimal times remain in a compact interval
of positive multiples of \(P^b\), approaching \(\mathcal S(P)\) as \(E\to\infty\).
The full cross curve is needed to locate this interval. Moreover,
\begin{equation}
    \mathcal L^*(P,E)
       =m(P)+\frac{\min_{\tau\in \mathcal S(P)}V(\tau)}{E}+o(E^{-1}),
    \qquad a<b.
    \label{eq:app_stopped_loss_below}
\end{equation}
For the \(E^{-1}\) excess-loss rate below, we additionally assume
\(\min_{\tau\in \mathcal S(P)}V(\tau)>0\); its coefficient then scales as
\(P^{-(a-1)}\) in the continuum description. If the cross minimum is
unique, \(\tau_\star(P,E)\) converges to its time. Thus the large-\(E\)
excess in this branch is measured above \(m(P)\), not \(C_\infty\).
If the minimum variance vanishes, \eqref{eq:app_stopped_loss_below}
only gives an \(o(E^{-1})\) excess, and its leading rate requires a
higher-order calculation.

\paragraph{Allocation over ensemble size.}
Let \(\tau_\star(P,E)\) be the earliest global minimizer when there is a
tie. The exact cost and feasible set for the allocation considered here are
\begin{equation}
    C_E=EP\tau_\star(P,E),\qquad
    \mathcal A(C;P)=\{E\in\mathbb N_{\ge1}:C_E\le C\}.
    \label{eq:app_stopped_cost}
\end{equation}
The ensemble identity gives \(V(\tau)\ge0\), so \(\mathcal L^*(P,E)\)
is nonincreasing in \(E\). When \(\mathcal A(C;P)\) is nonempty, a
largest affordable ensemble therefore attains the smallest loss in this
set. At small \(E\), feasibility is evaluated using stopping times obtained
from the full loss curve. If the set is empty,
no ensemble can reach its unconstrained minimum under this budget.

At large \(E\), the costs implied by
\eqref{eq:app_stopped_budget_loss}--\eqref{eq:app_stopped_loss_below} are
\[
 C_E\asymp
 \begin{cases}
 P^{b+1}E^2,&a>b,\\
 P^{b+1}E^{5/3},&a=b,\\
 P^{b+1}E,&a<b.
 \end{cases}
\]
Consequently, as \(C/P^{b+1}\to\infty\),
\begin{equation}
    E_\star(C;P)\asymp
    \begin{cases}
       (C/P^{b+1})^{1/2},&a>b,\\
       (C/P^{b+1})^{3/5},&a=b,\\
       C/P^{b+1},&a<b.
    \end{cases}
    \label{eq:app_stopped_optimal_ensemble}
\end{equation}
Integer rounding affects subleading terms in these large-budget laws.
For the loss over the feasible set, write
\[
    \mathcal L^*(C;P)
       =\min_{E\in\mathcal A(C;P)}\mathcal L^*(P,E),\qquad
    \mathcal L_\infty(P)=
    \begin{cases}
       C_\infty,&a\ge b,\\
       m(P),&a<b.
    \end{cases}
\]
Then
\begin{equation}
    \mathcal L^*(C;P)-\mathcal L_\infty(P)
       \asymp
    \begin{cases}
       P^{-(a-1)}(C/P^{b+1})^{-1/4},&a>b,\\
       P^{-(a-1)}(C/P^{b+1})^{-2/5},&a=b,\\
       P^{-(a-1)}(C/P^{b+1})^{-1},&a<b.
    \end{cases}
    \label{eq:app_stopped_compute_scaling}
\end{equation}
In particular, constant squared teacher coefficients give an excess-loss
compute exponent of \(-2/5\) at fixed data. The \(a<b\) rate assumes
the positive minimum variance specified above.

\paragraph{Scope of the allocation.}
The large-\(E\) conclusions for \(a\ge b\) use the differentiated
late-time estimates along diverging stopping times. The branch \(a<b\)
uses convergence to the finite minimum of the full cross curve.
Appendix~\ref{app:continuum_cross_stopping} establishes these properties
within the pure-power continuum GF equations. For a general discrete
spectrum, the leading cross transient depends on the target coefficients
throughout the spectrum. The exact cancellation giving the \(D/\tau\)
transient follows from constant squared teacher coefficients; matching
the tail exponents alone is insufficient.

The dataset and member size remain fixed across all configurations in
\eqref{eq:main_stopped_allocation}. This problem compares ensembles at
their own minima. The broader problem
\[
    \min_{\tau\ge0,\ E\in\mathbb N_{\ge1}:\,EP\tau\le C}
    \mathcal L(\tau,P,E),\qquad N=P\text{ fixed},
\]
also permits stopping before an ensemble's unconstrained minimum.
Appendix~\ref{app:width_ensemble_allocation} reduces this broader
checkpoint optimization to a scalar problem at fixed width and solves
its large-budget limit for constant squared teacher coefficients at
\(N=P\). It also gives a local comparison with changing width.

\subsection{Width and Ensembling at Fixed Data}
\label{app:width_ensemble_allocation}

Fix a dataset size $P$ and an initial member width $N_0<P$. Two uses of
an additional compute budget are to increase the width of one model, or
to train more models of width $N_0$. Allowing each configuration to stop
at any affordable time gives the comparison
\begin{equation}
 \inf_{\substack{N\in\mathbb N\cap[N_0,P],\ \tau\ge0\\N\tau\le C}}
       \mathcal L(\tau,N,P,1)
 \qquad\text{and}\qquad
 \inf_{\substack{E\in\mathbb N_{\ge1},\ \tau\ge0\\EN_0\tau\le C}}
       \mathcal L(\tau,N_0,P,E).
 \label{eq:app_width_ensemble_budget}
\end{equation}
Here the stopping time is optimized within the budget, whereas
\eqref{eq:app_stopped_cost} compares configurations at their unconstrained
loss minima.

\paragraph{Allocation at a finite compute budget.}
At fixed $P$, write $c(\tau,N)=C_{\mathrm{cross}}(\tau,N)$ and
$V(\tau,N)=C_{\mathrm{self}}(\tau,N)-c(\tau,N)\ge0$.
For each width and checkpoint, loss decreases with ensemble size.
The optimal integer ensemble at that checkpoint is therefore
$\lfloor C/(N\tau)\rfloor$, giving the exact reduction
\begin{equation}
 \inf_{\substack{E\in\mathbb N_{\ge1},\ \tau>0\\EN\tau\le C}}
 \mathcal L(\tau,N,P,E)
 =\inf_{0<\tau\le C/N}
 \left[c(\tau,N)+\frac{V(\tau,N)}{\lfloor C/(N\tau)\rfloor}\right].
 \label{eq:app_budget_integer_reduction}
\end{equation}
The initial loss is included through the limit $\tau\downarrow0$.
Allowing real $E\ge1$ replaces the objective on the right by
$c(\tau,N)+N\tau V(\tau,N)/C$. At an interior optimum,
\begin{equation}
 \partial_\tau c+\frac NC(V+\tau\partial_\tau V)=0,
 \qquad
 \partial_\tau\mathcal L=-\frac{V}{E\tau}.
 \label{eq:app_budget_stationarity}
\end{equation}
Thus, when $V>0$, the optimal checkpoint lies on a decreasing part of
the fixed-$E$ loss curve. This differs from training each ensemble to
its own loss minimum. Equation~\eqref{eq:app_budget_integer_reduction}
evaluated at $N=N_0$ gives the ensemble strategy in
\eqref{eq:app_width_ensemble_budget}; the width strategy minimizes
$c+V$ over the displayed feasible widths and checkpoints.

\paragraph{A local comparison of width and ensemble count.}
For a differentiable continuum width, compare
$(N,E,\tau)\mapsto(fN,E,\tau/f)$ and
$(N,E,\tau)\mapsto(N,fE,\tau/f)$ at $f=1$.
Both changes preserve $EN\tau=C$.
Their loss derivatives are respectively
$N\partial_N\mathcal L-\tau\partial_\tau\mathcal L$ and
$-V/E-\tau\partial_\tau\mathcal L$.
The width change has the smaller derivative precisely when
\begin{equation}
 N\partial_N c+\frac{V+N\partial_N V}{E}<0.
 \label{eq:app_budget_width_ensemble_comparison}
\end{equation}
The reduction of training time is the same for both changes and cancels
in this comparison. Width changes the cross correlation and the
variance; ensembling changes only the variance at a fixed checkpoint.
For integer widths and ensembles, the corresponding finite changes
are compared directly using \eqref{eq:app_width_ensemble_budget}.

\citet[Appendix~H.4]{bordelon2024dynamical} show that increasing width
is at least as effective as ensembling in the infinite-data limit,
under monotonicity conditions on their bias and variance terms.
Equation~\eqref{eq:app_budget_width_ensemble_comparison} gives a local
comparison at finite \(P\), without imposing these monotonicity
conditions.

\paragraph{Comparison with stopping each ensemble at its minimum.}
At $N=P$ with constant squared teacher coefficients, the late-time
expansions give $c-C_\infty\sim D/\tau$ and $V\sim H\sqrt\tau$,
with $D,H>0$. Since $c(\tau)>C_\infty$ at each finite time, an
optimal sequence approaching $C_\infty$ has $\tau\to\infty$ and
$E\to\infty$. The leading budget objective is therefore
$D/\tau+(PH/C)\tau^{3/2}$. Its minimum satisfies
\begin{equation}
 \begin{aligned}
 \tau_C&\sim\left(\frac{2DC}{3PH}\right)^{2/5},
 &E_C&\sim\frac{C}{P\tau_C},\\
 \inf_{EP\tau\le C}\mathcal L-C_\infty
 &\sim\frac{5D}{3}\left(\frac{3PH}{2DC}\right)^{2/5}.
 \end{aligned}
 \label{eq:app_budget_constant_teacher}
\end{equation}
Here the infimum is over integer $E\ge1$ and $\tau>0$.
The tail estimates bound the exact objective above and below by
arbitrarily close multiples of the leading objective along minimizing
sequences. Its unique minimum gives \eqref{eq:app_budget_constant_teacher};
rounding $E_C$ changes the variance term by a relative $O(E_C^{-1})$.
The compute exponent remains $-2/5$, while the optimal checkpoint
and leading loss coefficient differ from the allocation in
\eqref{eq:app_stopped_compute_scaling}.

The following results determine the width dependence entering this comparison.

\paragraph{Cross correlations at arbitrary width.}
Use the noiseless finite Gaussian model of
Appendix~\ref{app:finite_constant_target}, now with arbitrary $N$ and
$Q\in\mathbb R^{N\times M}$ having independent $\mathcal N(0,1/N)$ entries.
Keep $M>P$ and zero initialization. For a fixed full-row-rank dataset,
write $G=XX^\top$, $U=X^\top G^{-1/2}$, and $\Pi=UU^\top$.
Let $U_\perp$ be an orthonormal complement of $U$.
The independent Gaussian blocks $QU$ and $QU_\perp$ give
\begin{align}
 \mathbb E_Q[v_Q(\tau)\mid X]
   &=\bigl[I-\Pi+UF_{N,\tau}(G)U^\top\bigr]w^\star,\nonumber\\
 F_{N,\tau}(G)&=\mathbb E_Q
 e^{-\tau G^{1/2}(QU)^\top(QU)G^{1/2}/P}.
 \label{eq:app_general_width_conditional_mean}
\end{align}
Indeed, $QU_\perp$ has zero conditional mean and is independent of the
training predictions. Gaussian symmetry makes $F_{N,\tau}$ commute with
$G$, giving the displayed row-space term. Since
$0\preceq F_{N,\tau}\preceq I$, the conditional mean is a Euclidean
contraction and $C_{\mathrm{cross}}(\tau)\le
\lambda_{\max}(\Lambda)\|w^\star\|^2$ for every $N$ and $\tau$.
For $(w_k^\star)^2=q>0$, coordinate-sign symmetry further gives
\begin{equation}
 C_{\mathrm{cross}}(\tau)
 =C_\infty+q\mathbb E_X\operatorname{tr}
 [\Lambda UF_{N,\tau}^2U^\top],\qquad
 C_\infty=q\mathbb E_X\operatorname{tr}[\Lambda(I-\Pi)].
 \label{eq:app_general_width_cross_trace}
\end{equation}
The matrices $F_{N,\tau}$ are simultaneously diagonal in an eigenbasis
of $G$ and decrease with $\tau$, so cross is nonincreasing at every width.
For $N\ge P$, $F_{N,\tau}\to0$; for $N<P$, its limit is the mean
projection onto the training kernel's null space. Thus finite width can
also change the limiting cross correlation below interpolation.

\paragraph{Fully trained models above interpolation.}
For an arbitrary deterministic teacher and $N>P+1$, the exact finite-model
loss is
\begin{equation}
 \begin{aligned}
 \mathcal L(\infty,N,P,E)
 &=C_\infty+\frac{K(P)}{E(N-P-1)},\\
 C_\infty&=\mathbb E_X\|(I-\Pi)w^\star\|_\Lambda^2,\\
 K(P)&=\mathbb E_X\!\left[
 \operatorname{tr}[\Lambda(I-\Pi)]\|\Pi w^\star\|^2\right].
 \end{aligned}
 \label{eq:app_general_width_exact_plateau}
\end{equation}
To see this, put $A=QU$, $B=QU_\perp$, and $u=U^\top w^\star$.
The trained weights are $\theta_\infty=A(A^\top A)^{-1}u$, and
$v_Q(\infty)=(I-\Pi)w^\star-U_\perp B^\top\theta_\infty$.
Averaging over $B$ gives the variance
$\operatorname{tr}[\Lambda(I-\Pi)]\|\theta_\infty\|^2/N$.
The Wishart inverse mean
$\mathbb E_A(A^\top A)^{-1}=N I_P/(N-P-1)$ and the ensemble identity
then yield \eqref{eq:app_general_width_exact_plateau}.
If $(w_k^\star)^2=q$, sign symmetry gives $K(P)=P C_\infty$.
At $N\gg P$, the converged variance is consequently
$K(P)/(NE)\,[1+O((P+1)/N)]$.

\paragraph{Finite training time.}
The cross correlation also depends on width before convergence. Fix
$P,M,\tau>0$ and put $K_0=G/P$. Expanding the exponential in
\eqref{eq:app_general_width_conditional_mean} in the centered Wishart
fluctuation gives
\begin{align}
 F_{N,\tau}&=e^{-\tau K_0}+N^{-1}F_1(\tau)+o(N^{-1}),\nonumber\\
 F_1(\tau)&=\int_0^\tau(\tau-v)e^{-(\tau-v)K_0}
 \left[K_0^2e^{-vK_0}
       +K_0\operatorname{tr}(K_0e^{-vK_0})\right]dv.
 \label{eq:app_general_width_finite_time}
\end{align}
The two terms follow from
$N\mathbb E[\Delta_{ij}\Delta_{kl}]
=(K_0)_{ik}(K_0)_{jl}+(K_0)_{il}(K_0)_{jk}$ for
$\Delta=G^{1/2}A^\top A G^{1/2}/P-K_0$.
The third-order remainder is bounded by a fixed-time constant times
$\mathbb E\|\Delta\|^3=O(N^{-3/2})$; finite Gaussian data moments
also permit averaging over $X$. Every eigenvalue of $F_1(\tau)$ is
positive. For constant squared teacher coefficients,
\begin{equation}
 C_{\mathrm{cross}}(\tau,N)
 =C_{\mathrm{cross}}(\tau,\infty)+\frac{A(\tau)}N+o(N^{-1}),
 \qquad
 A(\tau)=2q\mathbb E_X\operatorname{tr}
 [\Lambda Ue^{-\tau K_0}F_1(\tau)U^\top]>0.
 \label{eq:app_general_width_cross_correction}
\end{equation}
This correction remains when $E\to\infty$. Thus the equivalence of
width and ensemble count in the converged variance does not extend
directly to the finite-time cross correlation.

\paragraph{Stationary correlations in the GF closure.}
For positive Laplace variables $z,z'$, set $h=z/(R_1R_3)$ and
$h'=h(z')$. With $S(h)=\sum_k\lambda_k/(\lambda_k+h)$, the
general-width response equations read
\begin{equation}
 \begin{aligned}
 R_1=1-\frac{S(h)}P,\qquad R_3=1-\frac{S(h)}N,
 \qquad
 z=h\left(1-\frac{S(h)}P\right)
    \left(1-\frac{S(h)}N\right).
 \end{aligned}
 \label{eq:app_general_width_responses}
\end{equation}
Define the spectral functions
\[
 (I_1,I_2,U,W)=\sum_k
 \frac{\bigl(\lambda_k,\lambda_k^2,
     \lambda_k(w_k^\star)^2,\lambda_k^2(w_k^\star)^2\bigr)}
      {(\lambda_k+h)(\lambda_k+h')}.
\]
Here $U$ denotes a scalar spectral function. Solving the self system in
\eqref{eq:app_gf_C0_fourier} and \eqref{eq:app_gf_C2_fourier}, together
with the GF limit of \eqref{eq:app_C0_cross}, gives
\begin{align}
 \widehat C_{\mathrm{cross}}(z,z')
 &=\frac{hh'}{zz'}\frac{U}{1-I_2/P},\nonumber\\
 \widehat C_{\mathrm{self}}(z,z')
 &=\frac{hh'}{zz'}
 \frac{(1-I_2/N)U+I_1W/N}
 {(1-I_2/P)(1-I_2/N)-hh'I_1^2/(NP)}.
 \label{eq:app_general_gf_correlations}
\end{align}
For the stationary branch at $N/P\ne1$, $h\to h_0>0$ with
$S(h_0)=m=\min(N,P)$. Writing $d=m-I_2(h_0,h_0)=h_0I_1(h_0,h_0)>0$,
the self denominator tends to
$(1-m/\max(N,P))d/m$. The stationary values follow from the
$1/(zz')$ residues, assuming the final-value limit for this branch.

For the pure-power continuum spectrum of
Appendix~\ref{app:continuum_cross_stopping}, put
$r=1/b$, $p=(a-1)/b$, $c_b=\pi/[b\sin(\pi/b)]$, and
$K=\pi/[b\sin(\pi p)]$. Direct integration gives
\[
 S(h)=c_bh^{-r},\quad h_0=(c_b/m)^b,\quad
 I_2=(1-r)m,\quad
 U=K(1-p)h_0^{p-2},\quad W=Kp h_0^{p-1}.
\]
Substitution in \eqref{eq:app_general_gf_correlations} yields
\begin{equation}
\begin{aligned}
 N<P:\qquad
 C_{\mathrm{self}}(\infty)
   &=\frac{K(c_b/N)^{a-1}}{1-N/P},\\
 C_{\mathrm{cross}}(\infty)
   &=\frac{K(1-p)(c_b/N)^{a-1}}{1-(1-r)N/P},\\
 N>P:\qquad
 C_{\mathrm{cross}}(\infty)
   &=\frac{K(1-p)}r(c_b/P)^{a-1},\\
 C_{\mathrm{self}}(\infty)-C_{\mathrm{cross}}(\infty)
   &=K(c_b/P)^{a-1}\frac P{N-P}.
\end{aligned}
 \label{eq:app_width_continuum_plateaus}
\end{equation}
The denominator $N-P$ in the continuum variance agrees to leading order
with $N-P-1$ in the exact finite-model expression at fixed $N/P>1$.
For $a=b=2$ and $\nu=N/P<1$, the ensemble plateau simplifies to
\begin{equation}
 \mathcal L(\infty,N,P,E)=\frac{\pi^2}{4P}
 \left[\frac1{\nu(2-\nu)}
       +\frac1{E\nu(1-\nu)(2-\nu)}\right].
 \label{eq:app_underparameterized_ensemble_plateau}
\end{equation}
Increasing width changes both terms, while increasing $E$ reduces the
second. The finite-budget comparison in
\eqref{eq:app_width_ensemble_budget} uses the full trajectories to account
for early stopping near interpolation.

\paragraph{Full training curves below interpolation.}
For noiseless GF with zero initialization, take $a=b=2$,
$\lambda_k=k^{-2}$, and $(w_k^\star)^2=1$, with spectral sums
replaced by integrals over $k\in(0,\infty)$. Define
\[
 x=\tau/P^2,\qquad \nu=N/P\in(0,1],\qquad c_2=\pi/2,
 \qquad d=\frac{c_2}{2}(\nu^{-1}-1),\qquad A=c_2+d.
\]
Let $S_\nu(x,x')=P C_{\mathrm{self}}(P^2x,P^2x')$ and
$X_\nu(x,x')=P C_{\mathrm{cross}}(P^2x,P^2x')$ at $N=\nu P$,
with $S_\nu(x)=S_\nu(x,x)$ and $X_\nu(x)=X_\nu(x,x)$.
The normalized ensemble loss is
$P\mathcal L(P^2x,\nu P,P,E)=X_\nu(x)+[S_\nu(x)-X_\nu(x)]/E$.
In the response equation, $w=P\sqrt h$ and $\zeta=P^2z$ satisfy
\[
 \zeta=(w-c_2)(w-c_2/\nu),\qquad
 w=A+u,\qquad u=\sqrt{\zeta+d^2}.
\]
Equation~\eqref{eq:app_general_gf_correlations} then gives the
two-time Laplace transforms of the normalized correlations as
\begin{equation}
 \widehat S_\nu(\zeta,\zeta')
   =\frac{c_2ww'}{\zeta\zeta'(u+u')},\qquad
 \widehat X_\nu(\zeta,\zeta')
   =\frac{c_2ww'}{\zeta\zeta'(u+u'+c_2/\nu)}.
 \label{eq:app_subcritical_full_transforms}
\end{equation}
Here the two time variables are both scaled by $P^2$.
Define
\begin{align}
 F(r,x)&=\frac12\left[
 e^{-dr}\operatorname{erfc}\!\left(\frac r{2\sqrt x}-d\sqrt x\right)
 +e^{dr}\operatorname{erfc}\!\left(\frac r{2\sqrt x}+d\sqrt x\right)
 \right],\nonumber\\
 f(r,x)&=AF(r,x)-\partial_rF(r,x).
 \label{eq:app_subcritical_inverse_kernel}
\end{align}
Since $\widehat F(r,\zeta)=e^{-r\sqrt{\zeta+d^2}}/\zeta$,
factorizing the denominators in
\eqref{eq:app_subcritical_full_transforms} yields the positive integrals
\begin{equation}
 S_\nu(x)=c_2\int_0^\infty f(r,x)^2\,dr,\qquad
 X_\nu(x)=c_2\int_0^\infty e^{-c_2r/\nu}f(r,x)^2\,dr.
 \label{eq:app_subcritical_full_curves}
\end{equation}
These formulas hold for every $x>0$ in the continuum closure, including
$\nu=1$. For $\nu<1$, $f(r,x)\to(c_2/\nu)e^{-dr}$ as $x\to\infty$,
recovering \eqref{eq:app_underparameterized_ensemble_plateau}.

\begin{theorem}[Ensembling below interpolation]
\label{thm:app_subcritical_ensemble_advantage}
Under the preceding continuum assumptions, every single model with
$0<N\le P$ satisfies, at every positive training time,
\begin{equation}
 \mathcal L(\tau,N,P,1)
 \ge\frac{c_2^2}{P}(1+P/N)\ge\frac{\pi^2}{2P}.
 \label{eq:app_single_model_all_time_bound}
\end{equation}
Fix $N_0=\nu_0P$ with $1-1/\sqrt2<\nu_0<1$. Any integer ensemble
count satisfying
\begin{equation}
 E>\frac1{(1-\nu_0)[2\nu_0(2-\nu_0)-1]}
 \label{eq:app_sufficient_ensemble_count}
\end{equation}
achieves strictly lower loss at a finite training time than every
single model with $N\in[N_0,P]$, allowing each single model to choose
its stopping time. Consequently, the ensemble strategy in
\eqref{eq:app_width_ensemble_budget} has strictly smaller loss for all
sufficiently large compute budgets. The sufficient counts are $E\ge5$
at $N_0=P/2$ and $E\ge14$ at $N_0=P/3$.
\end{theorem}
\begin{proof}
At fixed $x>0$, $F(0,x)=1$ and $F(\infty,x)=0$. With
$J=\int_0^\infty F^2\,dr$ and
$K=\int_0^\infty(\partial_rF)^2\,dr$, integration by parts gives
$\int F\partial_rF\,dr=-1/2$, hence $JK\ge1/4$. Therefore
\[
 S_\nu(x)=c_2(A+A^2J+K)
 \ge2c_2A=c_2^2(1+\nu^{-1}).
\]
This proves the bound uniformly over training time and width.
At the fixed starting width, the ensemble plateau is
\[
 \lim_{\tau\to\infty}P\mathcal L(\tau,N_0,P,E)
 =\frac{c_2^2}{\nu_0(2-\nu_0)}
   \left[1+\frac1{E(1-\nu_0)}\right].
\]
Condition~\eqref{eq:app_sufficient_ensemble_count} places this limit
strictly below $2c_2^2$. Convergence therefore gives a finite
$\tau_0$ at which the ensemble beats the single-model bound. The same
checkpoint is feasible for every $C\ge EN_0\tau_0$.
\end{proof}

\paragraph{Large budgets at a fixed width below interpolation.}
Keep $\nu\in(0,1)$ fixed and set $Q=C/P^3$. To optimize the stopping
time within the budget, define the normalized loss
\begin{equation}
 J_\nu(Q)=\inf_{\substack{E\in\mathbb N_{\ge1},\ x>0\\E\nu x\le Q}}
 \left\{X_\nu(x)+\frac{S_\nu(x)-X_\nu(x)}E\right\}.
 \label{eq:app_subcritical_budget_definition}
\end{equation}
Thus $J_\nu(Q)/P$ uses any affordable checkpoint, as in
\eqref{eq:app_width_ensemble_budget}. Its large-budget asymptotics are
\begin{align}
 x_\star(Q)&=\frac1{d^2}
  \left[\log Q-\frac32\log\log Q+O(1)\right],
 &E_\star(Q)&\sim\frac{d^2Q}{\nu\log Q},\nonumber\\
 J_\nu(Q)-\frac{c_2^2}{\nu(2-\nu)}
 &\sim\frac{4\nu^2}{(1-\nu)^3(2-\nu)}\frac{\log Q}{Q}.
 \label{eq:app_subcritical_large_budget}
\end{align}
The floor $c_2^2/[P\nu(2-\nu)]$ exceeds the critical-width floor
$c_2^2/P$. These asymptotics hold at fixed $\nu<1$; the gap $d^2$
vanishes as $\nu\to1$.

To obtain \eqref{eq:app_subcritical_large_budget}, first note that
$X_\nu(x)$ strictly decreases to its plateau. Indeed,
$\partial_xF>0$ for $r>0$,
$\partial_xF=\partial_r^2F-d^2F$, and integration by parts gives
\[
 X_\nu'(x)=2c_2\int_0^\infty e^{-c_2r/\nu}
 (\partial_rf-df)\partial_xF\,dr<0.
\]
The sign follows from
$\partial_rf-df=(c_2/\nu)(\partial_rF-dF)-\partial_xF$ and
$\partial_rF-dF<0$, directly from
\eqref{eq:app_subcritical_inverse_kernel}.
Expansion of $\widehat f=(A+\sqrt{\zeta+d^2})
e^{-r\sqrt{\zeta+d^2}}/\zeta$ at $\zeta=-d^2$ gives
\[
 f(r,x)-\frac{c_2}{\nu}e^{-dr}
 \sim\frac{1-Ar}{2d^2\sqrt\pi}\,x^{-3/2}e^{-d^2x}.
\]
Integrating the leading term in
\eqref{eq:app_subcritical_full_curves}, and differentiating the same
inverse transform, yields
\begin{align}
 X_\nu(x)-\frac{c_2^2}{\nu(2-\nu)}
 &\sim K_\nu x^{-3/2}e^{-d^2x},
 &X_\nu'(x)&\sim-d^2K_\nu x^{-3/2}e^{-d^2x},\nonumber\\
 K_\nu&=\frac{2c_2^2}{\nu d\sqrt\pi(d+c_2/\nu)^2}>0,
 &V_\nu(\infty)&=\frac{c_2^2}{\nu(1-\nu)(2-\nu)},
 \label{eq:app_subcritical_cross_tail}
\end{align}
Here $V_\nu=S_\nu-X_\nu$; the same expansion gives
$V_\nu'(x)=O(x^{-3/2}e^{-d^2x})$, hence $xV_\nu'(x)\to0$.
Taking $x$ of order $\log Q$ and $E$ of order $Q/\log Q$ approaches
the cross plateau. Strict decrease of $X_\nu$ and positivity of
$V_\nu$ then force any globally optimal sequence to satisfy
$x_\star\to\infty$ and $E_\star\to\infty$.
For real $E$, eliminating it gives
$X_\nu(x)+\nu xV_\nu(x)/Q$. Its stationary equation becomes
\[
 d^2K_\nu x^{-3/2}e^{-d^2x}
 \sim\frac{\nu V_\nu(\infty)}Q,
\]
which gives the first line of
\eqref{eq:app_subcritical_large_budget}. The variance term dominates
the excess loss at this optimum, with coefficient
$\nu V_\nu(\infty)/d^2=4\nu^2/[(1-\nu)^3(2-\nu)]$.
At late times the fixed-$E$ loss is decreasing near the optimum, so
the budget is saturated. Adjacent integer choices of $E$ give
checkpoints $x=Q/(\nu E)$ separated by
$O((\log Q)^2/Q)=o(1)$ there, preserving these asymptotics.

\subsection{Discrete Spectra at Interpolation}
\label{app:discrete_spectral_coefficients}
The critical coefficient can also be evaluated without replacing the
spectral sums by integrals. Set $N=P$, $\sigma=0$, and
$(w_k^\star)^2=q>0$. Take a finite positive spectrum with $M>P$,
or a summable positive infinite spectrum. Let $h_0$ be the unique positive solution of
\begin{equation}
 S(h_0)=\sum_k\frac{\lambda_k}{\lambda_k+h_0}=P.
 \label{eq:app_discrete_critical_scale}
\end{equation}
In \eqref{eq:app_general_gf_correlations}, $U=qI_1$ and $W=qI_2$, so
the self numerator reduces to $qI_1$. The denominator factors as
\[
 \frac{(P-I_2-\sqrt{hh'}I_1)(P-I_2+\sqrt{hh'}I_1)}{P^2}.
\]
The identities $S(h)=I_2+h'I_1$ and $S(h')=I_2+hI_1$ give
\[
 P-I_2-\sqrt{hh'}I_1
 =\frac P2(R(z)+R(z'))
   +\frac12(\sqrt h-\sqrt{h'})^2I_1.
\]
Since $R(z)=\sqrt{z/h(z)}$ and $h(z)=h_0+O(\sqrt z)$, the joint
low-frequency expansion is
\begin{equation}
 \widehat C_{\mathrm{self}}(z,z')
 =\frac{qP h_0^{3/2}}{zz'(\sqrt z+\sqrt{z'})}
   \bigl[1+O(\sqrt z+\sqrt{z'})\bigr].
 \label{eq:app_discrete_growth_transform}
\end{equation}
Under the same contour-inversion bounds used in
Lemma~\ref{lem:app_joint_critical_expansion},
\eqref{eq:app_double_inverse_equal_time} therefore gives the
discrete-spectrum GF coefficient
\begin{equation}
 C_{\mathrm{self}}(\tau)=H\sqrt\tau+o(\sqrt\tau),
 \qquad H=\frac{2(2-\sqrt2)}{\sqrt\pi}\,qP h_0^{3/2}.
 \label{eq:app_discrete_growth_coefficient}
\end{equation}

For $\lambda_k=k^{-2}$ with infinite cutoff, the sum is explicit:
\[
 S(h)=\frac{\pi}{2\sqrt h}\coth\!\left(\frac\pi{\sqrt h}\right)
       -\frac12.
\]
Consequently, relative to the continuum coefficient
$H_{\mathrm{cont}}=gqP[\pi/(2P)]^3$, the coefficient with cutoff $M$
satisfies, as $P\to\infty$ and $M/P\to\infty$,
\begin{equation}
 \frac{H_M}{H_{\mathrm{cont}}}
 =\left(\frac{P}{P+1/2}\right)^3
  \left[1-\frac{12(P+1/2)}{\pi^2 M}
       +O\!\left(\frac{P^2}{M^2}+e^{-4P}\right)\right].
 \label{eq:app_discrete_growth_correction}
\end{equation}
Here $g=2(2-\sqrt2)/\sqrt\pi$. The prefactor accounts for the
discrete spectrum, and the $M^{-1}$ term accounts for the feature
cutoff. To obtain the latter, subtract
$\sum_{k>M}(1+hk^2)^{-1}=1/(hM)+O(1/(hM^2)+1/(h^2M^3))$
from the infinite sum and expand its root.
These are coefficients of the GF closure with discrete spectral sums.
The exact finite Gaussian-model averages are given separately in
\eqref{eq:app_new_finite_cross_tail} and \eqref{eq:app_new_finite_self_tail}.
Comparisons over the stopping window use the time-dependent correlations.

\paragraph{Cross coefficient for the same spectrum.}
Writing $F(h,h')=zz'\widehat C_{\mathrm{cross}}(z,z')$, the identities above give
\[
 F(h,h_0)=F(h_0,h)=qPh_0,\qquad
 F_{hh'}(h_0,h_0)=\frac{qP}{h_0}
 \left(1+\frac{h_0S''(h_0)}{2S'(h_0)}\right).
\]
Using $h(z)-h_0=-P\sqrt z/[\sqrt{h_0}S'(h_0)]+O(z)$, the
mixed term of $F$ is $\pi D\sqrt{zz'}$, with
\begin{equation}
 D=\frac{qP^3}{\pi h_0^2[S'(h_0)]^2}
 \left(1+\frac{h_0S''(h_0)}{2S'(h_0)}\right)>0.
 \label{eq:app_discrete_cross_coefficient}
\end{equation}
The bracket equals
$\sum_k\lambda_k^2/(\lambda_k+h_0)^3\big/\sum_k\lambda_k/(\lambda_k+h_0)^2>0$.
With the corresponding contour and differentiated remainder bounds,
this is the coefficient of $\tau^{-1}$ in
$C_{\mathrm{cross}}(\tau)-qPh_0$ and of $-\tau^{-2}$ in its derivative.
Discreteness therefore changes both $D$ and $H$ in the late-time
balance $\tau_\star\sim[2D(E-1)/H]^{2/3}$.

\paragraph{Time-dependent corrections for $a=b=2$.}
The same spectral sums determine the correlations throughout the stopping
window. Retain $\lambda_k=k^{-2}$ and $(w_k^\star)^2=q$, and write
\[
 x=\frac{\tau}{P^2},\qquad \zeta=P^2z,\qquad
 w=P\sqrt{h(z)},\qquad c_2=\frac\pi2,\qquad
 \alpha=1+\frac1{2P}.
\]
Here $w'=P\sqrt{h(z')}$ denotes the response at the second transform
argument. The exact infinite-spectrum sum above becomes
\begin{equation}
 \frac{S(h)}P=\frac{c_2}{w}\coth\!\left(\frac{\pi P}w\right)
 -\frac1{2P},\qquad
 \sqrt\zeta=w\left(1-\frac{S(h)}P\right).
 \label{eq:app_discrete_window_response}
\end{equation}
To retain the leading discrete-spectrum correction, replace the
hyperbolic cotangent by one. We use this expansion in stopping
windows where the omitted term is small on the inverse Laplace contours.
The response then satisfies
\[
 w\simeq\frac{c_2+\sqrt\zeta}{\alpha}.
\]
All square roots use the physical branch continued from positive real
$z$. The identities
$I_1=[S(h)-S(h')]/(h'-h)$ and $I_2=S(h)-h'I_1$ give, within the same
approximation,
\begin{equation}
 \frac{I_1}{P^3}\simeq\frac{c_2}{ww'(w+w')},\qquad
 \frac{I_2}{P}\simeq\frac{c_2}{w+w'}-\frac1{2P}.
 \label{eq:app_discrete_window_spectral_functions}
\end{equation}
The coincident-argument values follow by continuity. Substituting these
expressions into \eqref{eq:app_general_gf_correlations} and inverting gives
\begin{equation}
 \begin{aligned}
 C_{\mathrm{self}}^{\mathrm{disc}}(P^2x)
 &\simeq\alpha^{-3}C_{\mathrm{self}}^{\mathrm{cont}}(P^2x),\\
 C_{\mathrm{cross}}^{\mathrm{disc}}(P^2x)
 &\simeq\alpha^{-2}C_{\mathrm{cross}}^{\mathrm{cont}}(P^2x).
 \end{aligned}
 \label{eq:app_discrete_window_curves}
\end{equation}
The continuum curves are those in
\eqref{eq:app_new_exact_flat_curves}, multiplied by $q$.
Thus the leading correction changes the self and cross amplitudes by
different factors, without rescaling $x$.

For the ensemble loss, \eqref{eq:app_discrete_window_curves} is equivalent
to
\begin{equation}
 \mathcal L_E^{\mathrm{disc}}(P^2x)
 \simeq\alpha^{-3}\frac{E_{\mathrm{eff}}}{E}
 \mathcal L_{E_{\mathrm{eff}}}^{\mathrm{cont}}(P^2x),
 \qquad E_{\mathrm{eff}}=1+\alpha(E-1).
 \label{eq:app_discrete_window_ensemble}
\end{equation}
Here the continuum ensemble formula is evaluated at the real argument
$E_{\mathrm{eff}}$; the actual member count remains the integer $E$.
The prefactor is independent of time, so the stopping time follows from
the continuum curve at this effective count. In particular, the
large-$E$ exponent remains $2/3$, with a modified coefficient.

\paragraph{Finite spectral cutoff.}
For $M/P\gg1$, let $\delta=P/M$. Subtracting the leading omitted tail
from the same infinite sum gives
\begin{equation}
 \frac{S_M(h)}P\simeq\frac{c_2}{w}-\frac1{2P}-\frac{\delta}{w^2},
 \qquad
 \alpha w^2-(c_2+\sqrt\zeta)w+\delta\simeq0.
 \label{eq:app_discrete_window_cutoff_response}
\end{equation}
The physical solution is the larger root on the positive real axis,
continued analytically from $\delta=0$:
\[
 w\simeq\frac{c_2+\sqrt\zeta+
 \sqrt{(c_2+\sqrt\zeta)^2-4\alpha\delta}}{2\alpha}.
\]
The associated spectral functions must be corrected together:
\begin{equation}
 \frac{I_1}{P^3}\simeq
 \frac{c_2}{ww'(w+w')}-\frac{\delta}{w^2w'^2},\qquad
 \frac{I_2}{P}\simeq\frac{c_2}{w+w'}-\frac1{2P}.
 \label{eq:app_discrete_window_cutoff_functions}
\end{equation}
The leading cutoff term cancels explicitly from $I_2$, while its
response arguments still depend on $\delta$. Equations
\eqref{eq:app_discrete_window_cutoff_response}--\eqref{eq:app_discrete_window_cutoff_functions}
can therefore be inserted into the full self and cross transforms to
approximate the stopping window. Unlike the infinite-cutoff correction
in \eqref{eq:app_discrete_window_curves}, this correction is not a
constant rescaling of the continuum curves. Its low-frequency limit
recovers the leading cutoff corrections to $H$ and $D$ above.

\subsection{Scaling with Model and Data Size}
\label{app:gf_scaling_model_data}

Consider the noiseless pure-power continuum spectrum
\(\lambda_k=k^{-b}\), \((w_k^\star)^2=k^{b-a}\), integrated over
\(k\in(0,\infty)\), with \(b>1\) and \(1<a<b+1\).
The general GF equations give a common scaling form at each fixed
\(\nu=N/P>0\). To derive it, set
\[
 k=Pu,\qquad h=P^{-b}y,\qquad z=P^{-b}\zeta.
\]
Then \(S(h)/P=c_by^{-1/b}\), and the response equation
\eqref{eq:app_general_width_responses} becomes
\(\zeta=y(1-c_by^{-1/b})(1-c_by^{-1/b}/\nu)\).
Its physical branch is independent of \(P\). The spectral functions in
\eqref{eq:app_general_gf_correlations} scale as
\[
 I_2=P\widetilde I_2,\qquad I_1=P^{b+1}\widetilde I_1,\qquad
 U=P^{2b-a+1}\widetilde U,\qquad W=P^{b-a+1}\widetilde W,
\]
where the functions with tildes depend only on \(y,y'\) and the exponents.
Both feedback denominators are dimensionless, while the numerator after
factoring out \(1/(zz')\) scales exactly as \(P^{1-a}\). Thus either correlation
transform has the exact scaling
\(\widehat C(z,z';P,\nu)=P^{2b+1-a}\widehat f(\zeta,\zeta';\nu)\).
On the physical branch with well-defined inverse transforms, rescaling
the two Laplace contours and applying the ensemble identity gives, at
positive times,
\begin{equation}
    \mathcal L(\tau,N,P,E)
    = P^{-(a-1)}
    F\!\left(\frac{\tau}{P^b},\frac NP,E\right).
    \label{eq:app_gf_similarity_candidate}
\end{equation}
This identity holds within the pure-power continuum closure. For the
discrete model, the continuum approximation predicts collapse of
\(P^{a-1}\mathcal L\) against \(\tau/P^b\) at fixed \(N/P,E\);
its accuracy depends on the discrete spectrum and feature cutoff.
The associated compute variable is
\[
 \frac{C}{P^{b+1}}=E\nu\frac{\tau}{P^b}.
\]
The scaling form fixes these units; determining the optimal allocation
across ratios requires the time dependence of \(F\).

At \(N=P\), the pure-power continuum calculation does supply a
specialization. Put \(x=\tau/P^b\), \(u=c_b^{b/2}\sqrt x\), and
\(f_\infty=P^{a-1}C_\infty\), which is independent of \(P\). Then
\begin{equation}
\begin{aligned}
 F(x,1,E)
  ={}&\left(1-\frac1E\right)f_\infty
         \left[1+\frac{j}{u}+\frac{d}{u^2}+O(u^{-3})\right]\\
     &+\frac{f_\infty}{E}
         \left[\frac{g}{b-a+1}u+O(1)\right],\qquad u\to\infty.
\end{aligned}
    \label{eq:app_gf_similarity_at_interpolation}
\end{equation}
The \(u^{-1}\) term vanishes for constant squared teacher coefficients, \(a=b\).
For \(a<b\), this late expansion does not locate the finite cross
minimum unless that minimum itself lies at large \(u\).

At fixed \(N/P\ne1\), the positive low-frequency determinant in
\eqref{eq:Gamma_noneq_limit} cuts off the critical growth on the regular
stationary branch, giving the plateaus in
\eqref{eq:app_width_continuum_plateaus}. The limit \(x\to\infty\)
therefore depends on the ratio: the critical \(\sqrt{x}/E\) term is
specific to \(\nu=1\). For \(a=b=2\) and \(0<\nu\le1\), the full
continuum curves in \eqref{eq:app_subcritical_full_curves} give the joint
time and aspect-ratio dependence of \(F\) at every \(x>0\), including
\(\nu=1\). A uniform near-interpolation expansion for general exponents
or for the approach from \(\nu>1\) requires the corresponding joint analysis.

\paragraph{Matching the abundant-data and interpolation scales.}
Within the single-model approximation
\eqref{eq:app_compute_bound_loss}, the optimal width reaches order \(P\)
at the same compute scale at which its training time reaches order \(P^b\):
\[
    N_\star(C)\asymp P,\qquad \tau_\star(C)\asymp P^b.
\]
Either matching condition gives
\begin{equation}
    C\asymp P^{b+1}.
    \label{eq:app_effective_data_crossover}
\end{equation}
For \(C\ll P^{b+1}\), the same abundant-data approximation gives
\(N_\star\ll P\) and \(\tau_\star\ll P^b\). The matching estimate
uses \eqref{eq:app_compute_bound_loss}; locating transitions in the
joint optimum over \(N,E,\tau\) requires the full width- and
time-dependent loss.

In unnormalized compute units, the matching cost is of order
\((B/\eta)P^{b+1}\), after order \(P^b/\eta\) updates. One pass through
the data instead takes \(P/B\) updates, or effective time \(\eta P/B\).
The abundant-data approximation treats the effect of reuse on the leading
loss as negligible over the training times considered.

\paragraph{The limit of large ensembles at fixed data.}
For the continuum GF loss at fixed \(N=P\),
\begin{equation}
    \inf_{E\in\mathbb N_{\ge1}}\mathcal L^*(P,E)
       =\inf_{\tau>0}C_{\mathrm{cross}}(\tau)
       =\mathcal L_\infty(P)
       \asymp P^{-(a-1)}.
    \label{eq:app_infinite_compute_data_law}
\end{equation}
The first equality follows from \(V\ge0\) and convergence of
\(\mathcal L(\tau,P,E)\) to cross at every fixed positive time.
For \(a\ge b\), the infimum of cross is its long-time limit
\(C_\infty\); for \(a<b\), it is the smaller value \(m(P)\) attained
at finite time.

To quantify the approach to this limit, consider a loss within
\(\epsilon P^{-(a-1)}\) of \(\mathcal L_\infty(P)\), where
\(0<\epsilon\ll1\). Equation~\eqref{eq:app_stopped_compute_scaling} gives
\begin{equation}
    C_\epsilon\asymp
    \begin{cases}
       P^{b+1}\epsilon^{-4},&a>b,\\
       P^{b+1}\epsilon^{-5/2},&a=b,\\
       P^{b+1}\epsilon^{-1},&a<b.
    \end{cases}
    \label{eq:app_fixed_data_tolerance_cost}
\end{equation}
Because \(\mathcal L_\infty(P)\asymp P^{-(a-1)}\), requiring a
relative excess of at most \(\epsilon\) gives the same exponents.
The \(a<b\) tolerance cost assumes \(\min_{\tau\in \mathcal S(P)}V(\tau)>0\).
The compute needed to approach the fixed-width limit thus depends on
the desired accuracy.

\subsection{Effect of Weak Label Noise}
\label{app:weak_label_noise_stopping}

Under the conditions of Appendix~\ref{app:label_noise_growth},
the signal source has order \(P^{-(a-1)}\), whereas the
additional source from labels has order \(\sigma^2\). Their ratio is
therefore of order
\begin{equation}
    \sigma^2P^{a-1}.
    \label{eq:app_label_noise_source_ratio}
\end{equation}
In particular,
\begin{equation}
    \sigma^2P^{a-1}\ll1
    \label{eq:app_weak_label_noise_condition}
\end{equation}
means that the label-noise source is small relative to the signal source.

Stability of the noiseless stopping time additionally requires an isolated
minimum and control of the noise-induced change in the loss derivative
near that minimum, including the derivative of the cross contribution.
A noisy allocation law requires these controls uniformly along the
ensemble sizes and stopping times being compared. For constant squared
teacher coefficients, noise can change the leading cross transient
because the noiseless \(\tau^{-1/2}\) coefficient cancels.
The additive irreducible test noise \(\sigma^2\) has zero time
derivative, so the stopping condition depends on the noise-induced
changes in the learned predictor.

\section{Proof of the Finite-Horizon DMFT Error Estimate}
\label{app:error_estimate_proof}

We first prove Lemma~\ref{lem:app_fixed_horizon_stability} and identify the
population limit of the closure. We then compare the finite microscopic
trajectory with that limit to prove
Theorem~\ref{thm:app_finite_horizon_error}.

\subsection{Causal recursion and the population limit}

Use the ordered indices \(i,j\in\{1,\ldots,S\}\). The response matrices
\(\bm H_k,\bm R_1,\bm R_3\) are lower triangular with unit diagonal.
Every matrix to be inverted in
\eqref{eq:app_H_boxed}--\eqref{eq:app_R3_boxed} has the form
\(\bm I+\bm K\), where \(\bm K\) is strictly lower triangular. Hence
\begin{equation}
    (\bm I+\bm K)^{-1}
    =\sum_{m=0}^{S-1}(-\bm K)^m.
    \label{eq:app_causal_finite_inverse}
\end{equation}
At lag \(i-j=d>0\), each feedback term contains a factor \(\bm\Theta\)
of strictly positive lag. Its other response entries therefore have lag
less than \(d\). The response equations can be solved successively in
\(d\), including strictly lower-triangular additive residuals.

Once the responses are determined, order the correlation entries by
\(i+j\). At a fixed pair \((i,j)\), solve them in the order
\(C_0,C_1,C_2,C_3\). Every feedback term in the \(C_0\) equation is
sandwiched between \(\bm H_k\bm\Theta\bm R_3\) and its transpose, so it
uses only indices \(p<i,q<j\). The \(C_1\) equation can then use
\(C_{0,ij}\). The self dependence in \(C_2\) again has a strictly lower
triangular factor on both sides, while its same-pair dependence on
\(C_{1,ij}\) is already known. Finally, \(C_3\) follows from \(C_2\).
Symmetric residuals preserve symmetry throughout this recursion.

This construction proves uniqueness of the causal solution. Every step
uses finite sums and products, including the inverse
\eqref{eq:app_causal_finite_inverse}, so the solution is polynomial in the
residual entries and \(1/N,1/B\). Its derivatives are bounded on a compact
neighborhood of zero residual and \((1/N,1/B)\in[0,1]^2\).
The mean value inequality gives
\eqref{eq:app_causal_stability_bound}, proving the lemma.

Set \(1/N=1/B=0\). The response solution becomes
\begin{equation}
    \bm R_1=\bm R_3=\bm I,
    \qquad
    \bm H_k=(\bm I+\eta\lambda_k\bm\Theta)^{-1},
    \qquad
    (\bm H_k\bm1)_{s+1}=(1-\eta\lambda_k)^s.
    \label{eq:app_closure_population_limit}
\end{equation}
Substitution into \eqref{eq:app_C0_boxed} gives the correlation of
\(\bar{\bm v}_s\) in \eqref{eq:app_error_population_reference}.
Polynomial dependence on \(1/N,1/B\) therefore implies
\begin{equation}
    \max_{0\leq s<S}
    |\mathcal L_{\mathrm{DMFT}}(s+1)-\mathcal L_{\mathrm{pop}}(s)|
    \leq K_{M,S}\left(\frac1N+\frac1B\right).
    \label{eq:app_dmft_population_comparison}
\end{equation}
Here and below, \(K_{M,S}\) denotes a finite constant depending on the
fixed parameters, whose value may change between inequalities.

\subsection{Concentration of the fixed training matrices}

Write
\begin{equation}
    \bm W=\frac{\bm A^\top\bm A}{N},
    \qquad
    \widehat{\bm\Sigma}_t=\frac{\bm\Psi(t)^\top\bm\Psi(t)}B,
    \qquad
    \bm g_t=\frac{\bm\Psi(t)^\top\bm\epsilon(t)}B.
    \label{eq:app_error_sample_matrices}
\end{equation}
Let \(u_N=\sqrt{(M+h)/N}\) and \(u_B=\sqrt{(M+h)/B}\), with \(h\) as
in \eqref{eq:app_fixed_horizon_error_rate}. Gaussian sample-covariance
bounds and a union bound over the \(T\) distinct minibatches give,
with probability at least \(1-\delta\),
\begin{align}
    d_N:=\|\bm W-\bm I\|_{\mathrm{op}}
    &\leq c(u_N+u_N^2),\nonumber\\
    d_B:=\max_{t<T}\|\widehat{\bm\Sigma}_t-\bm\Lambda\|_{\mathrm{op}}
    &\leq c\|\bm\Lambda\|_{\mathrm{op}}(u_B+u_B^2),
    \label{eq:app_sample_covariance_event}\\
    g:=\max_{t<T}\|\bm g_t\|_2
    &\leq
    \sqrt{\frac{\|\bm\Lambda\|_{\mathrm{op}}+d_B}{B}}
    \left(\sqrt M+\sqrt{2h}\right).
    \label{eq:app_sample_noise_event}
\end{align}
For the last inequality, condition on the feature matrices and use
\(\bm g_t\sim\mathcal N(\bm0,\widehat{\bm\Sigma}_t/B)\).
The universal constant \(c\) absorbs the Gaussian tail constants.
All subsequent estimates hold on this event. Repeated uses of a
minibatch involve the same matrices, so the event controls the whole
trajectory.

Choose \(c_0\) in the theorem small enough that
\(\varepsilon=u_N+u_B\leq c_0\) implies
\begin{equation}
    d_N<\frac12,
    \qquad
    \eta(1+d_N)(\|\bm\Lambda\|_{\mathrm{op}}+d_B)\leq1.
    \label{eq:app_sample_step_stability}
\end{equation}
The strict margin \(\rho\) ensures this choice is possible. Also set
\begin{equation}
    \Delta:=\max_{t<T}\|\bm W\widehat{\bm\Sigma}_t-\bm\Lambda\|_{\mathrm{op}}
    \leq(1+d_N)d_B+d_N\|\bm\Lambda\|_{\mathrm{op}}.
    \label{eq:app_sample_operator_difference}
\end{equation}

\subsection{Trajectory and loss comparison}

Flatten the microscopic residual to \(\bm v_s=\bm v_n^0(t)\), where
\(s=nT+t\) and \(t_s=s\bmod T\). Its exact recursion is
\begin{equation}
    \bm v_{s+1}
    =(\bm I-\eta\bm W\widehat{\bm\Sigma}_{t_s})\bm v_s
      -\eta\sigma\bm W\bm g_{t_s},
    \qquad \bm v_0=\bm w^\star.
    \label{eq:app_error_finite_recursion}
\end{equation}
Since \(d_N<1/2\), \(\bm W\) is positive definite. In coordinates
\(\bm z_s=\bm W^{-1/2}\bm v_s\), the homogeneous update is
\(\bm I-\eta\bm W^{1/2}\widehat{\bm\Sigma}_{t_s}\bm W^{1/2}\), a
positive semidefinite contraction by
\eqref{eq:app_sample_step_stability}. It follows that
\begin{equation}
    \|\bm v_s\|_2\leq V_0+\eta\sigma s(1+d_N)g,
    \qquad
    V_0=\sqrt{\frac{1+d_N}{1-d_N}}\|\bm w^\star\|_2.
    \label{eq:app_error_trajectory_bound}
\end{equation}

Let \(\bm e_s=\bm v_s-\bar{\bm v}_s\). Subtracting the population
recursion gives
\begin{equation}
    \bm e_{s+1}
    =(\bm I-\eta\bm\Lambda)\bm e_s
    -\eta(\bm W\widehat{\bm\Sigma}_{t_s}-\bm\Lambda)\bm v_s
    -\eta\sigma\bm W\bm g_{t_s}.
    \label{eq:app_error_difference_recursion}
\end{equation}
The population propagator also has norm at most one. Summing the forcing
terms and using \eqref{eq:app_error_trajectory_bound} yields
\begin{equation}
    \max_{0\leq s<S}\|\bm e_s\|_2\leq Q_S,
    \qquad
    Q_S=\eta S\Delta V_0+
    \eta\sigma S(1+d_N)g\left(1+\frac{\eta S\Delta}{2}\right).
    \label{eq:app_error_trajectory_comparison}
\end{equation}
Equations~\eqref{eq:app_sample_covariance_event}--\eqref{eq:app_sample_operator_difference}
give \(Q_S\leq K_{M,S}\varepsilon\) when \(\varepsilon\leq c_0\).
The test-loss difference therefore satisfies
\begin{align}
    \max_{0\leq s<S}
    |\mathcal L_{N,B}(s+1)-\mathcal L_{\mathrm{pop}}(s)|
    &\leq\|\bm\Lambda\|_{\mathrm{op}}
       \left(2\|\bm w^\star\|_2Q_S+Q_S^2\right)\nonumber\\
    &\leq K_{M,S}\varepsilon.
    \label{eq:app_finite_population_comparison}
\end{align}
The additive test-noise variance cancels in this difference.
Combining \eqref{eq:app_finite_population_comparison} with
\eqref{eq:app_dmft_population_comparison} proves
\eqref{eq:app_dmft_error_estimate}, since \(N^{-1}+B^{-1}\) is bounded
by a constant times \(\varepsilon\) in the stated range.

If \(\ell_S>0\), reduce \(c_0\) further so that
\eqref{eq:app_dmft_population_comparison} is at most \(\ell_S/2\).
Then \(\mathcal L_{\mathrm{DMFT}}(\alpha)\geq\ell_S/2\) for every
\(\alpha\leq S\). Dividing the absolute error bound by this lower bound
gives \eqref{eq:app_dmft_error_simple} and completes the proof.

The argument bounds absolute matrix and trajectory errors and normalizes
only the final loss. The constants depend on the fixed spectrum and horizon.

\clearpage
\section{Numerical Tests of the Random-Feature Theory}
\label{app:rf_numerics}

We test the learning curves, stopping laws, and compute allocations in the
Gaussian random-feature model. The experiments use direct simulations of
the finite model and compare them with the continuum predictions derived
above. Figure~\ref{fig:dmft_examples} gives the separate comparison with the
discrete-time DMFT equations.

The loss curves in Figure~\ref{fig:rf_compute_summary}A, B, and C--D are
also shown in Figures~\ref{fig:rf_compute_allocation}(b),
\ref{fig:rf_width_ensembles}(a), and \ref{fig:rf_batch_size}(a--b),
respectively. Panels A--B use \(P=256\), and C--D use \(P=64\), with
data size fixed within each comparison. The inset in
Figure~\ref{fig:rf_compute_summary}A shows the ensemble size selected by
the stopped allocation described in Appendix~\ref{app:rf_allocation}.

\subsection{Experimental setup}
\label{app:rf_setup}

We use population eigenvalues \(\lambda_k=k^{-b}\) and target coefficients
\((w_k^\star)^2=k^{b-a}\), for \(k=1,\ldots,M\). Thus the target power
in mode \(k\) is \(\lambda_k(w_k^\star)^2=k^{-a}\). All experiments in
this section have \(b=2\), noiseless labels \((\sigma=0)\), and zero
initialization. Unless we vary the target exponent explicitly, \(a=b=2\),
so the squared teacher coefficients are constant. Each configuration uses
20 datasets. Ensemble members share the training data and use independent
Gaussian random-feature matrices.

For GF, we evaluate the finite-model spectral solution at the saved times.
For SGD, we reuse the same ordered minibatches each epoch and evaluate the
exact affine epoch update in \eqref{eq:app_batch_epoch_map}. This permits
late checkpoints to be evaluated by matrix composition. The loss is the
population mean-squared prediction error, evaluated analytically in the
population eigenbasis.

We estimate \(C_{\mathrm{self}}\) by averaging squared errors across a pool
of models and \(C_{\mathrm{cross}}\) by averaging error inner products over
distinct pairs in that pool. Equation~\eqref{eq:app_ensemble_C0} then gives
the exact average loss over all size-\(E\) subsets of the sampled pool.
The dependence on \(E\) in this construction follows from the ensemble
identity; the numerical tests examine the time dependence of the two
correlations and the resulting stopping and allocation rules. Pool sizes
are specified in the captions.

Stopping minimizes the dataset-mean population loss over saved checkpoints.
We use \(\mathcal L^*(P,E)\) for this minimum when \(N=P\).
Uncertainty bands are pointwise 95\% percentile intervals from 1,000
whole-dataset bootstrap resamples, conditional on the sampled model pools.
When a quantity involves selection, the checkpoint and other allowed
choices are reselected within each resample. The curves therefore describe
oracle selection using population risk.

GF compute is \(C=EN\tau\), with the common factor \(B/\eta\) absorbed
into its units as in \eqref{eq:app_compute_budget}. When batch size or
learning rate varies, we instead use
\[
 C_{\mathrm{SGD}}=ENBs,
 \qquad \tau=\eta s,
 \qquad Bs=P\mathcal E.
\]
Each compute comparison holds \(P\) fixed. The cost counts the training
updates represented by a trajectory, rather than the runtime of the
spectral or epoch-update evaluator.

\subsection{Learning curves and ensemble correlations}
\label{app:rf_learning}

\begin{figure}[htbp]
 \centering
 \includegraphics[width=\linewidth]{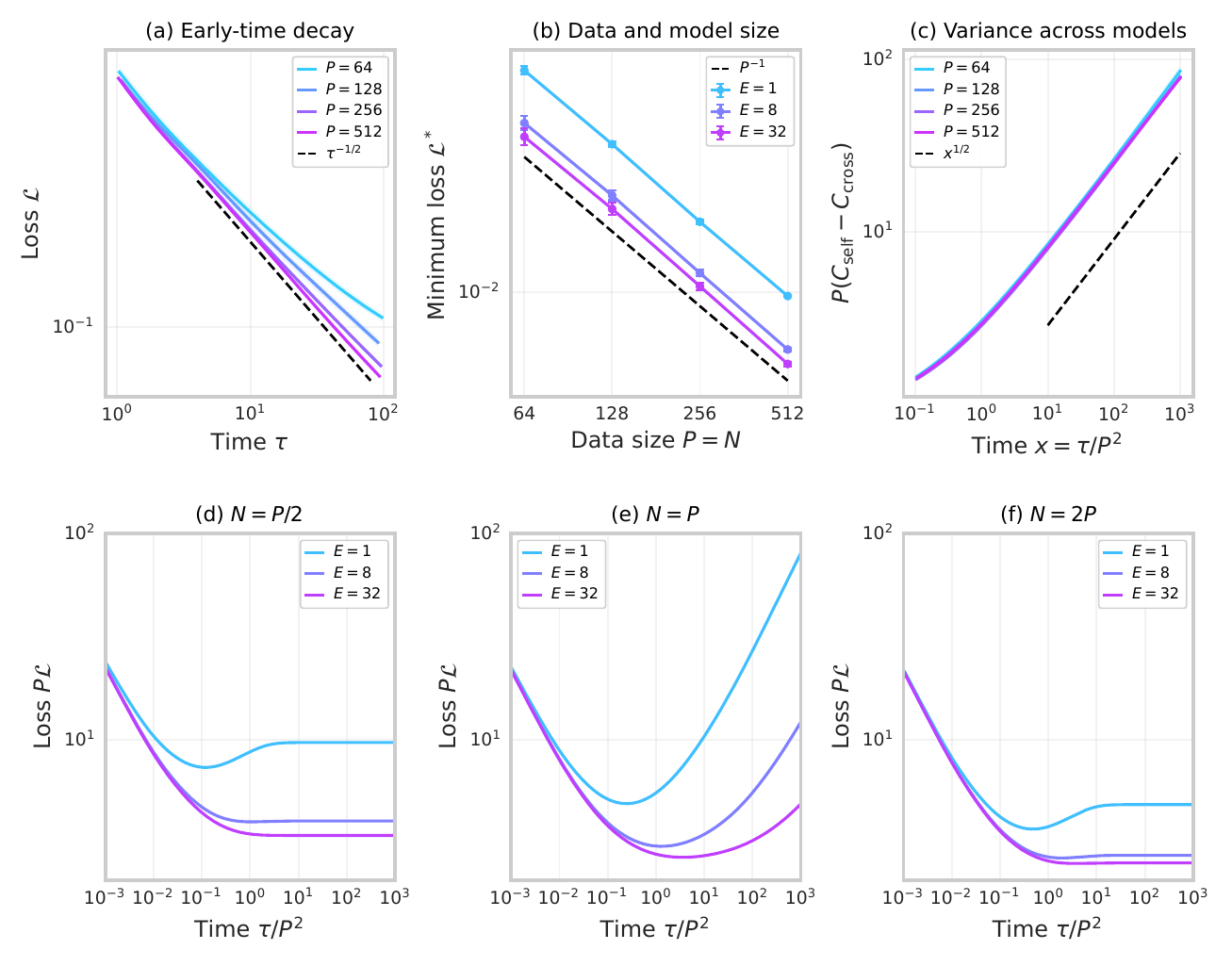}
 \caption{Learning curves in the noiseless random-feature model with
 \(a=b=2\) and \(M=16384\). (a) Single-model loss at early times for
 \(N=P\). (b) Minimum loss versus \(P\), also along \(N=P\), at fixed
 ensemble size. (c) The self-minus-cross correlation, multiplied by \(P\),
 versus scaled time. Dashed lines in (a)--(c) are slope guides with arbitrary
 amplitudes. Bands and bars are dataset-bootstrap intervals.
 (d)--(f) Ensemble learning curves at \(P=256\) and three width-to-data
 ratios. All panels use 20 datasets and 32 models per dataset.}
 \label{fig:rf_learning_curves}
\end{figure}

Figure~\ref{fig:rf_learning_curves} tests the time and data dependence at
\(a=b=2\). Along \(N=P\), the early-time loss approaches the predicted
\(\tau^{-1/2}\) decay as \(P\) increases. The minimum loss is close to
\(P^{-1}\) at each fixed ensemble size. Because width grows with data size
in this comparison, this tests the joint scaling along \(N=P\).
At late times, \(P(C_{\mathrm{self}}-C_{\mathrm{cross}})\) grows approximately
as \((\tau/P^2)^{1/2}\). The lower row shows how the learning curves change
below, at, and above interpolation. Ensembling reduces late-time loss in
all three examples, while the strong growth at \(N=P\) is specific to
the critical ratio.

\begin{figure}[htbp]
 \centering
 \includegraphics[width=\linewidth]{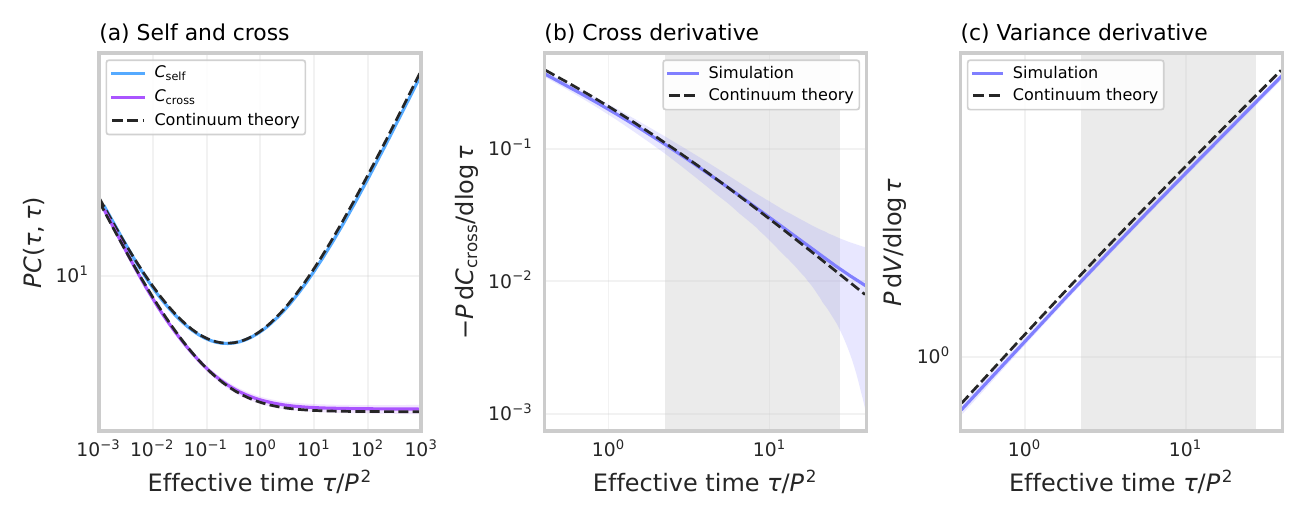}
 \caption{Self and cross correlations at interpolation, with
 \(P=N=256\), \(M=32768\), and \(a=b=2\). The simulations average
 20 datasets and 512 models per dataset. (a) Equal-time correlations.
 (b) The negative cross derivative. (c) The derivative of
 \(V=C_{\mathrm{self}}-C_{\mathrm{cross}}\).
 Gray shading in (b)--(c) spans the stopping times for \(E=16\) through
 \(512\); colored bands are dataset-bootstrap intervals.
 The continuum curves use the
 normalization of \eqref{eq:app_new_exact_flat_curves}, with no fitted
 amplitudes. Derivatives are with respect to \(\log x\), where
 \(x=\tau/P^2\); multiplying the stopping equation by \(\tau\) leaves
 its balance unchanged.}
 \label{fig:rf_correlations}
\end{figure}

Figure~\ref{fig:rf_correlations} resolves the contributions that determine
stopping at \(N=P\). For constant squared teacher coefficients, the
continuum cross correlation approaches its limit as \(1/\tau\), while
the self correlation grows as \(\sqrt\tau\). At an interior minimum,
\[
 -\partial_\tau C_{\mathrm{cross}}
 =\frac1E\partial_\tau(C_{\mathrm{self}}-C_{\mathrm{cross}}).
\]
The derivative comparison uses the full curves in
\eqref{eq:app_new_exact_flat_curves}, including their corrections at finite
time. This balance gives the \(E^{2/3}\) stopping law while retaining
\(1/E\) variance reduction at each fixed checkpoint.

\FloatBarrier
\subsection{Optimal stopping for different teacher spectra}
\label{app:rf_stopping}

We next vary \(a\) at fixed \(b=2\). For each of
\(a\in\{1.5,2,2.5\}\), we use \(N=P\in\{64,128\}\),
\(M=8192\), and pools of 256 models. A paired calculation at \(P=128\)
increases the cutoff to \(M=16384\), retaining the same underlying
Gaussian draws. These settings test the three branches of
\eqref{eq:app_new_continuum_stopping} and
\eqref{eq:app_new_below_cancellation}.

\begin{figure}[htbp]
 \centering
 \includegraphics[width=\linewidth]{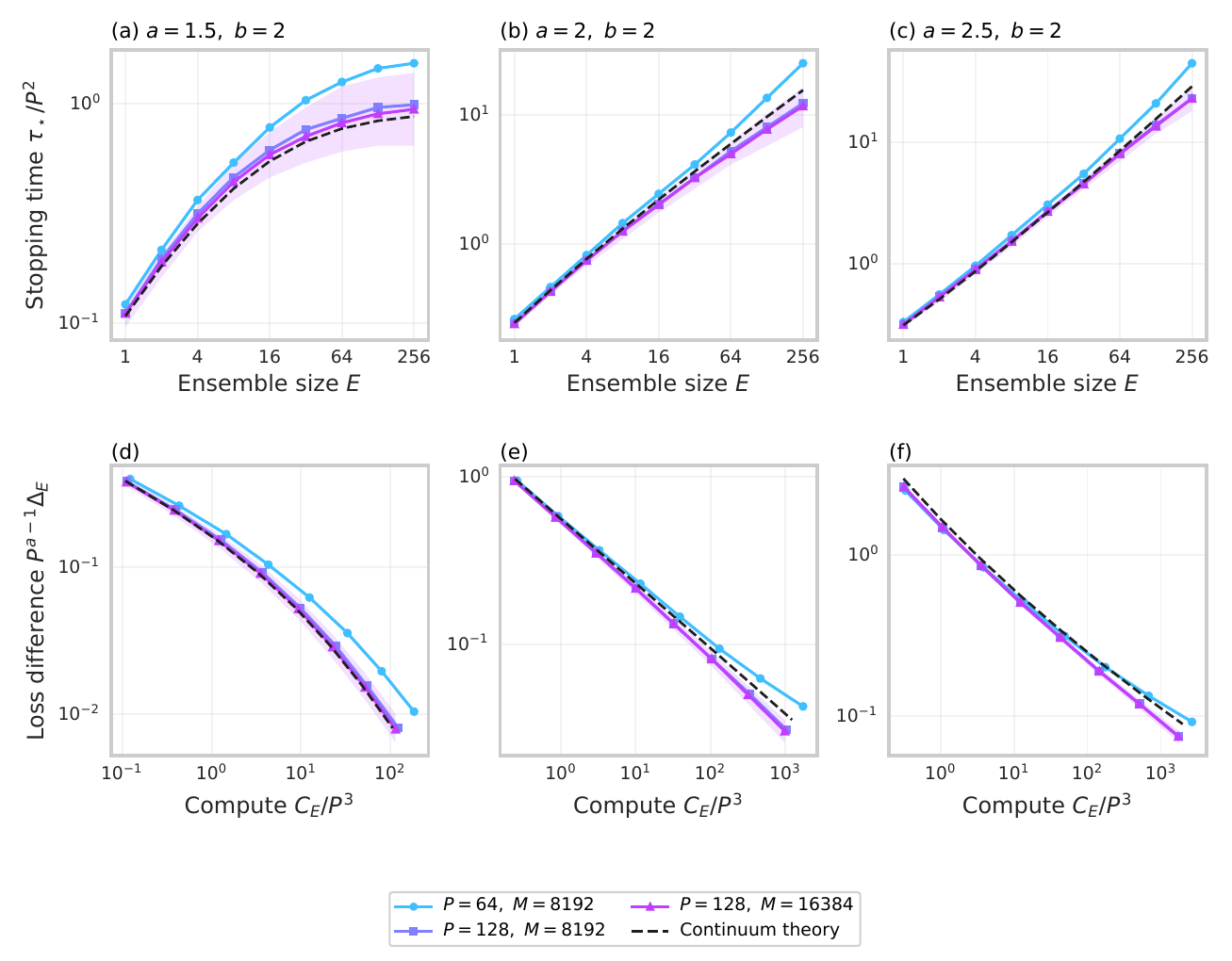}
 \caption{Teacher dependence of stopping and compute scaling at \(N=P\)
 and \(b=2\). Columns vary \(a\). Top: normalized stopping time versus
 ensemble size. Bottom: normalized adjacent-ensemble loss difference
 versus the smaller ensemble's stopping cost. Colors distinguish the
 simulated data sizes and cutoffs; shaded intervals are shown for
 \(P=128,M=16384\). In the lower row, the loss intervals are plotted
 at the central estimate of compute. Dashed curves are the full continuum predictions,
 with no fitted amplitudes. All simulations use 20 datasets and 256 models
 per dataset.}
 \label{fig:rf_teacher_stopping}
\end{figure}

For each ensemble size, we minimize the complete continuum loss
\((1-1/E)C_{\mathrm{cross}}+C_{\mathrm{self}}/E\) to obtain its theoretical
stopping point. The comparison in Figure~\ref{fig:rf_teacher_stopping}
uses these finite-\(E\) predictions. We also measure
\[
 \Delta_E=\mathcal L^*(P,E)-\mathcal L^*(P,2E),
 \qquad C_E=EP\tau_\star(P,E).
\]
This difference removes the common limiting loss without estimating it.
The horizontal coordinate \(C_E\) is the stopping cost of the smaller
ensemble. Each loss in the difference is evaluated at its own minimum.

The stopping time flattens for \(a<b\), whereas it continues to grow for
\(a=b\) and \(a>b\). Table~\ref{tab:rf_stopping_slopes} compares fits
on the same ensemble range. The finite-range continuum predictions fall
within the reported intervals at \(P=128,M=16384\). In particular, for
\(a>b\), the continuum slope in this range is about \(0.86\), before
it reaches its limiting value of one. The \(P=64\) curves show larger
deviations, which remain visible in the figure.

\begin{table}[htbp]
 \centering
 \caption{Stopping-time exponents at \(b=2\), fitted over
 \(E=16,32,64,128,256\). The simulation column uses
 \(P=N=128\), \(M=16384\), and a 256-member pool.
 Brackets give percentile dataset-bootstrap 95\% intervals for the fitted
 exponent. The continuum fit uses the same five ensemble sizes.}
 \label{tab:rf_stopping_slopes}
 \begin{tabular}{@{}cccc@{}}
 \toprule
 \(a\) & Asymptotic exponent & Continuum fit & Simulation fit \\
 \midrule
 1.5 & 0 & 0.167 & 0.172 [0.117, 0.234] \\
 2 & \(2/3\) & 0.704 & 0.633 [0.535, 0.727] \\
 2.5 & 1 & 0.860 & 0.776 [0.705, 0.865] \\
 \bottomrule
 \end{tabular}
\end{table}

\subsection{Compute allocation with fixed data}
\label{app:rf_allocation}

\begin{figure}[htbp]
 \centering
 \includegraphics[width=\linewidth]{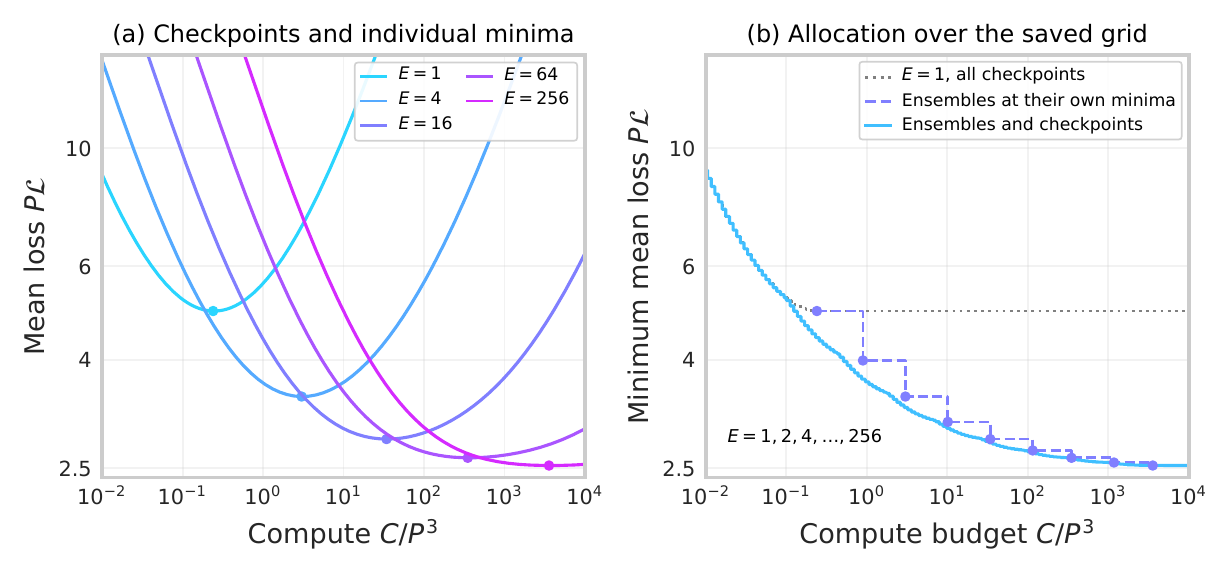}
 \caption{Compute allocation at fixed \(P=N=256\), with
 \(M=16384\), \(a=b=2\), 20 datasets, and a 256-member pool.
 Learning curves are plotted against \(C=EN\tau\), and their minima
 define the stopped configurations in (a). The budget comparison in (b) selects either
 an affordable stopped configuration or any affordable saved checkpoint.
 Ensemble sizes are \(E\in\{1,2,4,8,16,32,64,128,256\}\).
 The single-model checkpoint frontier is included for reference.
 Curves show means, and unused budget is allowed.}
 \label{fig:rf_compute_allocation}
\end{figure}

Figure~\ref{fig:rf_compute_allocation} compares the stopped allocation in
\eqref{eq:main_stopped_allocation} with the budget frontier in
\eqref{eq:app_budget_integer_reduction}. The first chooses among ensembles
that can afford to reach their own loss minimum. The second also allows
earlier checkpoints. Its feasible set is larger, giving lower loss over
part of the compute range. This distinction is visible even with fixed
\(N=P\), and it is separate from the loss-difference scaling in
Figure~\ref{fig:rf_teacher_stopping}.

For the inset in Figure~\ref{fig:rf_compute_summary}A, we first find each
ensemble's minimum dataset-mean population loss over the saved checkpoints.
At each budget, we select the lowest-loss configuration among these minima
whose cost \(EN\tau\) is affordable, and plot its ensemble size.
This is the allocation used by the dashed curve. The ensemble grid is
\(E\in\{1,2,4,8,16,32,64,128,256\}\), so the inset ends at the tested
maximum of 256 members.

\begin{figure}[htbp]
 \centering
 \includegraphics[width=\linewidth]{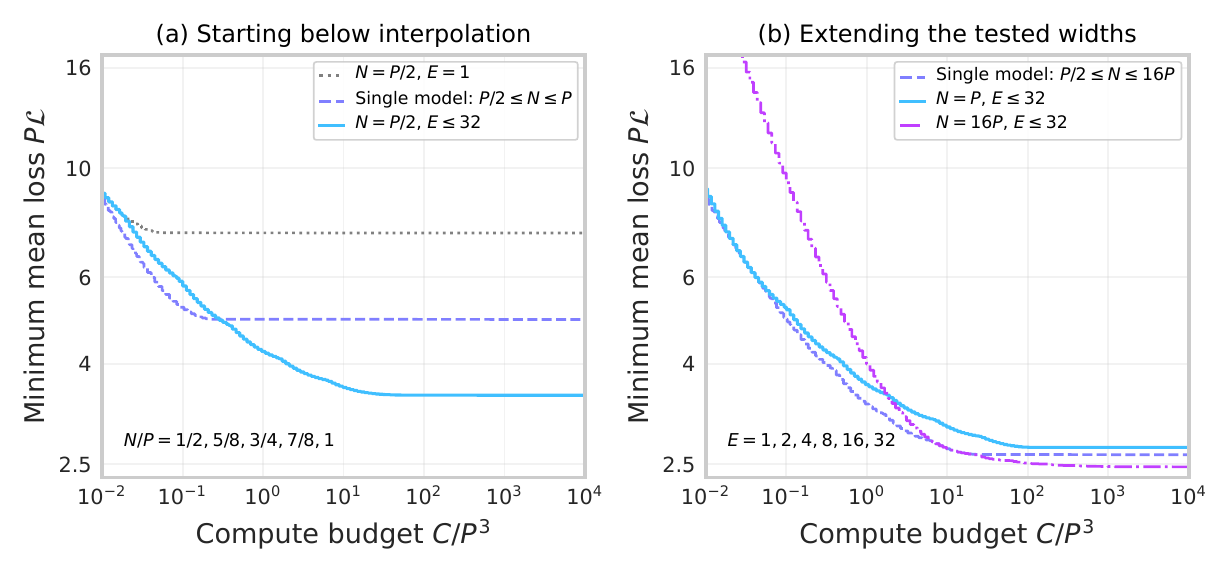}
 \caption{Width and ensemble allocation at fixed \(P=256\), with
 \(M=16384\) and \(a=b=2\). (a) Starting from \(N_0=P/2\), compare
 increasing single-model width through
 \(N/P\in\{1/2,5/8,3/4,7/8,1\}\) with increasing ensemble size at
 fixed \(N=N_0\). (b) Single-model widths through \(16P\), compared
 with ensembles at \(N=P\) and \(N=16P\). The full width grid is
 \(N/P\in\{1/2,5/8,3/4,7/8,1,5/4,3/2,2,4,8,16\}\).
 Each curve minimizes dataset-mean loss over its allowed configurations
 and affordable saved checkpoints. The figure uses 20 datasets and
 32 models per dataset at each width;
 \(E\in\{1,2,4,8,16,32\}\).
 Loss is averaged over datasets before selecting configurations.}
 \label{fig:rf_width_ensembles}
\end{figure}

The width comparison starts from \(N_0=P/2\), as in
\eqref{eq:app_width_ensemble_budget}. At fixed \(P=256\), we compare
increasing the width of one model toward \(P\) with training an ensemble
of models of width \(N_0\). Both strategies choose their best affordable
checkpoint. Figure~\ref{fig:rf_width_ensembles} also extends the width
range above interpolation. From \(N_0=P/2\), increasing width is more
effective at low budgets, while ensembling achieves lower loss at larger
budgets than the tested single models with \(N\le P\).
Allowing widths through \(16P\) changes the comparison: with ensembles
capped at 32 members, the widest single model achieves a lower loss than
the \(N=P\) ensembles at large budgets. Ensembling at \(N=16P\) lowers
the loss further. Thus the comparison depends on the permitted widths
and ensemble sizes.

\FloatBarrier
\subsection{Batch size and learning rate}
\label{app:rf_batch}

We hold \(P=64\) fixed and study \(N=32\) and \(N=64\), with
\(M=4096\), \(a=b=2\), and 128 models per dataset. Batch sizes are
\(B\in\{1,4,16,64\}\). We use the same initial learning-rate grid
\(\eta\in\{0.01,0.04,0.12,0.4,1.2\}\) for every batch size.
A candidate is retained when all its simulated members pass the
fixed-order epoch stability check. This retains 30 of the 40
\((N,B,\eta)\) settings: \(\eta=1.2\) is excluded throughout, and
\(\eta=0.4\) is additionally excluded at \(B=1\). Comparisons use the
remaining grid, with the same 20 datasets for every retained candidate.
After the first epoch, checkpoints are saved at whole-epoch boundaries.

\begin{figure}[htbp]
 \centering
 \includegraphics[width=\linewidth]{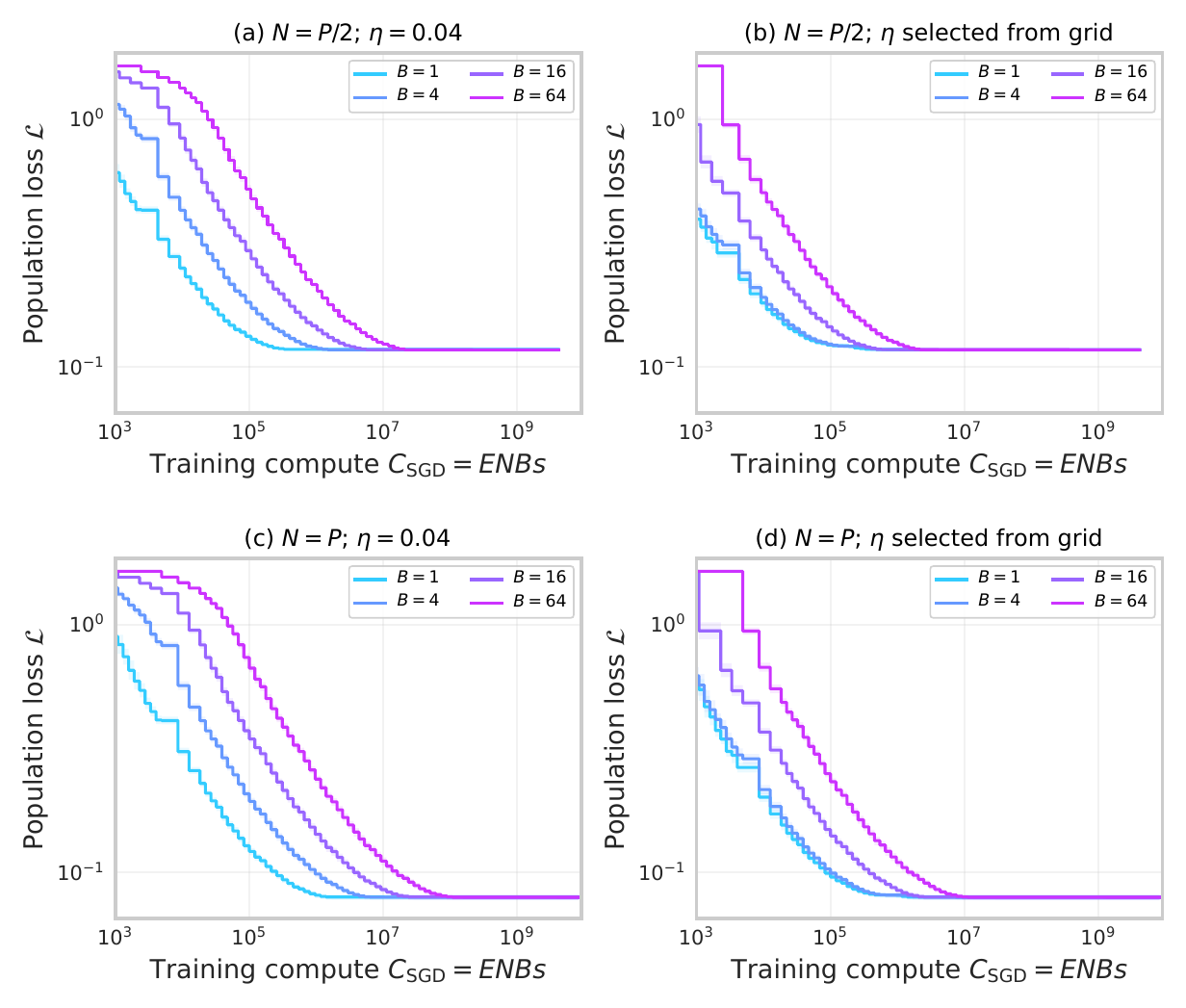}
 \caption{Batch size at matched training compute
 \(C_{\mathrm{SGD}}=ENBs\), with \(P=64\), \(M=4096\), and \(E=1\).
 Rows show \(N=P/2\) and \(N=P\). Left: fixed learning rate
 \(\eta=0.04\). Right: learning rate selected from the retained grid
 separately for each batch size and budget. All curves select an
 affordable checkpoint. Bands are pointwise dataset-bootstrap intervals,
 with selection repeated in each resample. Each setting uses 20 datasets
 and 128 models per dataset.}
 \label{fig:rf_batch_size}
\end{figure}

Figure~\ref{fig:rf_batch_size} first fixes \(\eta=0.04\), then permits
each batch size to select its learning rate under the compute budget.
With fixed \(\eta\), smaller batches make substantially more progress
per unit compute. Learning-rate selection narrows the gap. At
\(C_{\mathrm{SGD}}=10^4\) and \(E=1\), the loss reduction from
\(B=4\) to \(B=1\) falls from \(41.6\%\) to \(5.1\%\) for
\(N=P/2\), and from \(45.9\%\) to \(6.8\%\) for \(N=P\).
This agrees with the time conversion in
Appendix~\ref{app:finite_step_batches}: at fixed compute, the accessible
effective time depends on \(\eta/B\). The discrete epoch update determines
the remaining dependence on batch size and learning rate.

\begin{figure}[htbp]
 \centering
 \includegraphics[width=\linewidth]{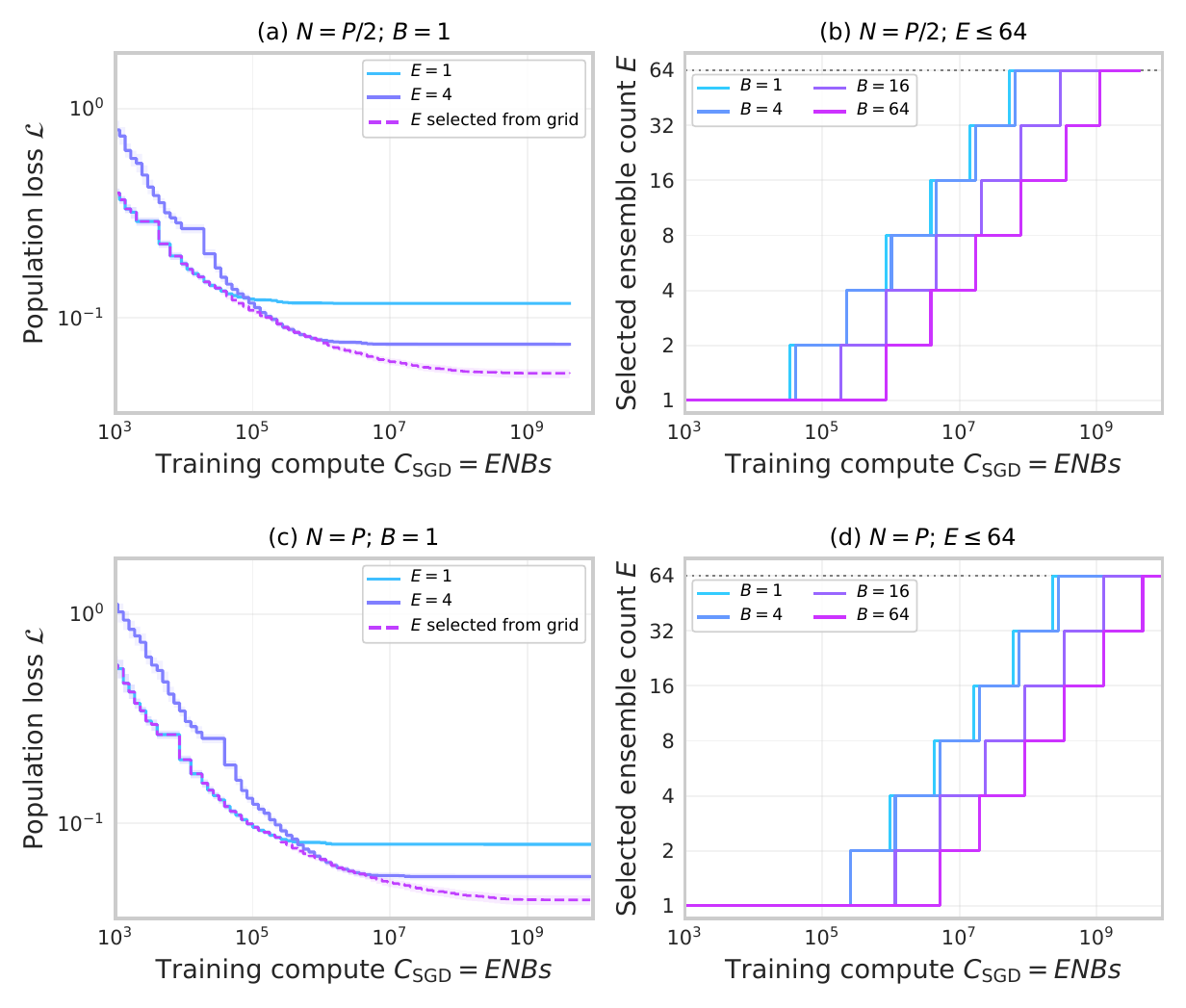}
 \caption{Ensemble allocation for cyclic SGD with the settings of
 Figure~\ref{fig:rf_batch_size}. Left: at \(B=1\), compare fixed
 \(E=1\), fixed \(E=4\), and selection over
 \(E\in\{1,2,4,8,16,32,64\}\), allowing learning-rate and checkpoint
 selection under each budget. Right: selected ensemble size for each
 batch size. Rows show \(N=P/2\) and \(N=P\). The upper ensemble limit
 is 64. At budgets permitting only the zero predictor, ties are assigned
 to \(E=1\).}
 \label{fig:rf_batch_ensembles}
\end{figure}

Figure~\ref{fig:rf_batch_ensembles} then allocates compute over ensemble
size. At low budgets, one model can achieve a lower loss than four models
because it receives more updates. At \(B=1\) and
\(C_{\mathrm{SGD}}=10^6\), four models instead reduce loss by
\(33.7\%\) for \(N=P/2\) and \(17.6\%\) for \(N=P\), after
learning-rate selection. The selected ensemble size increases with budget
and eventually reaches the tested cap of 64. At fixed \(E=1\) or
\(E=4\), the minimum losses across batch sizes are nearly equal at the
largest budgets. In this experiment, the main batch-size effect is the
compute needed to approach those losses.

\FloatBarrier
\subsection{Numerical checks}
\label{app:rf_checks}

\begin{figure}[htbp]
 \centering
 \includegraphics[width=\linewidth]{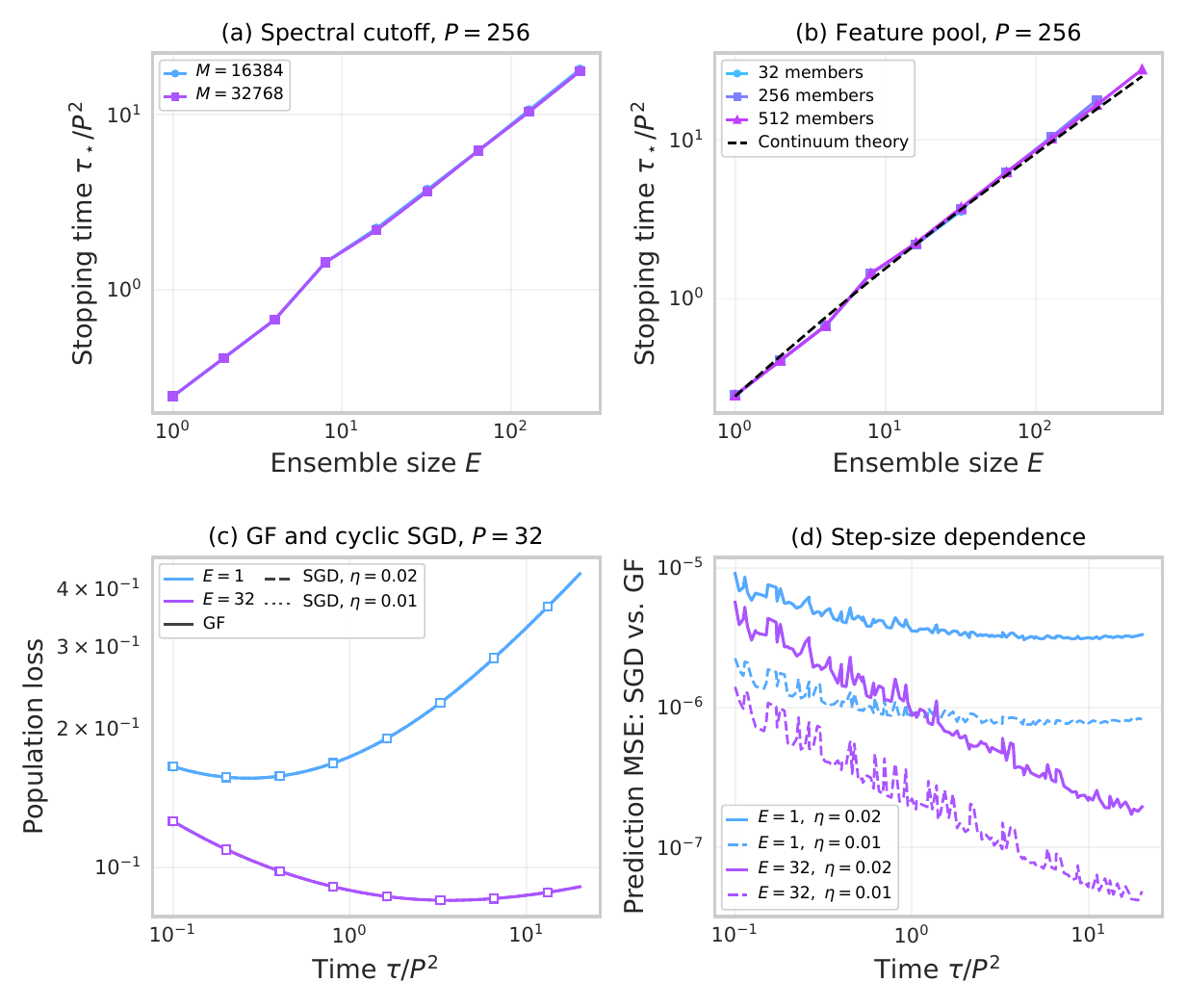}
 \caption{Numerical checks for \(a=b=2\). (a) Stopping times at
 \(P=N=256\) for paired cutoffs \(M=16384\) and \(32768\), using
 256 models per dataset. (b) Nested pools of 32, 256, and 512 models
 at \(P=N=256,M=32768\), together with the continuum prediction.
 (c) GF and fixed-order cyclic SGD at \(P=N=32,M=4096,B=4\), using
 32 models per dataset. (d) Population mean-squared difference between
 the paired SGD and GF predictions. Panels (c)--(d) compare matched
 effective time \(\tau=\eta s\). Every configuration uses 20 datasets.}
 \label{fig:rf_numerical_checks}
\end{figure}

Figure~\ref{fig:rf_numerical_checks} compares spectral cutoffs, feature
pools, and discrete updates. At \(P=N=256\), doubling \(M\) from
16384 to 32768 changes the stopping times only slightly. Increasing the
number of sampled models gives a separate check on the self/cross
estimates. With 512 models, the fitted stopping exponent over
\(E=32,64,128,256,512\) is \(0.716\), with a 95\% interval
\([0.589,0.874]\). This larger-size calculation is consistent with the
constant-teacher stopping law and its finite-\(E\) corrections.

The paired teacher-cutoff comparison in
Figure~\ref{fig:rf_teacher_stopping} gives a similar check for all three
target spectra. Doubling \(M\) changes the fitted stopping exponents
from \(0.171,0.652,0.778\) to \(0.172,0.633,0.776\) for
\(a=1.5,2,2.5\), respectively. Absolute losses are more sensitive to the
cutoff for \(a=1.5\), whose target power decays more slowly.
Using the saved analytic derivatives to interpolate stationary points
changes the largest-pool stopping exponents by less than \(0.006\)
across these nine teacher configurations.

For the small-step comparison, GF and cyclic SGD use identical datasets
and feature matrices at \(P=N=32\), \(M=4096\), and \(B=4\).
At matched effective time, the maximum relative loss difference over
\(0.1\le\tau/P^2\le20\) and \(E\in\{1,2,4,8,16,32\}\)
is \(0.127\%\) at \(\eta=0.02\) and \(0.064\%\) at
\(\eta=0.01\). The mean-squared difference between the paired SGD
and GF predictions decreases by approximately a factor of four when the
step size is halved at the final checkpoint. This is consistent with
first-order prediction error in the small-step limit established in
Appendix~\ref{app:finite_step_batches}.

\FloatBarrier

\end{document}